\documentclass[10pt,twocolumn,letterpaper]{article}

\usepackage{cvpr}              
\usepackage[export]{adjustbox}

\usepackage[utf8]{inputenc} 
\usepackage[T1]{fontenc}    
\usepackage{url}            
\usepackage{booktabs}       
\usepackage{amsfonts}       
\usepackage{nicefrac}       
\usepackage{microtype}      
\usepackage{xcolor}         
\usepackage{multirow}
\usepackage{graphicx}
\usepackage{subcaption}
\usepackage{caption}
\usepackage{array}
\usepackage{makecell}
\usepackage{amssymb}
\usepackage{xspace}
\usepackage{amsmath}
\usepackage[table]{xcolor}
\usepackage{bbm}
\usepackage{pifont}
\usepackage{makecell}
\usepackage{enumitem} 
\usepackage{tikz}
\definecolor{cvprblue}{rgb}{0.21,0.49,0.74}
\usepackage[pagebackref,breaklinks,colorlinks,allcolors=cvprblue]{hyperref}
\hypersetup{
    colorlinks=true,
    urlcolor=magenta
}

\def\paperID{3954} 
\def\confName{CVPR}
\def\confYear{2026}

\newcommand{\Enc}{\mathcal{E}}
\newcommand{\Dec}{\mathcal{D}}
\newcommand{\reconEncoder}{\Enc}
\newcommand{\reconDecoder}{\Dec}
\newcommand{\renderEncoder}{\widetilde{\Enc}}
\newcommand{\renderDecoder}{\widetilde{\Dec}}

\newcommand{\imagesIn}{\mathbf{I}}
\newcommand{\sceneMemory}{\mathbf{M}}

\newcommand{\globalpoints}{\mathbf{P}^{\text{global}}}
\newcommand{\localpoints}{\mathbf{P}^{\text{local}}}
\newcommand{\posesRecon}{\mathbf{v}}
\newcommand{\features}{\mathbf{F}}
\newcommand{\conf}{\mathbf{C}}

\newcommand{\taskG}{{t_\text{g}}}
\newcommand{\taskS}{{t_\text{s}}}

\newcommand{\imagesRender}{\widetilde{\mathbf{I}}}
\newcommand{\novelPosesInput}{\mathbf{T}}
\newcommand{\renderGlobalpoints}{{\widetilde{\mathbf{P}}^{\text{global}}}}
\newcommand{\renderLocalpoints}{{\widetilde{\mathbf{P}}^{\text{local}}}}

\newcommand{\renderFeatures}{\widetilde{\mathbf{F}}}
\newcommand{\renderConf}{\widetilde{\mathbf{C}}}
\newcommand{\novelInputfeatures}{\mathbf{X}}

\newcommand{\rendertaskG}{{\widetilde{t}_\text{g}}}
\newcommand{\rendertaskS}{{\widetilde{t}_\text{s}}}

\newcommand{\globalpointsgt}{\mathbf{P}^*}

\newcommand{\featuresgt}{\mathbf{F}^*}

\newcommand{\loss}{{\mathcal{L}}}
\newcommand{\lossGlobalpoints}{{\loss_{\rm glo}}}
\newcommand{\lossLocalpoints}{{\loss_{\rm loc}}}
\newcommand{\lossForcing}{{\loss_{\rm forcing}}}
\newcommand{\lossEncfeatures}{{\loss_{\rm enc}}}

\newcommand{\muster}{{MUSt3R}}

\newcommand{\cuter}{{CUT3R}}
\newcommand{\semantickitti}{\mbox{SemanticKITTI}}
\newcommand{\occDnuscenes}{\mbox{Occ3D-NuScenes}}
\newcommand{\occDwaymo}{\mbox{Occ3D-Waymo}}

\newcommand{\Singleview}{\mbox{Monocular}}
\newcommand{\singleview}{\mbox{monocular}}
\newcommand{\Surround}{\mbox{Surround-view}}
\newcommand{\surround}{\mbox{surround-view}}
\newcommand{\Sequence}{\mbox{Sequence}}
\newcommand{\sequence}{\mbox{sequence}}

\newcommand{\ttva}{{TTVA}}

\newcommand{\nrec}{{N_{rec}}}
\newcommand{\nren}{{N_{rnd}}}

\newcommand{\OURS}{OccAny} 

\definecolor{semantic}{HTML}{7e57c2}     
\definecolor{normal}{HTML}{3b8b5f}       
\definecolor{geometry}{HTML}{b4445c}        
\definecolor{todo}{HTML}{ff0000}

\definecolor{tud2b}{RGB}{0,131,204}
\definecolor{tud3b}{RGB}{0,157,129}
\definecolor{tud4b}{RGB}{153,192,0}
\definecolor{tud5b}{RGB}{201,212,0}
\definecolor{tud6b}{RGB}{253,202,0}
\definecolor{tud7b}{RGB}{245,163,0}

\definecolor{pastelred}{rgb}{1.0, 0.41, 0.38}
\newcommand{\redcross}{\textcolor{pastelred}{\ding{55}}}

\newcommand\cbarm[4][pastelred]{\colorbox{pastelred}{\color{black}\framebox(#2,#3){}}\,\colorbox{white}{\color{white}\framebox(#4,#3){}}}

\definecolor{important}{HTML}{e9e2ff}    

\newcommand{\indomain}[1]{\textcolor{gray}{#1}}
\newcommand{\best}[1]{\textbf{#1}}
\newcommand{\second}[1]{\underline{#1}}

\newcommand*\rot{\rotatebox{90}}

\newcommand{\condenseparagraph}[1]{{\vspace*{0.1em}\noindent\textbf{#1}\quad}}

\usepackage[capitalize]{cleveref}
\crefname{section}{Sec.}{Secs.}
\Crefname{section}{Section}{Sections}
\Crefname{table}{Table}{Tables}
\crefname{table}{Tab.}{Tabs.}

\newcolumntype{C}[1]{>{\centering\arraybackslash}p{#1}}

\definecolor{car}{rgb}{0.39215686, 0.58823529, 0.96078431}
\definecolor{bicycle}{rgb}{0.39215686, 0.90196078, 0.96078431}
\definecolor{motorcycle}{rgb}{0.11764706, 0.23529412, 0.58823529}
\definecolor{truck}{rgb}{0.31372549, 0.11764706, 0.70588235}
\definecolor{other-vehicle}{rgb}{0.39215686, 0.31372549, 0.98039216}
\definecolor{person}{rgb}{1.        , 0.11764706, 0.11764706}
\definecolor{bicyclist}{rgb}{1.        , 0.15686275, 0.78431373}
\definecolor{motorcyclist}{rgb}{0.58823529, 0.11764706, 0.35294118}
\definecolor{road}{rgb}{1.        , 0.        , 1.        }
\definecolor{parking}{rgb}{1.        , 0.58823529, 1.        }
\definecolor{sidewalk}{rgb}{0.29411765, 0.        , 0.29411765}
\definecolor{other-ground}{rgb}{0.68627451, 0.        , 0.29411765}
\definecolor{building}{rgb}{1.        , 0.78431373, 0.        }
\definecolor{fence}{rgb}{1.        , 0.47058824, 0.19607843}
\definecolor{vegetation}{rgb}{0.        , 0.68627451, 0.        }
\definecolor{trunk}{rgb}{0.52941176, 0.23529412, 0.        }
\definecolor{terrain}{rgb}{0.58823529, 0.94117647, 0.31372549}
\definecolor{pole}{rgb}{1.        , 0.94117647, 0.58823529}
\definecolor{traffic-sign}{rgb}{1.        , 0.        , 0.    }  
\definecolor{barrier}{rgb}{0.439, 0.502, 0.565}
\definecolor{bus}{rgb}{1.000, 0.498, 0.314}
\definecolor{construction-vehicle}{rgb}{0.914, 0.588, 0.275}
\definecolor{traffic-cone}{rgb}{0.184, 0.310, 0.310}
\definecolor{trailer}{rgb}{1.000, 0.549, 0.000}
\definecolor{truck}{rgb}{1.000, 0.384, 0.275}
\definecolor{other-flat}{rgb}{0.686, 0.000, 0.294}
\definecolor{manmade}{rgb}{1.000, 0.784, 0.000}

\makeatletter
\newcommand{\car@semkitfreq}{3.92}
\newcommand{\bicycle@semkitfreq}{0.03}
\newcommand{\motorcycle@semkitfreq}{0.03}
\newcommand{\truck@semkitfreq}{0.16}
\newcommand{\othervehicle@semkitfreq}{0.20}
\newcommand{\person@semkitfreq}{0.07}
\newcommand{\bicyclist@semkitfreq}{0.07}
\newcommand{\motorcyclist@semkitfreq}{0.05}
\newcommand{\road@semkitfreq}{15.30}  %
\newcommand{\parking@semkitfreq}{1.12}
\newcommand{\sidewalk@semkitfreq}{11.13}  %
\newcommand{\otherground@semkitfreq}{0.56}
\newcommand{\building@semkitfreq}{14.1}  %
\newcommand{\fence@semkitfreq}{3.90}
\newcommand{\vegetation@semkitfreq}{39.3}  %
\newcommand{\trunk@semkitfreq}{0.51}
\newcommand{\terrain@semkitfreq}{9.17} %
\newcommand{\pole@semkitfreq}{0.29}
\newcommand{\trafficsign@semkitfreq}{0.08}
\newcommand{\semkitfreq}[1]{{\csname #1@semkitfreq\endcsname}}

\definecolor{cvprblue}{rgb}{0.21,0.49,0.74}
\definecolor{plt:green}{HTML}{2ca02c}
\definecolor{plt:red}{HTML}{d62728}

\title{\OURS{}: Generalized Unconstrained Urban 3D Occupancy}

\author{
	Anh-Quan Cao \quad
	Tuan-Hung Vu\vspace{0.2cm}\\
	Valeo.ai, Paris, France\vspace{0.2cm} \\
    \href{https://valeoai.github.io/OccAny}{https://valeoai.github.io/OccAny}
}

\begin{document}
\newcolumntype{H}{>{\setbox0=\hbox\bgroup}c<{\egroup}}
\maketitle

\begin{abstract}
Relying on in-domain annotations and precise sensor-rig priors, existing 3D occupancy prediction methods are limited in both scalability and out-of-domain generalization.
While recent visual geometry foundation models exhibit strong generalization capabilities, they were mainly designed for general purposes and lack one or more key ingredients required for urban occupancy prediction, namely metric prediction, geometry completion in cluttered scenes and adaptation to urban scenarios.
We address this gap and present \emph{OccAny}, the first unconstrained urban 3D occupancy model capable of operating on out-of-domain uncalibrated scenes to predict and complete metric occupancy coupled with segmentation features.
\emph{OccAny} is versatile and can predict occupancy from sequential, monocular, or surround-view images.
Our contributions are three-fold: (i) we propose the first generalized 3D occupancy framework with (ii)
Segmentation Forcing that improves occupancy quality while enabling mask-level prediction, and (iii) a Novel View Rendering pipeline that infers novel-view geometry to enable test-time view augmentation for geometry completion.
Extensive experiments demonstrate that \emph{OccAny} outperforms all visual geometry baselines on 3D occupancy prediction task, while remaining competitive with in-domain self-supervised methods across three input settings on two established urban occupancy prediction datasets. \textbf{Our code is available at \href{https://github.com/valeoai/OccAny}{https://github.com/valeoai/OccAny}} .
\end{abstract}

\section{Introduction}
The innate ability to see and make sense of the world in three dimensions underpins how humans understand and navigate the space.
Advancing 3D scene understanding is crucial for spatial intelligent systems such as autonomous driving, robotics, and augmented reality.
A key task in this area is 3D occupancy prediction whose goal is to infer a voxelized map of the environment and, when required, provide the corresponding semantics.
Despite advances in architecture design~\cite{qi2017pointnet,thomas2019kpconv,choy20194d, wu2024point}, training algorithm~\cite{gsplat, nerf, monoscene, sdformer, SceneDINO} and dataset~\cite{kitti, scannet, shapenet, nuscenes}, current state-of-the-art 3D models still lack the generalization of human perception, typically requiring constrained setup with precise sensor calibration.
While humans can effortlessly infer complex 3D structures in any novel scenes, replicating this capability remains a demanding problem.

\begin{figure} [t!]
	\centering
	\includegraphics[width=0.99\linewidth]{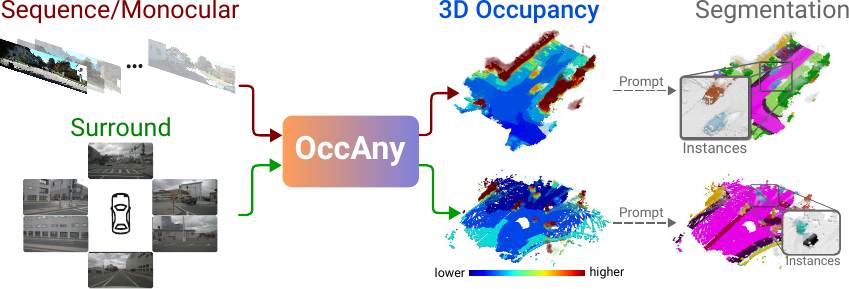}
	\caption{\textbf{\OURS{}} is a generalized 3D occupancy model that is trained once and can operate on out-of-domain sequential, monocular, or surround-view urban images. It produces SAM2-like features, enabling promptable segmentation.}
    \vspace{-0.3cm}
	\label{fig:teaser}
\end{figure}

State-of-the-art supervised approaches for 3D occupancy prediction~\cite{surroundocc, tpvformer, voxformer, sdformer, Wang2024H2GFormerHV, li2023stereoscene, CGFormer} achieve remarkable results when the training and test data are drawn from the same distribution, \ie both are collected using the same or a similar sensor rig under comparable conditions. A core component of these methods is the lifting of 2D features into 3D space, performed either via learnable mechanisms~\cite{voxformer, tpvformer} or via explicit camera modeling~\cite{monoscene, ndcscene}. However, this lifting operation inherently embeds sensor- and domain-specific biases into the models, which limits their ability to generalize to new sensor suites or environments.
Recent self-supervised works~\cite{scenerf, selfocc, bts, gausstr, GaussianOcc} remove the need for 3D supervision by formulating occupancy prediction as a differentiable volume-rendering problem, thereby leveraging advances in neural rendering~\cite{nerf, gsplat}.
Despite this, self-supervised models still struggle to generalize, as they remain specialized to a particular training domain with strong biases in camera poses and intrinsic parameters.
As we look toward a near future with millions of autonomous fleets equipped with different sensor configurations, advancing 3D occupancy prediction requires generalizable and efficient solutions capable of leveraging heterogeneous training data to overcome current generalization barriers.

The advent of visual geometry foundation models~\cite{duster, cuter, vggt, muster}, built around the concept of direct pointmap prediction, has demonstrated the strong generalization potential of large-scale transformer networks for 3D scene understanding.
However, their general-purpose design remains insufficient for urban occupancy prediction, which simultaneously requires metric-scale accuracy, cluttered geometry completion, and adaptation to the complex nature of urban environments.

We introduce a novel pipeline for urban 3D occupancy prediction that emphasizes scalability and generalization.
Our approach follows the recipe of geometry foundation models that train visual transformers with straightforward point-level objectives on diverse, large-scale datasets.
Unlike those prior works, we specialize in the task of occupancy prediction and focus exclusively on outdoor urban datasets, which we argue is essential for optimal adaptation to the unique characteristics of urban scene perception.
A major challenge in outdoor urban scenarios is the sparsity of supervised LiDAR point clouds, which leads to irregular predictions in non-supervised regions and exacerbates the difficulty of geometry completion, particularly in highly cluttered areas.
To address this, we introduce \emph{Segmentation Forcing}, a distillation strategy that enriches geometry-focused features with segmentation awareness and thus helps regularize predictions with consistent segmentation cues of object instances and homogeneous regions.
For geometry completion, we develop a \emph{Novel View Rendering} pipeline that infers arbitrary novel-view geometry from a global scene memory.
Our rendering pipeline enables Test-time View Augmentation, allowing us to densify and complete scenes at both the point- and voxel-levels.
~\cref{fig:teaser} illustrates our model.
In summary, our contributions are three-fold:
\begin{itemize}
\item We propose a generalized 3D occupancy framework, \emph{OccAny}, the first designed to infer dense 3D occupancy and segmentation features for out-of-domain unconstrained urban scenes. \underline{A unified \emph{OccAny} model} can operate on either sequential, monocular or surround-view images.
\item We introduce \emph{Segmentation Forcing}, a novel regularization strategy to mitigate the sparsity of LiDAR supervision.
\item We develop a \emph{Novel View Rendering} pipeline targeting geometry completion.
\end{itemize}
\emph{OccAny} is trained on five urban datasets and evaluated on two out-of-distribution occupancy datasets: SemanticKITTI and Occ3D-NuScenes.
\emph{OccAny} significantly outperforms baseline visual geometry networks and performs on par with domain-specific SOTA self-supervised occupancy networks trained directly on SemanticKITTI and Occ3D-NuScenes.

\section{Related works}
\condenseparagraph{Visual geometry foundation model.}
Dust3r~\cite{duster} introduced the visual geometry foundation model, which uses large-scale pointmap prediction to solve diverse 3D tasks. Research has rapidly expanded this paradigm beyond static, binocular inputs in several directions. One branch addresses dynamics by handling moving scenes~\cite{monst3r, dynamicpointmaps}, dynamic video pose estimation~\cite{AnyCam}, and camera rigs~\cite{rig3r}. A major thrust has been multi-frame processing through feed-forward, sequential, and memory-based architectures~\cite{spann3r, muster, fast3r, vggt, cuter}. Other works have explored downstream tasks such as indoor instance prediction~\cite{panst3r} and image matching~\cite{mast3r}, or have leveraged known camera parameters~\cite{pow3r}. While some methods explore novel view synthesis~\cite{cuter, anysplat}, they often prioritize image synthesis over geometric fidelity~\cite{anysplat} or exhibit limited applicability~\cite{cuter}. Unlike these approaches, we repurpose these models for occupancy prediction by introducing segmentation forcing to enhance geometric fidelity while enabling segmentation output. We further propose a novel pointmap rendering pipeline to enable complete geometry beyond visible scenery.

\condenseparagraph{3D occupancy prediction}. This task, which originates from 3D scene completion~\cite{SSCNet}, aims to assign an occupancy state to each voxel in a 3D volume. Initially proposed for indoor depth scenes~\cite{SSCNet}, it expanded to outdoor LiDAR~\cite{semkitti, lmscnet, pasco, scpnet} and was later adapted for multi-view images~\cite{monoscene}. Subsequent supervised research has focused on projection mechanisms~\cite{monoscene, voxformer, ndcscene}, efficient representations~\cite{tpvformer, GaussianFormer, SparseOcc, zuo2025quadricformer, OSP}, network architectures~\cite{OccFormer, voxformer, COTR}, and benchmark creation~\cite{Cam4DOcc, occ3d, sscbench}. However, these methods' reliance on dense, voxel-wise annotations limits their scalability.

\begin{figure*}[ht!]
	\centering
	\includegraphics[width=0.99\linewidth]{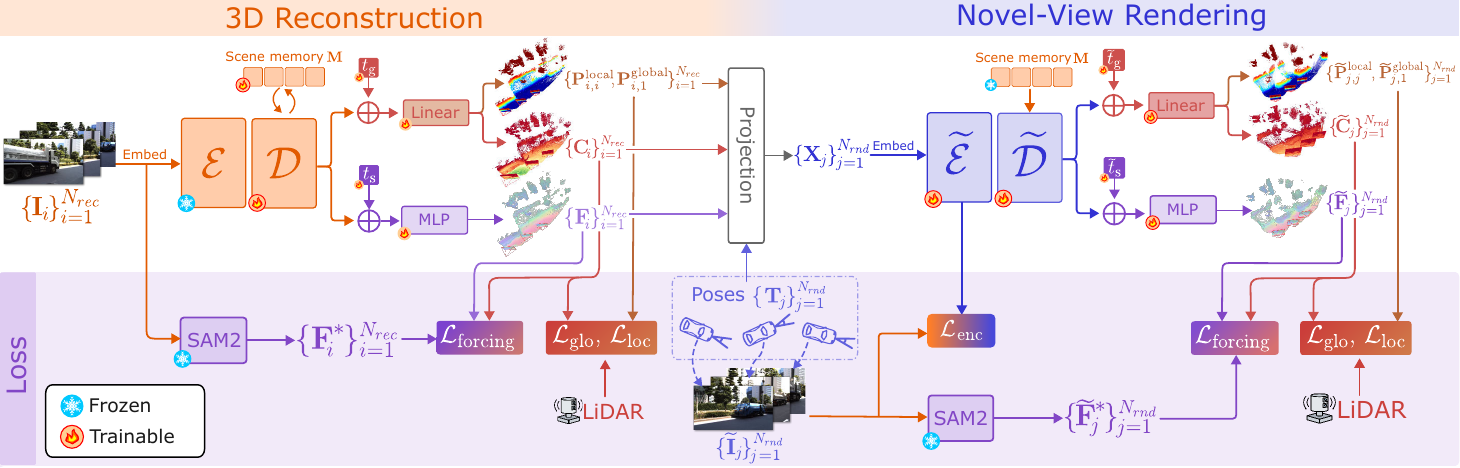}
	\vspace{-0.2cm}
	\caption{\textbf{\OURS{} Training} is done in two stages: (i) \emph{3D Reconstruction} infers 3D scene using $\nrec$ reconstruction frames and (ii) \emph{Novel-View Rendering} renders geometry of $\nren$ new views having camera poses $\{\novelPosesInput_j\}_{j=1}^\nren$. \emph{Segmentation Forcing} with SAM2 features helps regularize and improve geometry prediction.
		The scene memory $\sceneMemory$ is dynamically updated during reconstruction, while during rendering, the final scene memory output from the reconstruction stage is used without updating 
	}
	\vspace{-0.3cm}
	\label{fig:pipeline}
\end{figure*} 

Self-supervised methods mitigate this label dependency by training on posed images, often via volume rendering~\cite{bts,scenerf}. 
Subsequent NeRF-based approaches have improved performance through better losses~\cite{sfourc,selfocc,occnerf}, optimized ray sampling~\cite{occnerf, scenerf, higaussian}, and enhanced representations by distilling foundation models~\cite{distillnerf, OccFeat, SceneDINO}. 
More recently, 3D Gaussian Splatting has emerged as a more efficient alternative to NeRF~\cite{GaussianOcc,chambon2025gaussrender}. 
However, these approaches generally require precise camera information and in-domain training data. 
\cite{GaussianOcc} is a partial exception, avoiding 6D poses via camera overlap, but still requires camera intrinsics and domain-specific information (\ie, adjacent camera overlap). 
Other works~\cite{gausstr,gsocc3d,zhou2025autoocc,sfmocc} focus on pseudo-label generation, using open-vocabulary foundation models~\cite{gausstr,zhou2025autoocc} and sequence-level bundle adjustment~\cite{sfmocc}.
While models trained on these pseudo-labels show promising cross-dataset generalization, they remain limited to specific settings.

\section{Method}
We build \OURS{}, a 3D occupancy framework that can generalize to arbitrary out-of-domain urban scenes.
To this end, we adopt the transformer architecture from the Dust3r family and train the model on multiple urban datasets using standard point-level objectives commonly employed in prior works~\cite{duster,muster,vggt}.
\OURS{} is supervised with metric-scale point-clouds enabling metric predictions at test time, a key element in occupancy prediction.
We propose two novel strategies \emph{Segmentation Forcing} and \emph{Novel View Rendering} to accommodate the unique characteristics of 3D occupancy prediction in urban environments.

\cref{fig:pipeline} illustrates \OURS{} training process, which consists of two stages: \emph{3D Reconstruction} and \emph{Novel View Rendering}.
For each frame sequence, we randomly select $N$ frames for training.
In the reconstruction stage, we set the number of reconstruction frames to $\nrec = N$.
In the rendering stage, we use non-overlapping sets of $\nrec$ reconstruction frames and $\nren$ rendering frames, with $N = \nrec + \nren$.

\subsection{3D Reconstruction with Segmentation Forcing}
\label{sec:seg-forcing}

The \emph{3D Reconstruction} stage aims to recover the scene geometry from a set of reconstruction frames, providing the geometry basis for the subsequent novel-view rendering stage.
In this stage, \OURS{} extends \muster~\cite{muster}, a multi-view geometry network, by adding a SAM2 feature prediction head.
SAM2~\cite{sam2} is a foundation model designed for promptable visual segmentation in images and videos;
its features are thus rich in high-fidelity segmentation cues and are beneficial for resolving geometric ambiguity.
The \emph{Segmentation Forcing} loss compels \OURS{} to predict SAM2-like features.
Our strategy regularizes geometry prediction by leveraging segmentation cues to enforce spatial and temporal feature consistency, thereby improving performance, especially in regions where LiDAR supervision is sparse.

\OURS{} processes $\nrec$ reconstruction frames $\{\imagesIn_i\}_{i=1}^\nrec \in \mathbb{R}^{H\times W \times 3}$ as multi-view inputs to reconstruct the 3D scene.
We feed $\nrec$ frames in chronological order through a shared reconstruction encoder $\reconEncoder$ followed by a shared decoder $\reconDecoder$.
The first frame is always designated as the reference frame; all non-reference frames are identified by a specialized token added at the beginning of the shared decoder.
The two transformers produce, for each frame $\imagesIn_i$:
\begin{itemize}
    \item SAM2-like feature maps  $\features_i \in \mathbb{R}^{H' \times W' \times C}$, 
    \item global pointmaps $\globalpoints_{i,1} \in \mathbb{R}^{H \times W \times 3}$ in the global camera coordinate of the reference frame $1$, 
    \item local pointmaps $\localpoints_{i,i} \in \mathbb{R}^{H \times W \times 3}$ in the local camera coordinate of the current frame $i$,
    \item confidence maps $\conf_i \in \mathbb{R}^{H \times W}$, 
    \item and camera poses $\posesRecon_i \in \mathbb{R}^{7}$ inferred by registering the global and local pointmaps.
\end{itemize}
For each frame $i\in[3,\nrec]$, a scene memory $\sceneMemory_{i-1}$ of all historical reconstruction frames $1..i-1$ is used in the decoding process to infer the geometry of the current frame $i$ via cross-attention between tokens of frame $i$ and memory tokens in $\sceneMemory_{i-1}$.
The scene memory $\sceneMemory_{i}$ is then constructed by concatenating $\sceneMemory_{i-1}$ with the decoder tokens of the current frame $i$.
To initialize, $\sceneMemory_{2}$ is formed by concatenating the decoder tokens of the first two frames.
With a slight abuse of notation, we use $\sceneMemory$ without a subscript to denote the final global scene memory, which aggregates information from the entire sequence; that is, $\sceneMemory \equiv \sceneMemory_{\nrec}\in \mathbb{R}^{H' \times W' \times (C \cdot \nrec)}$.

The decoder is followed by linear heads for pointmap and confidence prediction, and an MLP head for SAM2-like feature prediction.
Because the geometry and segmentation tasks differ in nature, we introduce two learnable task tokens: $\taskG$ for the pointmap heads and $\taskS$ for the SAM2 head.
These tokens are added to all decoder tokens before the corresponding head is applied.
For clarity, we omit task tokens in the equations and only visualize them in~\cref{fig:pipeline}.

The SAM2 head consists of an MLP with two linear layers followed by two upsampling layers.
Each upsampling layer uses bilinear interpolation to resize the features, followed by a convolution, layer norm, and GELU.

In summary, the output of this stage is:
\begin{equation}
\reconDecoder(\reconEncoder(\{\imagesIn_i\}_{i=1}^{\nrec})) = \left(\sceneMemory, \{\features_i, \globalpoints_{i,1}, \localpoints_{i,i}, \conf_i, \posesRecon_i\}_{i=1}^{\nrec}\right).
\label{eqn:rec}
\end{equation}

\subsection{Novel-View Rendering}
\label{sec:novel-semantic-pointmap-synthesis}
We train a rendering encoder $\renderEncoder$ and decoder $\renderDecoder$ to predict pointmaps and SAM2-like features for arbitrary novel views along the reconstruction camera trajectories $\{\posesRecon_i\}_{i=1}^{\nrec}$ (\cf~\cref{eqn:rec}).
The reconstruction modules $\reconEncoder$, $\reconDecoder$ are frozen, and their outputs serve as inputs to the rendering stage.

During training, we sample $\nrec$ reconstruction frames and $\nren$ rendering frames from the same sequence; the first frame always belongs to the reconstruction set.
Let $\{\novelPosesInput_j\}_{j=1}^{\nren}$ be the camera poses of rendering frames $\imagesRender_j$.
Our goal is to render pointmaps and SAM2-like features for each $\novelPosesInput_j$, conditioned on reconstruction outputs.
Rendering frames are used only for loss computation.

\condenseparagraph{Tokenization.} 
We merge the global pointmaps $\{\globalpoints_{i,1}\}_{i=1}^{\nrec}$ into a single point cloud $\globalpoints$ in the reference-frame coordinate system.
Projecting $\globalpoints$ into $\{\novelPosesInput_j\}_{j=1}^\nren$ yields $\nren$ xyz-images and point-to-pixel correspondences, enabling 2D projection of SAM2-like features and confidence maps into each novel view.
Each modality image is processed by an MLP; the results are concatenated and linearly projected to form novel-view tokens $\{\novelInputfeatures_j\}_{j=1}^{\nren}$.
RoPE is used for positional encoding universally.

\condenseparagraph{Rendering.}
Because reconstruction frames cover the scene only partially, projected novel views contain missing areas and projection artifacts.
The rendering transformers learn to complete missing geometry and correct projection errors, producing denser pointmaps.

The rendering encoder $\renderEncoder$ contains $6$ transformer blocks, processing novel-view token representations $\novelInputfeatures$ to predict encoder tokens.
During training, we distill knowledge from the large reconstruction encoder $\reconEncoder$ (24 transformer blocks) to the small rendering encoder $\renderEncoder$ through the $\lossEncfeatures$ loss (defined in~\cref{sec:losses}).
This helps facilitate the optimization process by providing an auxiliary supervision signal, encouraging the rendering encoder to mimic the tokens produced by the larger teacher reconstruction encoder.

The rendering decoder $\renderDecoder$ has the same architecture as the reconstruction decoder $\reconDecoder$ and is initialized from its weights.
We also introduce two learnable task tokens $\rendertaskG$ and $\rendertaskS$, initialized from $\taskG$ and $\taskS$.
The scene memory $\sceneMemory$ obtained from the reconstruction stage remained fixed (\cf~\cref{eqn:rec}) and is used by $\renderDecoder$ to render the final set of outputs.
During decoding, $\renderDecoder$ applies cross-attention between decoder tokens and the memory tokens in $\sceneMemory$, making possible reference to the whole reconstructed scene.
Intuitively, the explicit reconstruction outputs from the previous stage guides the rendering, while the implicit memory provides supporting information to correct and complete the scene.
\emph{Segmentation Forcing} is also applied to regularize novel-view predictions.

In summary, output of the rendering stage is written:
\begin{equation} \renderDecoder(\sceneMemory, \renderEncoder(\{\novelInputfeatures\}_{j=1}^{\nren})) = \{\renderFeatures_j, \renderGlobalpoints_{j,1}, \renderLocalpoints_{j,j}, \renderConf_j\}_{j=1}^{\nren}.
\label{eqn:nvr}
\end{equation}

\subsection{\OURS{} Inference}

\begin{figure}
\centering
\includegraphics[width=\linewidth]{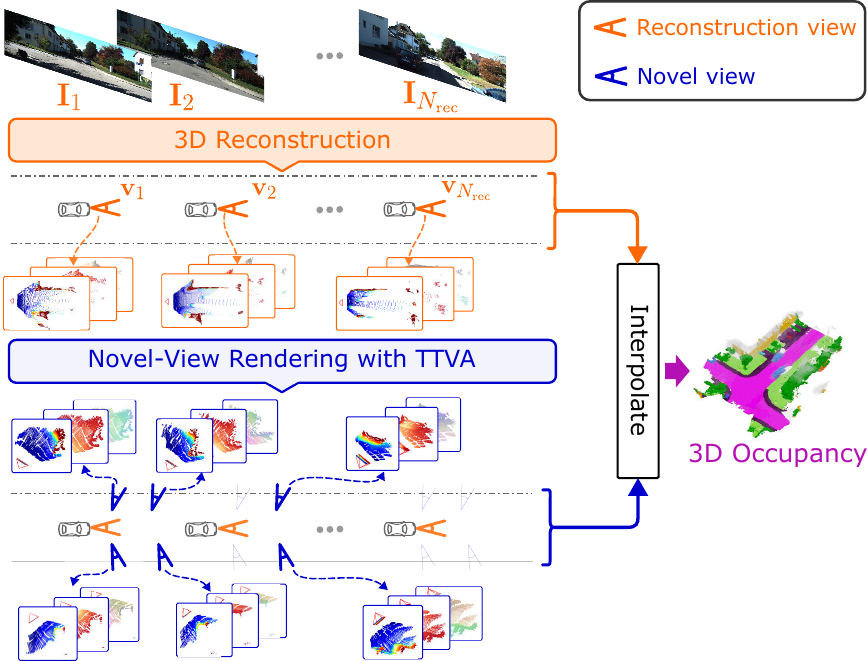}
\caption{\textbf{\OURS{} inference } undergoes two stages: (i) 3D reconstruction to retrieve $\nrec$ pointmaps with predicted camera poses $\{\posesRecon_i\}_{i=1}^{\nrec}$, and (ii) novel-view rendering with TTVA sampled along the trajectory of $\{\posesRecon_i\}_{i=1}^{\nrec}$. 3D occupancy is obtained by aggregating all pointmaps and voxelizing them with trilinear interpolation.}
\vspace{-0.3cm}
\label{fig:inference}
\vspace{-0.3cm}
\end{figure}

We first retrieve the reconstructed pointmaps, SAM2-like features and the registered camera poses of all $\nrec$ input frames from the \emph{3D Reconstruction} stage.
We then randomly sample novel views around the predicted camera trajectory $\{\posesRecon_i\}_{i=1}^{\nrec}$ (\cf~\cref{eqn:rec}) and pass them through the \emph{Novel View Rendering} (NVR) stage to infer the novel-view pointmaps and segmentation features (\cf~\cref{eqn:nvr}).
The final 3D occupancy is obtained by aggregating all pointmaps from both stages and voxelizing them into a dense grid via trilinear interpolation. The inference protocol is visualized in~\cref{fig:inference}. \OURS{} is versatile and can predict 3D occupancy for either sequential, monocular, or surround-view inputs.
Predicted SAM2-like features can be directly used for segmentation.

\condenseparagraph{NVR Inference.}
Thanks to NVR, we can use arbitrary views at test-time to help infer occlusion; this strategy is coined Test-time View Augmentation (TTVA).
We first position novel camera views uniformly every $\rho_{\rm fwd}$ meters along a straight path along the trajectory of predicted poses $\{\posesRecon_i\}_{i=1}^{\nrec}$.
At each of those $N_{\rm fwd}$ sampled positions, we vary horizontal viewing angles $\Phi{=}\{0,\pm\phi\}$ and shift the camera by a lateral amount of $\pm \rho_{\rm lat}$.
~\cref{fig:ablation_ttva} illustrates the NVR setups.

\condenseparagraph{Segmentation w/ SAM2-like features.}
We apply Grounded SAM2~\cite{gsam} pipeline by feeding the first frame to GroundingDINO~\cite{gdino} and obtain candidate bounding boxes of all semantic classes of interest.
We then use the pretrained prompt decoder of SAM2 to prompt \OURS{}'s predicted SAM2-like features with the obtained bounding boxes, resulting in dense semantic masks for the first frame.
Semantic masks are then propagated through the entire scene with SAM2 video tracking.
Finally we assign the predicted occupancy voxels with predicted semantic classes.

\subsection{Training Losses}

Both stages are trained using the same set of losses, \ie global- and local- pointmap loss $\lossGlobalpoints$, $\lossLocalpoints$, and Segmentation Forcing loss $\lossForcing$, with the exception of the rendering encoder distillation loss $\lossEncfeatures$, which is applied only in the rendering stage.
We only describe common losses in the reconstruction stage for brevity.

\label{sec:losses}

\condenseparagraph{Pointmap Losses $\lossGlobalpoints$, $\lossLocalpoints$.} 
The loss weights the difference between the predicted pointmap $\globalpoints_{i,1}$ and ground truth $\globalpointsgt_{i,1}$ using the predicted confidence map $\conf_i$~\cite{muster}:
\begin{equation}
\lossGlobalpoints = \frac{1}{|s|}\sum_{i=1}^{\nrec}
\Big\lVert \conf_i \odot \big(\globalpoints_{i,1} - \globalpointsgt_{i,1}\big) \Big\rVert_{1}
\;-\; \alpha \,\log\!\big(\conf_i\big), \notag \nonumber
\end{equation}
where $\odot$ denotes element-wise multiplication with channel-wise broadcasting, and $\alpha$ controls the regularization strength, and $s$ is the normalization scale~\cite{muster,cuter}.
The local pointmap loss $\lossLocalpoints$ is formulated identically.

\condenseparagraph{Geometry-aware Segmentation Forcing Loss $\lossForcing$.}
We employ a Mean Squared Error (MSE) loss. 
We use the same confidence map $\conf$ in pointmap losses above to weight the MSE error:
\begin{equation}
\lossForcing = \frac{1}{H^{'}W^{'}}\sum_{i=1}^{\nrec}  \Big\lVert \conf_i \odot  (\features_{i} - \featuresgt_{i}) \Big\rVert^2_2, 
\label{eqn:forcing_loss}
\end{equation}
where $\nrec$ is the number of reconstruction frames.
Since $\conf$ represents the geometry confidence learned by the pointmap head, our weighting forces the network to focus on high-confidence areas and ignore low-confidence ones like sky.
We note that $\lossForcing$ does not update the confidence head.

\condenseparagraph{Encoder Distillation Loss $\lossEncfeatures$.} This loss distills knowledge from the larger teacher reconstruction encoder $\reconEncoder$ (24 layers) to the smaller student rendering encoder $\renderEncoder$ (6 layers).
It minimizes the squared L2 distance between the output tokens from both encoders.
Given the output tokens from the rendering encoder $\renderEncoder(\{\novelInputfeatures_j\}_{j=1}^{\nren})$ and the reconstruction encoder $\reconEncoder(\{\imagesRender_j\}_{j=1}^{\nren})$, the loss is written as:
\begin{equation}
\lossEncfeatures = \sum_{j=1}^{\nrec} \Big\lVert \reconEncoder(\imagesRender_{j}) - \renderEncoder(\novelInputfeatures_j)\Big\rVert_2^2,  \notag \nonumber
\end{equation}
where $\{\imagesRender_j\}_{j=1}^{\nrec}$ are the novel-view images.

\begin{figure*}[ht!]
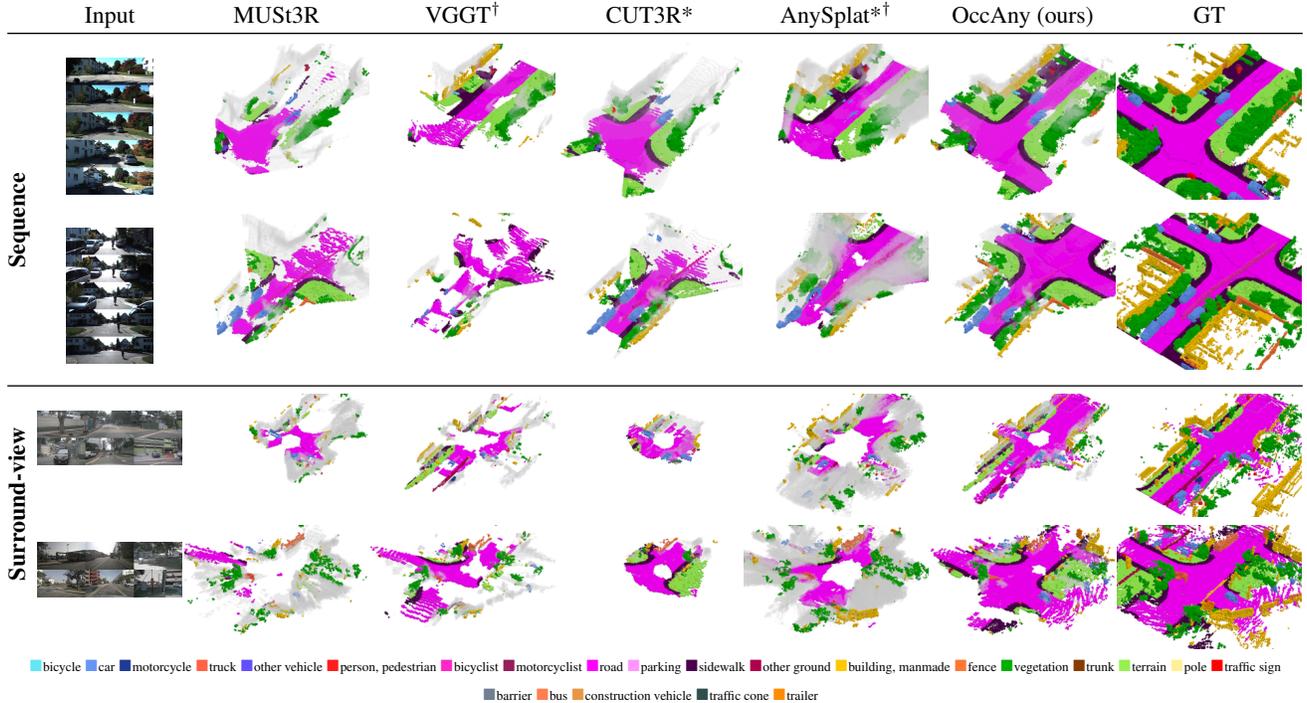

	\centering
	\small
	\setlength{\tabcolsep}{0.5pt}
	\renewcommand{\arraystretch}{0}
	\newcommand{\imgsizer}{.14\textwidth}
	\newcommand{\trimWidth}{0.45}
	\newcommand{\trimHeight}{0.42}
	\newcommand{\trimLeft}{0.1}
	\newcommand{\trimBot}{0.05}
	\newcommand{\kittiOccRow}[2]{%
        \makecell[t]{\adjincludegraphics[width=.6\linewidth]{figs/kitti_5frames_#2opa/#1_stacked_recon.png}} &
        \makecell{\adjincludegraphics[width=\linewidth,trim={\trimLeft\width} {\trimBot\height} {\trimWidth\width} {\trimHeight\height}, clip]{figs/kitti_5frames_#2opa/#1_must3r.png}} &
        \makecell{\adjincludegraphics[width=\linewidth,trim={\trimLeft\width} {\trimBot\height} {\trimWidth\width} {\trimHeight\height}, clip]{figs/kitti_5frames_#2opa/#1_VGGT.png}} &
        \makecell{\adjincludegraphics[width=\linewidth,trim={\trimLeft\width} {\trimBot\height} {\trimWidth\width} {\trimHeight\height}, clip]{figs/kitti_5frames_#2opa/#1_cut3r.png}} &
        \makecell{\adjincludegraphics[width=\linewidth,trim={\trimLeft\width} {\trimBot\height} {\trimWidth\width} {\trimHeight\height}, clip]{figs/kitti_5frames_#2opa/#1_AnySplat.png}} &
        \makecell{\adjincludegraphics[width=\linewidth,trim={\trimLeft\width} {\trimBot\height} {\trimWidth\width} {\trimHeight\height}, clip]{figs/kitti_5frames_#2opa/#1_occany512_original.png}} &
        \makecell{\adjincludegraphics[width=\linewidth,trim={\trimLeft\width} {\trimBot\height} {\trimWidth\width} {\trimHeight\height}, clip]{figs/kitti_5frames_#2opa/#1_ground_truth.png}}%
    }

	\newcommand{\trimNuScenesWidth}{0.35}
	\newcommand{\trimNuScenesHeight}{0.34}
	\newcommand{\trimNuScenesLeft}{0.22}
	\newcommand{\trimNuScenesBot}{0.26}

    \newcommand{\occRow}[2]{%
        \makecell{\adjincludegraphics[width=\linewidth]{figs/#2/#1_stacked_recon.png}} &
        \makecell{\adjincludegraphics[width=\linewidth,trim={\trimNuScenesLeft\width} {\trimNuScenesBot\height} {\trimNuScenesWidth\width} {\trimNuScenesHeight\height}, clip]{figs/#2/#1_must3r.png}} &
        \makecell{\adjincludegraphics[width=\linewidth,trim={\trimNuScenesLeft\width} {\trimNuScenesBot\height} {\trimNuScenesWidth\width} {\trimNuScenesHeight\height}, clip]{figs/#2/#1_VGGT.png}} &
        \makecell{\adjincludegraphics[width=\linewidth,trim={\trimNuScenesLeft\width} {\trimNuScenesBot\height} {\trimNuScenesWidth\width} {\trimNuScenesHeight\height}, clip]{figs/#2/#1_cut3r.png}} &
        \makecell{\adjincludegraphics[width=\linewidth,trim={\trimNuScenesLeft\width} {\trimNuScenesBot\height} {\trimNuScenesWidth\width} {\trimNuScenesHeight\height}, clip]{figs/#2/#1_AnySplat.png}} &
        \makecell{\adjincludegraphics[width=\linewidth,trim={\trimNuScenesLeft\width} {\trimNuScenesBot\height} {\trimNuScenesWidth\width} {\trimNuScenesHeight\height}, clip]{figs/#2/#1_occany512_original.png}} &
        \makecell{\adjincludegraphics[width=\linewidth,trim={\trimNuScenesLeft\width} {\trimNuScenesBot\height} {\trimNuScenesWidth\width} {\trimNuScenesHeight\height}, clip]{figs/#2/#1_ground_truth.png}}
    }
		\begin{tabular}{m{0.02\textwidth} m{0.11\textwidth}m{\imgsizer}m{\imgsizer}m{\imgsizer}m{\imgsizer}m{\imgsizer}m{\imgsizer}}
			& \multicolumn{1}{c}{Input\vphantom{$^\dagger$}} & \multicolumn{1}{c}{\muster{}\vphantom{$^\dagger$}} & \multicolumn{1}{c}{VGGT$^\dagger$} & \multicolumn{1}{c}{\cuter{}*\vphantom{$^\dagger$}} & \multicolumn{1}{c}{{AnySplat*}$^\dagger$} & \multicolumn{1}{c}{\OURS{} (ours)\vphantom{$^\dagger$}} & \multicolumn{1}{c}{GT\vphantom{$^\dagger$}} \\
			\midrule
			\multirow{2}{*}{\rotatebox[origin=c]{90}{\parbox{2.5cm}{\centering\textbf{\Sequence{}}}}} & \kittiOccRow{08_001390}{3} \\ 
			 & \kittiOccRow{08_001245}{3} \\ 
             \midrule
			\multirow{2}{*}{\rotatebox[origin=c]{90}{\parbox{2.cm}{\centering\textbf{\Surround{}}}}} & \occRow{scene-0276_46a4a199e44544018017be7434e46fc8}{nuscenes_surround_10opa}\\ 
			 & \occRow{scene-0552_5ef15615bc08428bab23b2a84e84961e}{nuscenes_surround_10opa}\\
		\end{tabular}
		{				
			\tiny
			\textcolor{bicycle}{$\blacksquare$}bicycle~%
			\textcolor{car}{$\blacksquare$}car~%
			\textcolor{motorcycle}{$\blacksquare$}motorcycle~%
			\textcolor{truck}{$\blacksquare$}truck~%
			\textcolor{other-vehicle}{$\blacksquare$}other vehicle~%
			\textcolor{person}{$\blacksquare$}person, pedestrian~%
			\textcolor{bicyclist}{$\blacksquare$}bicyclist~%
			\textcolor{motorcyclist}{$\blacksquare$}motorcyclist~%
			\textcolor{road}{$\blacksquare$}road~%
			\textcolor{parking}{$\blacksquare$}parking~%
			\textcolor{sidewalk}{$\blacksquare$}sidewalk~%
			\textcolor{other-ground}{$\blacksquare$}other ground~%
			\textcolor{building}{$\blacksquare$}building, manmade~%
			\textcolor{fence}{$\blacksquare$}fence~%
			\textcolor{vegetation}{$\blacksquare$}vegetation~%
			\textcolor{trunk}{$\blacksquare$}trunk~%
			\textcolor{terrain}{$\blacksquare$}terrain~%
			\textcolor{pole}{$\blacksquare$}pole~%
			\textcolor{traffic-sign}{$\blacksquare$}traffic sign~%
		 	
		 	\textcolor{barrier}{$\blacksquare$}barrier
		 	\textcolor{bus}{$\blacksquare$}bus
		 	\textcolor{construction-vehicle}{$\blacksquare$}construction vehicle
		 	\textcolor{traffic-cone}{$\blacksquare$}traffic cone
		 	\textcolor{trailer}{$\blacksquare$}trailer
		 }
    \vspace{-0.2cm}
	\caption{\textbf{Occupancy predictions} of \OURS{} and baselines on a sequence and a surround view. We visualize here predicted voxels. For qualitative analysis, we overlay the semantic ground-truth colors on predicted voxels to better highlight class-wise gains. False positive voxels are painted in gray without any overlayed color. Compared to baselines, our occupancy predictions are denser and more accurate.}
	\vspace{-0.2cm}
	\label{fig:qual_res}
\end{figure*}

\section{Experiments}
\condenseparagraph{Training.}
\OURS{} is trained on a mixture of five urban datasets, using images from all cameras and projected LiDAR pointmap as ground truth: Waymo~\cite{waymo}, DDAD~\cite{ddad}, PandaSet~\cite{pandaset}, VKITTI2~\cite{vkitti2}, and ONCE~\cite{once}.

In the reconstruction stage, we initialize with \muster{}~\cite{muster}, freeze the encoder $\reconEncoder{}$ and only train the decoder $\reconDecoder{}$ for 3D reconstruction.
Input frames are resized to $512$-width with varying aspect ratios.
We sample training sequences with minimum length $N{=}6$ and maximum length
$N{=}10$. Frames are sampled at 2Hz in all datasets.

In the rendering stage, we initialize $\renderDecoder{}$ with the pretrained weights of $\reconDecoder{}$.
We keep the same sequence length $N\in[6,10]$, and randomly select among those $\nren$ frames as rendering views; the remaining $\nrec = N-\nren$ are used for reconstruction.
The first frame serves as reference and it is always part of the reconstruction set.

\condenseparagraph{Evaluation.}
We evaluate the generalization of \OURS{} on two out-of-domain benchmarks: \semantickitti~\cite{semkitti} and \occDnuscenes~\cite{occ3d}, detailed in~\cref{app:tech_details}.

We use three evaluation settings:
\begin{itemize}
    \item \emph{\Sequence:} a sequence of $5$ frames coming from a single camera on \semantickitti{} and \occDnuscenes{},
    \item \emph{\Singleview:} a single input frame on \semantickitti{},
    \item \emph{\Surround:} all surrounding frames at a single timestep on \occDnuscenes{}.
\end{itemize}

\condenseparagraph{NVR inference.} In the \emph{Sequence} and \emph{Surround-view} settings, we use the augmentation strategy TTVA with $N_{\rm fwd}=10$, forward shift $\rho_{\rm fwd}$ of 3\,m, and lateral shift $\rho_{\rm lat}$ of 2\,m.
In the \emph{\Singleview} setting, we sample denser and use $N_{\rm fwd}=50$, forward shift $\rho_{\rm fwd}$ of 1\,m, lateral shift $\rho_{\rm lat}$ of 2\,m. All settings use horizontal angle $\phi$ of $\{0^\circ, \pm 60^\circ\}$.

\condenseparagraph{Baselines.} We compare \OURS{} against four strong baselines: \muster~\cite{muster}, \cuter~\cite{cuter}, VGGT~\cite{vggt}, AnySplat~\cite{anysplat}, and Depth Anything 3 (DA3)~\cite{depthanything3}. Among them, \cuter{} is trained only in the online setting. AnySplat is an VGGT extension with Gaussian Splatting~\cite{gsplat} for novel view synthesis and for improving geometric consistency.
\muster{} and \cuter{} output metric-scale pointmaps, whereas VGGT and AnySplat produce scale-invariant pointmaps.
To resolve the scale ambiguity of VGGT and AnySplat, we calibrate their depth predictions with Metric3Dv2~\cite{metric3d} using their predicted camera intrinsics; those two variants are presented as VGGT$^{\dagger}$ and AnySplat$^{\dagger}$. For DA3, we use DA3-LARGE to estimate global point map and DA3METRIC-LARGE for metric scaling.
Since AnySplat and~\cuter{} support novel-view synthesis, we also apply our proposed \ttva{} strategy to improve those, referred to as \cuter{}* and AnySplat*$^{\dagger}$. 
All models are tested on the same input resolution, with a very slight difference depending on the patch-size.

For reference, we also report published results from vision-based self-supervised occupancy models trained \emph{in-domain}, which are heavily biased to dataset-specific characteristics especially camera intrinsics and extrinsics.
We compare against self-supervised methods as both do not require in-domain 3D ground-truth for training.
However, \OURS{} is completely zero-shot while self-supervised methods are trained on in-domain calibrated data. 

\begin{table}[t]
	\scriptsize
	\setlength{\tabcolsep}{0.0055\linewidth}
	\centering
	\resizebox{\columnwidth}{!}{
		\begin{tabular}{l|c|c|cc >{\columncolor{important}}c|c|cc >{\columncolor{important}}c}
			\toprule
			\multirow{2}{*}{\textbf{Method}} & \multirow{2}{*}{\textbf{Venue}} & \multicolumn{4}{c|}{\textbf{Semantic KITTI}} & \multicolumn{4}{c}{\textbf{Occ3D-NuScenes}} \\
			\cmidrule(lr){3-6} \cmidrule(lr){7-10} 
			&& \textbf{Res.} & \textbf{Prec.} & \textbf{Rec.} & \textbf{IoU} & \textbf{Res.} & \textbf{Prec.} & \textbf{Rec.} & \textbf{IoU}  \\
			\midrule
			\muster~\cite{muster} & CVPR'25 & 512x160 & 18.38 & 25.58 & 11.97 & 512x288 & 19.27 & 28.60 & 13.01 \\
			\cuter~\cite{cuter} & CVPR'25 & 512x160  & 25.72 & 21.11 & 13.11 & 512x288 & 24.69 & 16.57 & 11.01\\
			\cuter*~\cite{cuter} & CVPR'25 & 512x160  &  27.05 & 27.92 & 15.93 & 512x288 & 29.44 & 30.50 & 17.62 \\
			VGGT$^\dagger$~\cite{vggt} & CVPR'25 & 518x168 &  36.35 & 22.62 & 15.20 & 518x294 & {38.34} & 26.23 & {18.45} \\
			AnySplat$^\dagger$~\cite{anysplat} & TOG'25 & 518x168 & 18.22 & 35.62 & 11.67 & 518x294 & 26.67 & 36.93 & 18.33 \\
			{AnySplat*}$^\dagger$~\cite{anysplat} &  TOG'25 & 518x168 & 14.53 & \best{47.48} & 12.39 & 518x294 &   24.42 & \best{42.48} & 18.35 \\
			DA3~\cite{depthanything3} &  ICLR'26 & 518x168 &   26.37 & 28.13 & 15.76 &  518x294 &  \best{51.25} & 23.64 & \second{19.30}   \\
			\OURS{}$_\text{base}$ & -- & 512x160 & \best{43.38} & 20.37 & \second{16.09}   & 512x288 & \second{48.09} & 20.97 & 17.10\\
			\textbf{\OURS{}} & -- & 512x160 &  \second{36.79} & \second{46.70} & \best{25.91}   & 512x288 &  36.09 & \second{40.39} & \best{23.55}  \\
			\bottomrule
            \multicolumn{7}{l}{\tiny{*: use \ttva{}\,\,\,\,\,$^\dagger$: scaled with Metric3Dv2~\cite{metric3d}.}}\\
			\multicolumn{7}{l}{\tiny{\OURS{}$_\text{base}$: w/o Segmentation Forcing \& Novel-view Rendering.}}
		\end{tabular}
	}
    \vspace{-0.2cm}
	\caption{\textbf{Sequence setting.} Occupancy prediction on SemanticKITTI and Occ3D-NuScenes.}
	\label{tbl:sequence_res}
\end{table}  

\begin{table}[t]
	\centering
	\setlength{\tabcolsep}{0.0055\linewidth}
	\scriptsize
	\resizebox{0.9\columnwidth}{!}{%
		\begin{tabular}{c|l|c|c|cc >{\columncolor{important}}c}
			\toprule
			\textbf{Test} & \textbf{Method} & \textbf{Venue}  & \textbf{Res.} & \textbf{Prec.} & \textbf{Rec.} & \textbf{IoU} \\
			\midrule
			\multirow{6}{*}{\rotatebox{90}{\indomain{\textbf{in-domain}}}} & \indomain{MonoScene~\cite{monoscene} }  & \indomain{CVPR'22} &  \indomain{1220x370} & \indomain{13.15} & \indomain{40.22} & \indomain{11.18} \\
			&\indomain{SceneRF~\cite{scenerf}} & \indomain{ICCV'23}  & \indomain{1220x370} & \indomain{17.28} & \indomain{40.96} & \indomain{13.84} \\
			& \indomain{SelfOcc}~\cite{selfocc} & \indomain{CVPR'24}  & \indomain{1220x370} & \indomain{34.83} & \indomain{37.31} & \indomain{21.97} \\
			& \indomain{Splatter Image~\cite{splatterimage}} & \indomain{CVPR'24}  & \indomain{1220x370} & \indomain{11.30} & \indomain{53.93} & \indomain{10.30}    \\
			& \indomain{Hi-Gaussian}~\cite{higaussian}  & \indomain{ICCV'25}  & \indomain{1220x370} & \indomain{17.39} & \indomain{59.72} & \indomain{15.56} \\
			& \indomain{OccNeRF~\cite{occnerf}} & \indomain{TIP'25}  & \indomain{1220x370} & \indomain{35.25} & \indomain{39.27} & \indomain{22.81} \\
			\midrule
			\multirow{8}{*}{\rotatebox{90}{\textbf{out-of-domain}}} & \muster~\cite{muster} & CVPR'25 & 512x160 &   15.29 & 12.24 & 7.29\\
			& \cuter~\cite{cuter} & CVPR'25 &  512x160 & 33.32 & 8.64 & 7.37 \\
			& \cuter{}*~\cite{cuter} & CVPR'25 & 512x160 &  33.47  & 17.59 & \second{13.03} \\
			& VGGT$^\dagger$~\cite{vggt} & CVPR'25 & 518x168 & 25.59 & 14.49 & 10.19 \\
			& AnySplat$^\dagger$~\cite{anysplat} & TOG'25 & 518x168 &  17.97 & 20.39 & 10.56 \\
			& {AnySplat*}$^\dagger$~\cite{anysplat} & TOG'25 &  518x168 & 14.61 & \best{35.21} & 11.52\\
            & DA3~\cite{depthanything3} &  ICLR'26 & 518x168 & 23.98 & 14.54 & 9.95 \\
			& \OURS{}$_\text{base}$ & -- &  512x160 & \second{41.24} & 14.49 & 12.01 \\
			& \textbf{\OURS{}} & -- &   512x160 & \best{45.64} & \second{33.66} & \best{24.03} \\
			\bottomrule
			\multicolumn{7}{l}{\tiny{*: use \ttva{}\,\,\,\,\,$^\dagger$: scaled with Metric3Dv2~\cite{metric3d}.}}\\
			\multicolumn{7}{l}{\tiny{\OURS{}$_\text{base}$: w/o Segmentation Forcing \& Novel-view Rendering.}}
		\end{tabular}
	}
	\vspace{-0.2cm}
	\caption{\textbf{\Singleview{} setting.} Occupancy results with \Singleview{} input on \semantickitti{} following \cite{selfocc,scenerf}. Results for MonoScene and Splatter Image are taken from \cite{scenerf, higaussian}.}
	\label{tbl:monocular_skitti}
\end{table}

\begin{table}[t]
	\scriptsize
	\centering
	\resizebox{0.95\columnwidth}{!}{
		\setlength{\tabcolsep}{0.0085\linewidth}
		\begin{tabular}{c|l|c|c|cc >{\columncolor{important}}c}
			\toprule
			\textbf{Test} & \textbf{Method} & \textbf{Venue} &  \textbf{Res.} &\textbf{Prec.} & \textbf{Rec.} & \textbf{IoU} \\
			\midrule
			\multirow{5}{*}{\rotatebox{90}{\indomain{\textbf{in-domain}}}} & \indomain{SelfOcc~\cite{selfocc}} & \indomain{CVPR'24} & \indomain{800x450} & \indomain{--} & \indomain{--} & \indomain{45.01} \\
			&\indomain{OccNeRF~\cite{occnerf}} & \indomain{TIP'25} & \indomain{672x336} & \indomain{57.20} & \indomain{55.47} & \indomain{39.20} \\
			&\indomain{DistillNeRF~\cite{distillnerf}} & \indomain{NeuRIPS'24} & \indomain{400×228} & \indomain{--} & \indomain{--} & \indomain{29.11} \\
			&\indomain{SimpleOcc~\cite{simpleocc}} & \indomain{TIV'24} &  \indomain{672x336} & \indomain{41.91} & \indomain{64.02} & \indomain{33.92} \\
			&\indomain{GaussTR~\cite{gausstr}} & \indomain{CVPR'25} & \indomain{896x504} & \indomain{--} & \indomain{--} & \indomain{45.19} \\
			\midrule
			\multirow{8}{*}{\rotatebox{90}{\textbf{out-of-domain}}} & \muster~\cite{muster} & CVPR'25 &  512x288 & 20.79 & 28.29 & 13.61 \\
			& \cuter~\cite{cuter} & CVPR'25 &  512x288 & 32.19 & 7.93 & 6.79 \\
			&\cuter{}*~\cite{cuter} & CVPR'25 &  512x288 & 40.60 & 26.73 & 19.21 \\
			& VGGT$^\dagger$~\cite{vggt}& CVPR'25 &  518x294 & 41.56 & 28.64 & 20.42\\
			& AnySplat$^\dagger$~\cite{anysplat} & TOG'25 &   518x294 & 29.35 & 40.80 & 20.59\\
			& {AnySplat*}$^\dagger$~\cite{anysplat} & TOG'25 &  518x294 &  24.52 & \second{57.65} & \second{20.78} \\
            & DA3~\cite{depthanything3} & ICLR'26 &  518x294 &  \second{53.26} & 23.75 & 19.65 \\
			& \OURS{}$_\text{base}$ & -- & 512x288 & \best{59.58} & 21.19 & 18.53  \\
			& \textbf{\OURS{}} & -- & 512x288 & {45.04} & \best{58.54} & \best{34.15}  \\
			\bottomrule
			\multicolumn{7}{l}{\tiny{*: use \ttva{}\,\,\,\,\,$^\dagger$: scaled with Metric3Dv2~\cite{metric3d}.}}\\
			\multicolumn{7}{l}{\tiny{\OURS{}$_\text{base}$: w/o Segmentation Forcing \& Novel-view Rendering.}}
			
	\end{tabular}}
	\\ 
	\vspace{-0.2cm}
	\caption{\textbf{Surround-view setting}. More results are in~\cref{tab:additional_method_comparison}.}
    \vspace{-0.2cm}
	\label{tab:surround-occ3d-nuscenes}
\end{table}

\condenseparagraph{Metrics.} Similar to~\cite{monoscene, selfocc}, we use the standard 3D occupancy metrics Precision, Recall, and Intersection over Union (IoU) to assess geometry quality; mean IoU (mIoU) is used for semantic segmentation.
Following open-vocabulary LiDAR semantic segmentation works~\cite{sal, leap, losc}, we also report performance on super classes, denoted as mIoU$^{\rm sc}$.
This helps evaluate results at a coarser semantic level, alleviating the impact of ``prompting and text-to-image alignment'' limitations~\cite{sal} especially on semantically confusing classes, \eg, ``car'' \vs ``other-vehicle''.

\subsection{Main results}

\begin{figure*}[ht!]
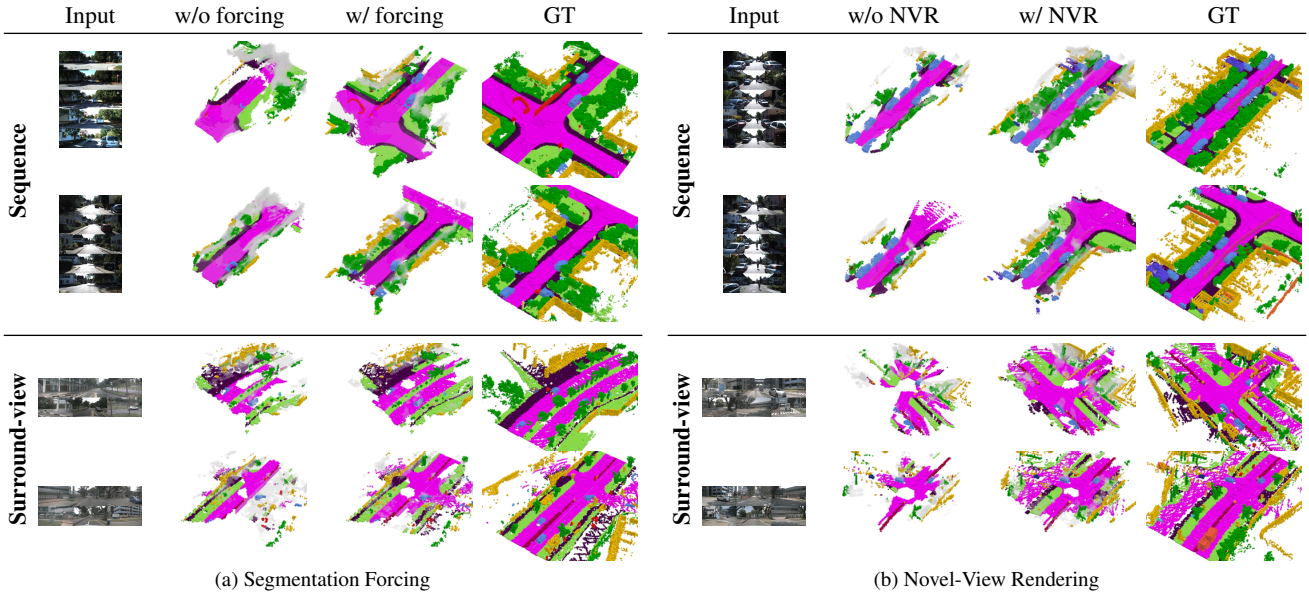

	\centering
	\small
	\setlength{\tabcolsep}{0pt}
	\renewcommand{\arraystretch}{0}
	\newcommand{\imgsizer}{.24\textwidth}
	\newcommand{\trimWidth}{0.45}
	\newcommand{\trimHeight}{0.4}
	\newcommand{\trimLeft}{0.1}
	\newcommand{\trimBot}{0.05}
	\newcommand{\kittiOccRow}[2]{%
        \makecell{\adjincludegraphics[width=0.6\linewidth]{figs/kitti_5frames_#2opa/#1_stacked_recon.png}} &
        \makecell{\adjincludegraphics[width=\linewidth,trim={\trimLeft\width} {\trimBot\height} {\trimWidth\width} {\trimHeight\height}, clip]{figs/kitti_5frames_#2opa/#1_occany512_noDistill.png}} &
        \makecell{\adjincludegraphics[width=\linewidth,trim={\trimLeft\width} {\trimBot\height} {\trimWidth\width} {\trimHeight\height}, clip]{figs/kitti_5frames_#2opa/#1_occany512_original.png}} &
        \makecell{\adjincludegraphics[width=\linewidth,trim={\trimLeft\width} {\trimBot\height} {\trimWidth\width} {\trimHeight\height}, clip]{figs/kitti_5frames_#2opa/#1_ground_truth.png}}
    }
    
	\newcommand{\kittiOccRowNoTTVA}[2]{%
        \makecell{\adjincludegraphics[width=0.6\linewidth]{figs/kitti_5frames_#2opa/#1_stacked_recon.png}} &
        \makecell{\adjincludegraphics[width=\linewidth,trim={\trimLeft\width} {\trimBot\height} {\trimWidth\width} {\trimHeight\height}, clip]{figs/kitti_5frames_#2opa/#1_occany512_noTTVA.png}} &
        \makecell{\adjincludegraphics[width=\linewidth,trim={\trimLeft\width} {\trimBot\height} {\trimWidth\width} {\trimHeight\height}, clip]{figs/kitti_5frames_#2opa/#1_occany512_original.png}} &
        \makecell{\adjincludegraphics[width=\linewidth,trim={\trimLeft\width} {\trimBot\height} {\trimWidth\width} {\trimHeight\height}, clip]{figs/kitti_5frames_#2opa/#1_ground_truth.png}}
    }

	\newcommand{\trimNuScenesWidth}{0.35}
	\newcommand{\trimNuScenesHeight}{0.33}
	\newcommand{\trimNuScenesLeft}{0.22}
	\newcommand{\trimNuScenesBot}{0.25}
	\newcommand{\occRow}[2]{%
		\adjincludegraphics[width=\linewidth]{figs/#2/#1_stacked_recon.png} &
		\adjincludegraphics[width=\linewidth,trim={\trimNuScenesLeft\width} {\trimNuScenesBot\height} {\trimNuScenesWidth\width} {\trimNuScenesHeight\height}, clip]{figs/#2/#1_occany512_noDistill.png} &
		\adjincludegraphics[width=\linewidth,trim={\trimNuScenesLeft\width} {\trimNuScenesBot\height} {\trimNuScenesWidth\width} {\trimNuScenesHeight\height}, clip]{figs/#2/#1_occany512_original.png} &
		\adjincludegraphics[width=\linewidth,trim={\trimNuScenesLeft\width} {\trimNuScenesBot\height} {\trimNuScenesWidth\width} {\trimNuScenesHeight\height}, clip]{figs/#2/#1_ground_truth.png} 
	}

	\newcommand{\occRowNoTTVA}[2]{%
		\adjincludegraphics[width=\linewidth]{figs/#2/#1_stacked_recon.png} &
		\adjincludegraphics[width=\linewidth,trim={\trimNuScenesLeft\width} {\trimNuScenesBot\height} {\trimNuScenesWidth\width} {\trimNuScenesHeight\height}, clip]{figs/#2/#1_occany512_noTTVA.png} &
		\adjincludegraphics[width=\linewidth,trim={\trimNuScenesLeft\width} {\trimNuScenesBot\height} {\trimNuScenesWidth\width} {\trimNuScenesHeight\height}, clip]{figs/#2/#1_occany512_original.png} &
		\adjincludegraphics[width=\linewidth,trim={\trimNuScenesLeft\width} {\trimNuScenesBot\height} {\trimNuScenesWidth\width} {\trimNuScenesHeight\height}, clip]{figs/#2/#1_ground_truth.png} 
	}
	
	\begin{subfigure}{0.49\textwidth}
		\centering
        \setlength{\tabcolsep}{2pt}
		\begin{tabular}{m{0.03\textwidth} m{0.16\textwidth}m{\imgsizer}m{\imgsizer}m{\imgsizer}}
			& \multicolumn{1}{c}{Input\vphantom{$^\dagger$}} & \multicolumn{1}{c}{w/o forcing} & \multicolumn{1}{c}{w/ forcing\vphantom{$^\dagger$}} & \multicolumn{1}{c}{GT\vphantom{$^\dagger$}} \\
			\midrule
			\multirow{2}{*}{\rotatebox[origin=c]{90}{\parbox{2.4cm}{\centering\textbf{\Sequence{}}}}} & \kittiOccRow{08_001475}{3} \\
            & \kittiOccRow{08_000645}{3} \\
            \midrule
			\multirow{2}{*}{\rotatebox[origin=c]{90}{\parbox{1.7cm}{\centering\textbf{\Surround{}}}}} & \occRow{scene-0273_c0ac3dc491e7414184f44c2b32d24ad8}{nuscenes_surround_10opa} \\
            & \occRow{scene-0277_aae0304d89854d818cb48c0aa52a4ebd}{nuscenes_surround_10opa} \\
		\end{tabular}
		\caption{Segmentation Forcing}
		\label{fig:qualitative-occupancy-a}
	\end{subfigure}
	\hfill
	\begin{subfigure}{0.49\textwidth}
		\centering
        \setlength{\tabcolsep}{2pt}
		\begin{tabular}{m{0.03\textwidth} m{0.16\textwidth}m{\imgsizer}m{\imgsizer}m{\imgsizer}}
			 & \multicolumn{1}{c}{Input\vphantom{$^\dagger$}} & \multicolumn{1}{c}{w/o NVR} & \multicolumn{1}{c}{w/ NVR\vphantom{$^\dagger$}} & \multicolumn{1}{c}{GT\vphantom{$^\dagger$}} \\
			\midrule
            \multirow{2}{*}{\rotatebox[origin=c]{90}{\parbox{2.4cm}{\centering\textbf{\Sequence{}}}}}& \kittiOccRowNoTTVA{08_001185}{3} \\
			& \kittiOccRowNoTTVA{08_001230}{3} \\
            \midrule
			\multirow{2}{*}{\rotatebox[origin=c]{90}{\parbox{1.7cm}{\centering\textbf{\Surround{}}}}} & \occRowNoTTVA{scene-0968_22cc5dafc805425b9bcec94512093825}{nuscenes_surround_10opa} \\
            &\occRowNoTTVA{scene-0035_de550f4c35284f90844c07a64253639b}{nuscenes_surround_10opa} \\
		\end{tabular}
		\caption{Novel-View Rendering}
		\label{fig:qualitative-occupancy-b}
	\end{subfigure}
	\vspace{-0.2cm}
	\caption{\textbf{Qualitative ablation} shows the gains from \emph{Segmentation Forcing} and \emph{Novel-View Rendering}. Voxel colorization follows~\cref{fig:qual_res}. The two proposed strategies significantly improve the density and the accuracy of occupancy predictions.}
    \vspace{-0.2cm}
	\label{fig:qualitative-occupancy}
\end{figure*}

\condenseparagraph{\Sequence{}.}
In the \emph{Sequence} setting (~\cref{tbl:sequence_res}), \OURS{} surpasses all other zero-shot baselines.
On \semantickitti{}, it reaches $25.91\%$ IoU, surpassing the nearest baseline (\cuter*) by roughly $10$ points.
A similar trend is observed on Occ3D-NuScenes, where \OURS{} achieves $23.55\%$ IoU, significantly outperforming baselines; of note, some baselines are already enhanced with post-hoc metric scaling and, if applicable, TTVA.
This demonstrates \OURS{}'s ability to effectively complete geometry from limited-view sequence without in-domain training, thanks to \emph{Segmentation Forcing} and \emph{Novel-View Rendering}.
The \OURS{}$_\text{base}$ variant, which is equivalent to fine-tuning \muster{} on our datasets, was trained without the two proposed strategies and obtained only marginal improvements over baselines.

Wrong metric reasoning leads to voxels predicted outside of the scene, significantly degrading the performance.
The scale-invariant design of VGGT and AnySplat is not well-suited for the occupancy task, unlike~\OURS{} with metric prediction by design.
The Gaussian Splatting of AnySplat, while favorable for synthesizing compelling images, produces lots of geometric artifacts, thereby hallucinating lots of noises and harming geometry prediction.
~\cref{fig:qual_res} visualizes the occupancy results. 

\condenseparagraph{\Singleview{}.}
In the more challenging \emph{\Singleview{}} setting on SemanticKITTI (\cref{tbl:monocular_skitti}), \OURS{} demonstrates remarkable generalization.
It achieves $24.03\%$ IoU, outperforming all other zero-shot baselines by significant margins (\eg, $+11.00\%$ IoU over \cuter{}* w/ TTVA).
Notably, it significantly surpasses several in-domain self-supervised methods like
SceneRF ($+10.19\%$); \OURS{} even surpasses self-supervised SOTAs SelfOcc ($+2.06\%$) and OccNeRF ($+1.22\%$), despite never been trained on SemanticKITTI.

\condenseparagraph{\Surround{}.}
In the \emph{\Surround{}} setting on Occ3D-NuScenes~\cref{tab:surround-occ3d-nuscenes}, \OURS{} maintains its lead among zero-shot methods with $34.15\%$ IoU, and achieves better performance than some in-domain approaches like DistillNeRF/SimpleOcc, yet remains behind more recent methods.

\begin{table}[t]
	\small
	\resizebox{\columnwidth}{!}{
		\centering
		\begin{tabular}{l|c|c|>{\columncolor{important}}c>{\columncolor{important}}c|c|>{\columncolor{important}}c>{\columncolor{important}}c}
			\toprule
			\multirow{2}{*}{\textbf{Method}} & \multirow{2}{*}{\textbf{Venue}} & \multicolumn{3}{c|}{\textbf{Semantic KITTI \sequence{}}} & \multicolumn{3}{c}{\textbf{Occ3D-NuScenes \surround{}}} \\
			\cmidrule(lr){3-5} \cmidrule(lr){6-8} 
			& & \textbf{Res.} & \textbf{mIoU} & \textbf{mIoU$^{\rm sc}$} & \textbf{Res.} &  \textbf{mIoU} & \textbf{mIoU$^{\rm sc}$}  \\
			\midrule
			\muster~\cite{muster} + SAM2~\cite{sam2} & CVPR'25 & 512x160 & 3.22 & 5.96 & 512x288 &  2.43 & 3.84   \\
			\cuter~\cite{cuter} + SAM2~\cite{sam2} & CVPR'25 & 512x160  & 4.15&6.72& 512x288 & 2.40 & 2.75 \\
			\cuter{}*~\cite{cuter} + SAM2~\cite{sam2} & CVPR'25 & 512x160  &4.53&8.18& 512x288 & 3.06	& 3.99\\
			VGGT$^{\dagger}$~\cite{vggt} + SAM2~\cite{sam2}  & CVPR'25 & 518x294 & 3.47 & 6.76 & 518x294 & 4.39 & 6.49\\
			AnySplat$^{\dagger}$~\cite{anysplat}+ SAM2~\cite{sam2} & TOG'25 & 518x168 & 3.37 & 6.83 & 518x294 & 3.96 & 5.97 \\
			AnySplat*$^{\dagger}$~\cite{anysplat}+ SAM2~\cite{sam2} & TOG'25 & 518x168 & 3.86 & 7.51 & 518x294 & 4.44 & 6.51\\
            DA3~\cite{depthanything3}+ SAM2~\cite{sam2} & ICLR'26 & 518x168 & 4.92 &  9.56 & 518x294 & 4.55 & 6.29 \\
			\OURS{} w/o forcing + SAM2~\cite{sam2} & --  & 512x160 & \second{6.83} & \second{12.01}  & 512x288 & \second{6.17} & \second{8.96} \\
			\textbf{\OURS{}} & -- & 512x160 &  \best{7.28} & \best{13.53} & 512x288 & \best{6.66} & \best{10.32} \\
			\bottomrule
			\multicolumn{8}{l}{\small{*: use \ttva{}\,\,\,\,\,$^\dagger$: scaled with Metric3Dv2~\cite{metric3d}.}}
		\end{tabular}
	}
	\\
	\vspace{-0.2cm}
	\caption{\textbf{Semantic Occupancy Prediction} with GSAM2~\cite{gsam}.}
	\label{tbl:kitti_semantic}
    \vspace{-0.2cm}
\end{table}

\condenseparagraph{Semantic Occupancy.}
We further evaluate 3D semantic occupancy~(\cref{tbl:kitti_semantic}) by applying Grounded SAM2 pipeline directly on \OURS{}'s segmentation features.
\OURS{} achieves the highest mIoU and mIoU$^{\rm sc}$ across both datasets, compared to baselines using a separated SAM2 model to produce segmentation features.
The comparison with the variant ``\OURS{} w/o forcing + SAM2'' confirms that our \emph{Segmentation Forcing} strategy leads to a unified and simpler solution to better predict geometry and segmentation.

\condenseparagraph{Impact of base foundation models.}
We change the foundation models used in OccAny to DA3~\cite{depthanything3} and SAM3~\cite{sam3}, resulting in the \textbf{OccAny+} variant, detailed in~\cref{supp:occanyplus}. ~\cref{tbl:based_foundation_model} and ~\cref{app:more_studies} show that \OURS{} benefits from advances in generic foundation models, while being \emph{independently and orthogonally} effective for occupancy prediction.

\begin{table}[t]
\scriptsize
\setlength{\tabcolsep}{0.0055\linewidth}
\centering
\resizebox{\columnwidth}{!}{
\begin{tabular}{l|c|cc>{\columncolor{important}}c>{\columncolor{important}}c>{\columncolor{important}}c|c|cc>{\columncolor{important}}c>{\columncolor{important}}c>{\columncolor{important}}c}
\toprule
Method & \multicolumn{6}{c|}{\textbf{Semantic KITTI sequence}} & \multicolumn{6}{c}{\textbf{Occ3D-NuScenes surround-view}} \\
\cmidrule(lr){2-7} \cmidrule(lr){8-13}
& \textbf{Res.} & Pre. & Rec. & \textbf{IoU} & \textbf{mIoU} & \textbf{mIoU$^{\rm sc}$} & \textbf{Res.} & Pre. & Rec. & \textbf{IoU} & \textbf{mIoU} & \textbf{mIoU$^{\rm sc}$} \\
\midrule
\textbf{OccAny} & 512x160 & 36.79 & 46.70 & 25.91 & \textbf{7.28} & \textbf{13.53} & 512x288 & \textbf{45.04} & \textbf{58.54} & \textbf{34.15} & 6.66 & 10.32 \\
\textbf{OccAny+} & 512x160 & \textbf{38.12} & \textbf{49.14} & \textbf{27.33} & 6.48 & 13.30 & 512x288 & 46.38 & 54.66 & 33.49 & \textbf{7.20} & \textbf{11.50}\\
\bottomrule
\end{tabular}}
\caption{\textbf{Changing the base foundation models} used in OccAny to DA3~\cite{depthanything3} and SAM3~\cite{sam3} results in the \textbf{OccAny+} variant.}
\label{tbl:based_foundation_model}
\end{table}

\begin{table}[t]
    \scriptsize
	\setlength{\tabcolsep}{0.0055\linewidth}
    \setlength\fboxsep{0pt}
	\centering
    \resizebox{0.8\columnwidth}{!}{
	\begin{tabular}{ccccc|cl|cl}
		\toprule
    	\multirow{2}{*}{\rot{\text{{NVR}}}} & \multirow{2}{*}{\rot{$\mathcal{L}_\text{forcing}$}} & \multirow{2}{*}{\rot{\tiny{geo-aware}}} & \multirow{2}{*}{\rot{$\taskG+\taskS$}} & \multirow{2}{*}{\rot{$\mathcal{L}_\text{enc}$}} & \multicolumn{2}{c|}{\textbf{SemKITTI seq.}}  &  \multicolumn{2}{c}{\textbf{SemKITTI single}}    \\
        \cmidrule(lr){6-7} \cmidrule(lr){8-9}
        &&&&& \textbf{IoU} & \makecell{\textbf{$\Delta$ \textbf{IoU}}} & \textbf{IoU} & \makecell{\textbf{$\Delta$ \textbf{IoU}}} \\
		\midrule
		\redcross& & & & & 19.64  & \cbarm{40.8}{6}{0} -6.27  & 11.55 & \cbarm{49.95}{6}{0}-12.48 \\
          & \redcross &  &  &  & 24.23 & \cbarm{16.0}{6}{0} -1.68 & 21.73 & \cbarm{21.14}{6}{0}-2.30\\
          
         & & \redcross &  &  & 24.88 & \cbarm{9.5}{6}{0} -1.03 & 23.02 & \cbarm{8.5}{6}{0} -1.01\\
          & &  & \redcross &  & 25.04 & \cbarm{7.9}{6}{0} -0.87  & 22.67 & \cbarm{12.0}{6}{0} -1.36\\
          &  &  &  &  \redcross & 24.99 & \cbarm{8.4}{6}{0} -0.92  & 23.46 & \cbarm{4.1}{6}{0} -0.57\\
         \rowcolor{gray!20} \multicolumn{5}{c|}{\textbf{\OURS{}}}  & 25.91 &  --- & 24.03 & --- \\
		\bottomrule
	\end{tabular}
	}
    \vspace{-0.2cm}
    \caption{\textbf{Ablation results} on SemanticKitti. The ``geo-aware'' stands for applying geometry confidence maps $\conf$ in the segmentation forcing loss (\cf~\cref{eqn:forcing_loss}).}
    \vspace{-0.2cm}
	\label{tbl:ablation_res}
\end{table}

\subsection{Analysis}
\condenseparagraph{Method ablation.}
Tab.~\ref{tbl:ablation_res} analyzes the contribution of each proposed component.
Removing Test-Time View Augmentation (TTVA) causes the most significant drop ($-6.27\%$ in sequence- and $-12.47\%$ in \singleview{} setting), highlighting its critical role in geometry completion.
The rendering-specific losses $\mathcal{L}_\text{Enc}$, geometry-aware $\mathcal{L}_{\rm forcing}$, and the task tokens also consistently contribute to the final performance, proving their effectiveness.
~\cref{fig:qualitative-occupancy} shows gains brought by Segmentation Forcing and Novel-view Rendering (TTVA).

\begin{figure}[t!]
    \centering
    \includegraphics[width=.68\linewidth]{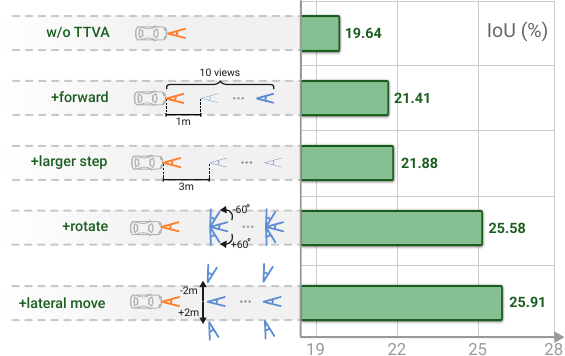}
    \vspace{-0.2cm}
    \caption{\textbf{Ablating NVR inference} on SemanticKITTI}
    \label{fig:ablation_ttva}
    \vspace{-0.2cm}
\end{figure}
\condenseparagraph{NVR inference.}
We ablate NVR inference in~\cref{fig:ablation_ttva}.
Starting from the baseline without TTVA, adding simple forward movement helps complete distant geometry ($+1.83\%$).
Introducing rotations and lateral shifts further helps complete the geometry by resolving occlusions from diverse views, improving IoU by $+4.15\%$ and resulting in the final $25.91\%$.

\begin{figure}[t!]
    \centering
    \footnotesize
    \setlength{\tabcolsep}{0pt}
    \renewcommand{\arraystretch}{0}
    \newcommand{\imgsizer}{.24\linewidth}
    \newcommand{\pcaRow}[2]{%
    \includegraphics[width=\imgsizer, trim=#2, clip]{figs/pca/#1_saved_colors.png} &
    \includegraphics[width=\imgsizer, trim=#2, clip]{figs/pca/#1_pca_image_embed.png} &
    \includegraphics[width=\imgsizer, trim=#2, clip]{figs/pca/#1_pca_high_res_1.png} &
    \includegraphics[width=\imgsizer, trim=#2, clip]{figs/pca/#1_pca_high_res_0.png}}
    \begin{tabular}{cccc}    
        RGB & low & med & high \\ 
        \pcaRow{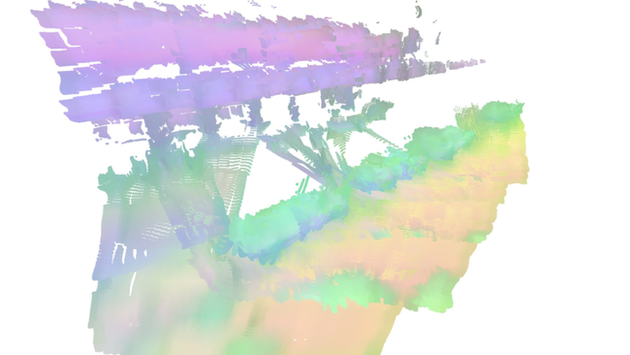}{1cm 1cm 1cm 1cm} \\
        \pcaRow{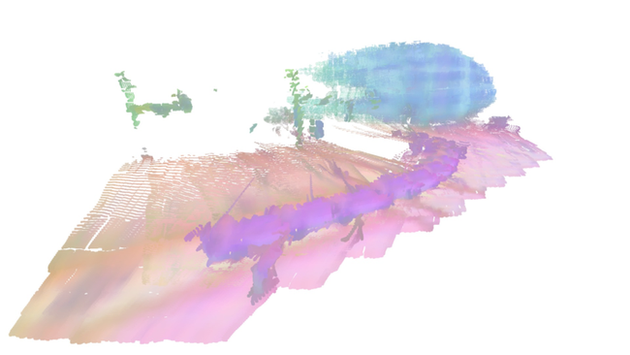}{1cm 1cm 1cm 1cm} \\

    \end{tabular}
    \vspace{-0.2cm}
    \caption{\textbf{PCA visualization of our segmentation features} of multi-view sequences. Low-resolution features capture high-level semantics (\eg, separating cars, buildings, and roads), while high-resolution features capture low-level details such as boundaries and textures. Features remain consistent across different views.}
    \vspace{-0.2cm}
    \label{fig:PCA_sam2}
\end{figure}

\condenseparagraph{Promptable segmentation feature.}
We visualize the segmentation features of \OURS{} using PCA, as shown in~\cref{fig:PCA_sam2}.
Low-resolution features appear to cluster semantically similar regions, while high-resolution features seem to capture fine details like boundaries and textures, both helping regularize and improve occupancy prediction (\cf~\cref{fig:qualitative-occupancy} \& ~\cref{tbl:ablation_res}).

\begin{figure}[t!]
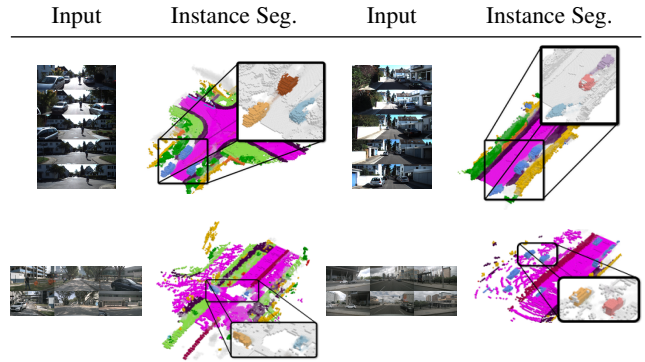

	\centering
	\small
	
	\setlength{\tabcolsep}{0pt}
	\newcommand{\imgsizer}{.14\textwidth}
	\newcommand{\trimWidth}{0.35}
	\newcommand{\trimHeight}{0.35}
	\newcommand{\trimLeft}{0.00}
	\newcommand{\trimBot}{0.05}
	\newcommand{\kittiInsRow}[5]{%
		\makebox[\linewidth]{\adjincludegraphics[width=.6\linewidth]{figs/kitti_5frames_3opa_instance/#1_stacked_recon.png}} &
		\adjincludegraphics[width=\linewidth,trim={#2\width} {#3\height} {#4\width} {#5\height}, clip]{figs/kitti_5frames_3opa_instance/#1_ins.png} 
	}
	\newcommand{\nuscInsRow}[5]{%
		\makebox[\linewidth]{\adjincludegraphics[width=\linewidth]{figs/nuscenes_surround_10opa_instance/#1_stacked_recon.png}} &
		\adjincludegraphics[width=\linewidth,trim={#2\width} {#3\height} {#4\width} {#5\height}, clip]{figs/nuscenes_surround_10opa_instance/#1.png} 
	}

	\begin{tabular}{m{0.1\textwidth}m{\imgsizer}m{0.1\textwidth}m{\imgsizer}}
		\multicolumn{1}{c}{Input} 
		& \multicolumn{1}{c}{Instance Seg.} & \multicolumn{1}{c}{Input} 
		& \multicolumn{1}{c}{Instance Seg.}  \\
		\midrule
		\kittiInsRow{08_001250}{0.15}{0.15}{0.35}{0.35} & \kittiInsRow{08_002280}{0.15}{0.10}{0.4}{0.34} \\
		\nuscInsRow{scene-0035_5f15115414584a23bdc0d0982c46e264}{0.25}{0.28}{0.32}{0.25} & \nuscInsRow{scene-0098_d5c86824b36b4dbba467399bf3123d90}{0.25}{0.4}{0.35}{0.32}
		
	\end{tabular}
    \vspace{-0.2cm}
	\caption{\textbf{Instance segmentation} of cars with \OURS{}'s features.}
    \vspace{-0.2cm}
	\label{fig:instance}
\end{figure}

Similar to SAM2, our segmentation features remain spatially and temporally consistent.
This consistency enables instance segmentation via prompting with object instances detected by Grounding DINO.
In~\cref{fig:instance}, we show some qualitative results when performing instance segmentation directly on our segmentation features.


\section{Conclusion}
We propose for the first time a generalized 3D occupancy network, called~\OURS{}, that is trained once and perform zero-shot inference on arbitrary out-of-domain sequential, monocular and surround-view unposed data.
With the proposed Segmentation Forcing and Novel-View Rendering strategies,~\OURS{} outperforms generic visual-geometry foundation models on occupancy prediction.
\OURS{} surpasses several in-domain self-supervised models, while remaining behind more recent ones.
Our work introduces a novel framework for occupancy prediction prioritizing scalability and generalization, paving the way toward the next generation of versatile and generalized occupancy networks.
The gap to fully-supervised in-domain performance remains substantial, leaving room for future improvements in this direction.

\newpage
{\footnotesize
\condenseparagraph{Acknowledgment.} 
{This
work was granted access to the HPC resources of IDRIS
under the allocations AD011014102R2, AD011013540R1 made by GENCI. We acknowledge EuroHPC Joint Undertaking for awarding the project ID
EHPC-REG-2025R01-032 access to Karolina, Czech Republic. This work was supported by the European Union’s Horizon Europe research and innovation programme under grant agreement No 101214398 (ELLIOT). }\par}

{
    \small
    \bibliographystyle{ieeenat_fullname}
    \bibliography{main}

\begin{thebibliography}{89}
\providecommand{\natexlab}[1]{#1}
\providecommand{\url}[1]{\texttt{#1}}
\expandafter\ifx\csname urlstyle\endcsname\relax
  \providecommand{\doi}[1]{doi: #1}\else
  \providecommand{\doi}{doi: \begingroup \urlstyle{rm}\Url}\fi

\bibitem[Behley et~al.(2019)Behley, Garbade, Milioto, Quenzel, Behnke, Stachniss, and Gall]{semkitti}
Jens Behley, Martin Garbade, Andres Milioto, Jan Quenzel, Sven Behnke, Cyrill Stachniss, and Juergen Gall.
\newblock Semantickitti: A dataset for semantic scene understanding of lidar sequences.
\newblock In \emph{ICCV}, 2019.

\bibitem[Cabon et~al.(2020)Cabon, Murray, and Humenberger]{vkitti2}
Yohann Cabon, Naila Murray, and Martin Humenberger.
\newblock Virtual kitti 2.
\newblock In \emph{arXiv}, 2020.

\bibitem[Cabon et~al.(2025)Cabon, Stoffl, Antsfeld, Csurka, Chidlovskii, Revaud, and Leroy]{muster}
Yohann Cabon, Lucas Stoffl, Leonid Antsfeld, Gabriela Csurka, Boris Chidlovskii, Jerome Revaud, and Vincent Leroy.
\newblock Must3r: Multi-view network for stereo 3d reconstruction.
\newblock In \emph{CVPR}, 2025.

\bibitem[Caesar et~al.(2020)Caesar, Bankiti, Lang, Vora, Liong, Xu, Krishnan, Pan, Baldan, and Beijbom]{nuscenes}
Holger Caesar, Varun Bankiti, Alex~H. Lang, Sourabh Vora, Venice~Erin Liong, Qiang Xu, Anush Krishnan, Yu Pan, Giancarlo Baldan, and Oscar Beijbom.
\newblock nuscenes: A multimodal dataset for autonomous driving.
\newblock In \emph{CVPR}, 2020.

\bibitem[Cao and de~Charette(2022)]{monoscene}
Anh-Quan Cao and Raoul de Charette.
\newblock Monoscene: Monocular 3d semantic scene completion.
\newblock In \emph{CVPR}, 2022.

\bibitem[Cao and de~Charette(2023)]{scenerf}
Anh-Quan Cao and Raoul de Charette.
\newblock Scenerf: Self-supervised monocular 3d scene reconstruction with radiance fields.
\newblock In \emph{ICCV}, 2023.

\bibitem[Cao et~al.(2024)Cao, Dai, and de~Charette]{pasco}
Anh-Quan Cao, Angela Dai, and Raoul de Charette.
\newblock Pasco: Urban 3d panoptic scene completion with uncertainty awareness.
\newblock In \emph{CVPR}, 2024.

\bibitem[Carion et~al.(2026)Carion, Gustafson, Hu, Debnath, Hu, Suris, Ryali, Alwala, Khedr, Huang, Lei, Ma, Guo, Kalla, Marks, Greer, Wang, Sun, Rädle, Afouras, Mavroudi, Xu, Wu, Zhou, Momeni, Hazra, Ding, Vaze, Porcher, Li, Li, Kamath, Cheng, Dollár, Ravi, Saenko, Zhang, and Feichtenhofer]{sam3}
Nicolas Carion, Laura Gustafson, Yuan-Ting Hu, Shoubhik Debnath, Ronghang Hu, Didac Suris, Chaitanya Ryali, Kalyan~Vasudev Alwala, Haitham Khedr, Andrew Huang, Jie Lei, Tengyu Ma, Baishan Guo, Arpit Kalla, Markus Marks, Joseph Greer, Meng Wang, Peize Sun, Roman Rädle, Triantafyllos Afouras, Effrosyni Mavroudi, Katherine Xu, Tsung-Han Wu, Yu Zhou, Liliane Momeni, Rishi Hazra, Shuangrui Ding, Sagar Vaze, Francois Porcher, Feng Li, Siyuan Li, Aishwarya Kamath, Ho~Kei Cheng, Piotr Dollár, Nikhila Ravi, Kate Saenko, Pengchuan Zhang, and Christoph Feichtenhofer.
\newblock Sam 3: Segment anything with concepts.
\newblock In \emph{ICLR}, 2026.

\bibitem[Chambon et~al.(2025)Chambon, Zablocki, Boulch, Chen, and Cord]{chambon2025gaussrender}
Loick Chambon, Eloi Zablocki, Alexandre Boulch, Mickael Chen, and Matthieu Cord.
\newblock Gaussrender: Learning 3d occupancy with gaussian rendering.
\newblock In \emph{CVPR}, 2025.

\bibitem[Chang et~al.(2015)Chang, Funkhouser, Guibas, Hanrahan, Huang, Li, Savarese, Savva, Song, Su, et~al.]{shapenet}
Angel~X Chang, Thomas Funkhouser, Leonidas Guibas, Pat Hanrahan, Qixing Huang, Zimo Li, Silvio Savarese, Manolis Savva, Shuran Song, Hao Su, et~al.
\newblock Shapenet: An information-rich 3d model repository.
\newblock \emph{arXiv}, 2015.

\bibitem[Chen et~al.(2025)Chen, Fang, Han, Cheng, Yin, Xu, Khan, and Shen]{alocc}
Dubing Chen, Jin Fang, Wencheng Han, Xinjing Cheng, Junbo Yin, Chenzhong Xu, Fahad~Shahbaz Khan, and Jianbing Shen.
\newblock Alocc: adaptive lifting-based 3d semantic occupancy and cost volume-based flow prediction.
\newblock In \emph{ICCV}, 2025.

\bibitem[Choy et~al.(2019)Choy, Gwak, and Savarese]{choy20194d}
Christopher Choy, JunYoung Gwak, and Silvio Savarese.
\newblock 4d spatio-temporal convnets: Minkowski convolutional neural networks.
\newblock In \emph{CVPR}, 2019.

\bibitem[Dai et~al.(2017)Dai, Chang, Savva, Halber, Funkhouser, and Nie{\ss}ner]{scannet}
Angela Dai, Angel~X Chang, Manolis Savva, Maciej Halber, Thomas Funkhouser, and Matthias Nie{\ss}ner.
\newblock Scannet: Richly-annotated 3d reconstructions of indoor scenes.
\newblock In \emph{CVPR}, 2017.

\bibitem[Gan et~al.(2024)Gan, Mo, Xu, and Yokoya]{simpleocc}
Wanshui Gan, Ningkai Mo, Hongbin Xu, and Naoto Yokoya.
\newblock A comprehensive framework for 3d occupancy estimation in autonomous driving.
\newblock \emph{IEEE TIV}, 2024.

\bibitem[Gan et~al.(2025)Gan, Liu, Xu, Mo, and Yokoya]{GaussianOcc}
Wanshui Gan, Fang Liu, Hongbin Xu, Ningkai Mo, and Naoto Yokoya.
\newblock Gaussianocc: Fully self-supervised and efficient 3d occupancy estimation with gaussian splatting.
\newblock In \emph{ICCV}, 2025.

\bibitem[Gao et~al.(2024)Gao, Yang, Chen, Chitta, Qiu, Geiger, Zhang, and Li]{vista}
Shenyuan Gao, Jiazhi Yang, Li Chen, Kashyap Chitta, Yihang Qiu, Andreas Geiger, Jun Zhang, and Hongyang Li.
\newblock Vista: A generalizable driving world model with high fidelity and versatile controllability.
\newblock In \emph{NeurIPS}, 2024.

\bibitem[Gebraad et~al.(2025)Gebraad, Palffy, and Caesar]{leap}
Simon Gebraad, Andras Palffy, and Holger Caesar.
\newblock Leap: Consistent multi-domain 3d labeling using foundation models.
\newblock In \emph{ICRA}, 2025.

\bibitem[Geiger et~al.(2012)Geiger, Lenz, and Urtasun]{kitti}
Andreas Geiger, Philip Lenz, and Raquel Urtasun.
\newblock Are we ready for autonomous driving? the kitti vision benchmark suite.
\newblock In \emph{CVPR}, 2012.

\bibitem[Guizilini et~al.(2020)Guizilini, Ambrus, Pillai, Raventos, and Gaidon]{ddad}
Vitor Guizilini, Rares Ambrus, Sudeep Pillai, Allan Raventos, and Adrien Gaidon.
\newblock 3d packing for self-supervised monocular depth estimation.
\newblock In \emph{CVPR}, 2020.

\bibitem[Hassan et~al.(2025)Hassan, Stapf, Rahimi, Rezende, Haghighi, Brüggemann, Katircioglu, Zhang, Chen, Saha, Cannici, Aljalbout, Ye, Wang, Davtyan, Salzmann, Scaramuzza, Pollefeys, Favaro, and Alahi]{gem}
Mariam Hassan, Sebastian Stapf, Ahmad Rahimi, Pedro M.~B. Rezende, Yasaman Haghighi, David Brüggemann, Isinsu Katircioglu, Lin Zhang, Xiaoran Chen, Suman Saha, Marco Cannici, Elie Aljalbout, Botao Ye, Xi Wang, Aram Davtyan, Mathieu Salzmann, Davide Scaramuzza, Marc Pollefeys, Paolo Favaro, and Alexandre Alahi.
\newblock Gem: A generalizable ego-vision multimodal world model for fine-grained ego-motion, object dynamics, and scene composition control.
\newblock In \emph{CVPR}, 2025.

\bibitem[Hayler et~al.(2024)Hayler, Wimbauer, Muhle, Rupprecht, and Cremers]{sfourc}
Adrian Hayler, Felix Wimbauer, Dominik Muhle, Christian Rupprecht, and Daniel Cremers.
\newblock S4c: Self-supervised semantic scene completion with neural fields.
\newblock In \emph{3DV}, 2024.

\bibitem[Hu et~al.(2024)Hu, Yin, Zhang, Cai, Long, Chen, Wang, Yu, Shen, and Shen]{metric3d}
Mu Hu, Wei Yin, Chi Zhang, Zhipeng Cai, Xiaoxiao Long, Hao Chen, Kaixuan Wang, Gang Yu, Chunhua Shen, and Shaojie Shen.
\newblock Metric3d v2: A versatile monocular geometric foundation model for zero-shot metric depth and surface normal estimation.
\newblock \emph{IEEE TPAMI}, 2024.

\bibitem[Huang et~al.(2023)Huang, Zheng, Zhang, Zhou, and Lu]{tpvformer}
Yuanhui Huang, Wenzhao Zheng, Yunpeng Zhang, Jie Zhou, and Jiwen Lu.
\newblock Tri-perspective view for vision-based 3d semantic occupancy prediction.
\newblock In \emph{CVPR}, 2023.

\bibitem[Huang et~al.(2024{\natexlab{a}})Huang, Zheng, Zhang, Zhou, and Lu]{selfocc}
Yuanhui Huang, Wenzhao Zheng, Borui Zhang, Jie Zhou, and Jiwen Lu.
\newblock Selfocc: Self-supervised vision-based 3d occupancy prediction.
\newblock In \emph{CVPR}, 2024{\natexlab{a}}.

\bibitem[Huang et~al.(2024{\natexlab{b}})Huang, Zheng, Zhang, Zhou, and Lu]{GaussianFormer}
Yuanhui Huang, Wenzhao Zheng, Yunpeng Zhang, Jie Zhou, and Jiwen Lu.
\newblock Gaussianformer: Scene as gaussians for vision-based 3d semantic occupancy prediction.
\newblock In \emph{ECCV}, 2024{\natexlab{b}}.

\bibitem[Jang et~al.(2025)Jang, Weinzaepfel, Leroy, Agapito, and Revaud]{pow3r}
Wonbong Jang, Philippe Weinzaepfel, Vincent Leroy, Lourdes Agapito, and Jerome Revaud.
\newblock Pow3r: Empowering unconstrained 3d reconstruction with camera and scene priors.
\newblock In \emph{CVPR}, 2025.

\bibitem[Jevti{\'c} et~al.(2025)Jevti{\'c}, Reich, Wimbauer, Hahn, Rupprecht, Roth, and Cremers]{SceneDINO}
Aleksandar Jevti{\'c}, Christoph Reich, Felix Wimbauer, Oliver Hahn, Christian Rupprecht, Stefan Roth, and Daniel Cremers.
\newblock Feed-forward scenedino for unsupervised semantic scene completion.
\newblock In \emph{ECCV}, 2025.

\bibitem[Jiang et~al.(2025{\natexlab{a}})Jiang, Liu, Cheng, Wang, Lin, Su, Liu, and Wang]{gausstr}
Haoyi Jiang, Liu Liu, Tianheng Cheng, Xinjie Wang, Tianwei Lin, Zhizhong Su, Wenyu Liu, and Xinggang Wang.
\newblock Gausstr: Foundation model-aligned gaussian transformer for self-supervised 3d spatial understanding.
\newblock In \emph{CVPR}, 2025{\natexlab{a}}.

\bibitem[Jiang et~al.(2025{\natexlab{b}})Jiang, Mao, Xu, Lu, Ren, Jin, Xu, Yu, Pang, Zhao, Lin, and Dai]{anysplat}
Lihan Jiang, Yucheng Mao, Linning Xu, Tao Lu, Kerui Ren, Yichen Jin, Xudong Xu, Mulin Yu, Jiangmiao Pang, Feng Zhao, Dahua Lin, and Bo Dai.
\newblock Anysplat: Feed-forward 3d gaussian splatting from unconstrained views.
\newblock \emph{ACM TOG}, 2025{\natexlab{b}}.

\bibitem[Kerbl et~al.(2023)Kerbl, Kopanas, Leimk{\"u}hler, and Drettakis]{gsplat}
Bernhard Kerbl, Georgios Kopanas, Thomas Leimk{\"u}hler, and George Drettakis.
\newblock 3d gaussian splatting for real-time radiance field rendering.
\newblock \emph{ACM TOG}, 2023.

\bibitem[Leroy et~al.(2024)Leroy, Cabon, and Revaud]{mast3r}
Vincent Leroy, Yohann Cabon, and Jerome Revaud.
\newblock Grounding image matching in 3d with mast3r.
\newblock In \emph{ECCV}, 2024.

\bibitem[Li et~al.(2024{\natexlab{a}})Li, Sun, Jin, Zeng, Zhu, Wang, Zhang, Okae, Xiao, and Du]{li2023stereoscene}
Bohan Li, Yasheng Sun, Xin Jin, Wenjun Zeng, Zheng Zhu, Xiaoefeng Wang, Yunpeng Zhang, James Okae, Hang Xiao, and Dalong Du.
\newblock Stereoscene: Bev-assisted stereo matching empowers 3d semantic scene completion.
\newblock In \emph{IJCAI}, 2024{\natexlab{a}}.

\bibitem[Li et~al.(2025)Li, Kachana, Chidananda, Nair, Furukawa, and Brown]{rig3r}
Samuel Li, Pujith Kachana, Prajwal Chidananda, Saurabh Nair, Yasutaka Furukawa, and Matthew Brown.
\newblock Rig3r: Rig-aware conditioning for learned 3d reconstruction.
\newblock In \emph{NeurIPS}, 2025.

\bibitem[Li et~al.(2023)Li, Yu, Choy, Xiao, Alvarez, Fidler, Feng, and Anandkumar]{voxformer}
Yiming Li, Zhiding Yu, Christopher Choy, Chaowei Xiao, Jose~M Alvarez, Sanja Fidler, Chen Feng, and Anima Anandkumar.
\newblock Voxformer: Sparse voxel transformer for camera-based 3d semantic scene completion.
\newblock In \emph{CVPR}, 2023.

\bibitem[Li et~al.(2024{\natexlab{b}})Li, Li, Liu, Gong, Li, Chen, Wang, Li, Jiang, Yu, Wang, Zhao, Yu, and Feng]{sscbench}
Yiming Li, Sihang Li, Xinhao Liu, Moonjun Gong, Kenan Li, Nuo Chen, Zijun Wang, Zhiheng Li, Tao Jiang, Fisher Yu, Yue Wang, Hang Zhao, Zhiding Yu, and Chen Feng.
\newblock Sscbench: A large-scale 3d semantic scene completion benchmark for autonomous driving.
\newblock In \emph{IROS}, 2024{\natexlab{b}}.

\bibitem[Lin et~al.(2025)Lin, Chen, Liew, Chen, Li, Shi, Feng, and Kang]{depthanything3}
Haotong Lin, Sili Chen, Jun~Hao Liew, Donny~Y. Chen, Zhenyu Li, Guang Shi, Jiashi Feng, and Bingyi Kang.
\newblock Depth anything 3: Recovering the visual space from any views.
\newblock \emph{arXiv}, 2025.

\bibitem[Liu et~al.(2024{\natexlab{a}})Liu, Wang, Chen, Yang, Zeng, Chen, and Wang]{SparseOcc}
Haisong Liu, Haiguang Wang, Yang Chen, Zetong Yang, Jia Zeng, Li Chen, and Limin Wang.
\newblock Fully sparse 3d panoptic occupancy prediction.
\newblock In \emph{ECCV}, 2024{\natexlab{a}}.

\bibitem[Liu et~al.(2024{\natexlab{b}})Liu, Zeng, Ren, Li, Zhang, Yang, Jiang, Li, Yang, Su, et~al.]{gdino}
Shilong Liu, Zhaoyang Zeng, Tianhe Ren, Feng Li, Hao Zhang, Jie Yang, Qing Jiang, Chunyuan Li, Jianwei Yang, Hang Su, et~al.
\newblock Grounding dino: Marrying dino with grounded pre-training for open-set object detection.
\newblock In \emph{ECCV}, 2024{\natexlab{b}}.

\bibitem[Loshchilov and Hutter(2019)]{adamw}
Ilya Loshchilov and Frank Hutter.
\newblock Decoupled weight decay regularization.
\newblock In \emph{ICLR}, 2019.

\bibitem[Ma et~al.(2024{\natexlab{a}})Ma, Chen, Huang, Xu, Luo, Xu, Gu, Ai, and Wang]{Cam4DOcc}
Junyi Ma, Xieyuanli Chen, Jiawei Huang, Jingyi Xu, Zhen Luo, Jintao Xu, Weihao Gu, Rui Ai, and Hesheng Wang.
\newblock Cam4docc: Benchmark for camera-only 4d occupancy forecasting in autonomous driving applications.
\newblock In \emph{CVPR}, 2024{\natexlab{a}}.

\bibitem[Ma et~al.(2024{\natexlab{b}})Ma, Tan, Qu, Ma, Zhang, and Xie]{COTR}
Qihang Ma, Xin Tan, Yanyun Qu, Lizhuang Ma, Zhizhong Zhang, and Yuan Xie.
\newblock Cotr: Compact occupancy transformer for vision-based 3d occupancy prediction.
\newblock In \emph{CVPR}, 2024{\natexlab{b}}.

\bibitem[Mao et~al.(2021)Mao, Niu, Jiang, Liang, Chen, Liang, Li, Ye, Zhang, Li, et~al.]{once}
Jiageng Mao, Minzhe Niu, Chenhan Jiang, Hanxue Liang, Jingheng Chen, Xiaodan Liang, Yamin Li, Chaoqiang Ye, Wei Zhang, Zhenguo Li, et~al.
\newblock One million scenes for autonomous driving: Once dataset.
\newblock In \emph{NeurIPS}, 2021.

\bibitem[Marcuzzi et~al.(2025)Marcuzzi, Nunes, Marks, Wiesmann, L\"abe, Behley, and Stachniss]{sfmocc}
R. Marcuzzi, L. Nunes, E.A. Marks, L. Wiesmann, T. L\"abe, J. Behley, and C. Stachniss.
\newblock {SfmOcc: Vision-Based 3D Semantic Occupancy Prediction in Urban Environments}.
\newblock \emph{RA-L}, 2025.

\bibitem[Mildenhall et~al.(2021)Mildenhall, Srinivasan, Tancik, Barron, Ramamoorthi, and Ng]{nerf}
Ben Mildenhall, Pratul~P. Srinivasan, Matthew Tancik, Jonathan~T. Barron, Ravi Ramamoorthi, and Ren Ng.
\newblock Nerf: representing scenes as neural radiance fields for view synthesis.
\newblock \emph{Commun. ACM}, 2021.

\bibitem[Ošep et~al.(2024)Ošep, Meinhardt, Ferroni, Peri, Ramanan, and Leal-Taixé]{sal}
Aljoša Ošep, Tim Meinhardt, Francesco Ferroni, Neehar Peri, Deva Ramanan, and Laura Leal-Taixé.
\newblock Better call sal: Towards learning to segment anything in lidar.
\newblock In \emph{ECCV}, 2024.

\bibitem[Qi et~al.(2017)Qi, Su, Mo, and Guibas]{qi2017pointnet}
Charles~R Qi, Hao Su, Kaichun Mo, and Leonidas~J Guibas.
\newblock Pointnet: Deep learning on point sets for 3d classification and segmentation.
\newblock In \emph{CVPR}, 2017.

\bibitem[Ravi et~al.(2025)Ravi, Gabeur, Hu, Hu, Ryali, Ma, Khedr, R{\"a}dle, Rolland, Gustafson, Mintun, Pan, Alwala, Carion, Wu, Girshick, Doll{\'a}r, and Feichtenhofer]{sam2}
Nikhila Ravi, Valentin Gabeur, Yuan-Ting Hu, Ronghang Hu, Chaitanya Ryali, Tengyu Ma, Haitham Khedr, Roman R{\"a}dle, Chloe Rolland, Laura Gustafson, Eric Mintun, Junting Pan, Kalyan~Vasudev Alwala, Nicolas Carion, Chao-Yuan Wu, Ross Girshick, Piotr Doll{\'a}r, and Christoph Feichtenhofer.
\newblock Sam 2: Segment anything in images and videos.
\newblock In \emph{ICLR}, 2025.

\bibitem[Ren et~al.(2024)Ren, Liu, Zeng, Lin, Li, Cao, Chen, Huang, Chen, Yan, et~al.]{gsam}
Tianhe Ren, Shilong Liu, Ailing Zeng, Jing Lin, Kunchang Li, He Cao, Jiayu Chen, Xinyu Huang, Yukang Chen, Feng Yan, et~al.
\newblock Grounded sam: Assembling open-world models for diverse visual tasks.
\newblock In \emph{arXiv}, 2024.

\bibitem[Rold{\~a}o et~al.(2020)Rold{\~a}o, de~Charette, and Verroust-Blondet]{lmscnet}
Luis Rold{\~a}o, Raoul de Charette, and Anne Verroust-Blondet.
\newblock Lmscnet: Lightweight multiscale 3d semantic completion.
\newblock In \emph{3DV}, 2020.

\bibitem[Samet et~al.(2026)Samet, Puy, and Marlet]{losc}
Nermin Samet, Gilles Puy, and Renaud Marlet.
\newblock Losc: Lidar open-voc segmentation consolidator.
\newblock In \emph{3DV}, 2026.

\bibitem[Shi et~al.(2024)Shi, Cheng, Zhang, Liu, and Wang]{OSP}
Yiang Shi, Tianheng Cheng, Qian Zhang, Wenyu Liu, and Xinggang Wang.
\newblock Occupancy as set of points.
\newblock In \emph{ECCV}, 2024.

\bibitem[Sirko-Galouchenko et~al.(2024)Sirko-Galouchenko, Boulch, Gidaris, Bursuc, Vobecky, P{\'e}rez, and Marlet]{OccFeat}
Sophia Sirko-Galouchenko, Alexandre Boulch, Spyros Gidaris, Andrei Bursuc, Antonin Vobecky, Patrick P{\'e}rez, and Renaud Marlet.
\newblock Occfeat: Self-supervised occupancy feature prediction for pretraining bev segmentation networks.
\newblock In \emph{CVPR}, 2024.

\bibitem[Song et~al.(2017)Song, Yu, Zeng, Chang, Savva, and Funkhouser]{SSCNet}
Shuran Song, Fisher Yu, Andy Zeng, Angel~X Chang, Manolis Savva, and Thomas Funkhouser.
\newblock Semantic scene completion from a single depth image.
\newblock In \emph{CVPR}, pages 1746--1754, 2017.

\bibitem[Sucar et~al.(2025)Sucar, Lai, Insafutdinov, and Vedaldi]{dynamicpointmaps}
Edgar Sucar, Zihang Lai, Eldar Insafutdinov, and Andrea Vedaldi.
\newblock Dynamic point maps: A versatile representation for dynamic 3d reconstruction.
\newblock In \emph{ICCV}, 2025.

\bibitem[Sun et~al.(2020)Sun, Kretzschmar, Dotiwalla, Chouard, Patnaik, Tsui, Guo, Zhou, Chai, Caine, et~al.]{waymo}
Pei Sun, Henrik Kretzschmar, Xerxes Dotiwalla, Aurelien Chouard, Vijaysai Patnaik, Paul Tsui, James Guo, Yin Zhou, Yuning Chai, Benjamin Caine, et~al.
\newblock Scalability in perception for autonomous driving: Waymo open dataset.
\newblock In \emph{CVPR}, 2020.

\bibitem[Szymanowicz et~al.(2024)Szymanowicz, Rupprecht, and Vedaldi]{splatterimage}
Stanislaw Szymanowicz, Chrisitian Rupprecht, and Andrea Vedaldi.
\newblock Splatter image: Ultra-fast single-view 3d reconstruction.
\newblock In \emph{CVPR}, 2024.

\bibitem[Teed and Deng(2021)]{droidslam}
Zachary Teed and Jia Deng.
\newblock {DROID-SLAM: Deep Visual SLAM for Monocular, Stereo, and RGB-D Cameras}.
\newblock In \emph{NeurIPS}, 2021.

\bibitem[Thomas et~al.(2019)Thomas, Qi, Deschaud, Marcotegui, Goulette, and Guibas]{thomas2019kpconv}
Hugues Thomas, Charles~R Qi, Jean-Emmanuel Deschaud, Beatriz Marcotegui, Fran{\c{c}}ois Goulette, and Leonidas~J Guibas.
\newblock Kpconv: Flexible and deformable convolution for point clouds.
\newblock In \emph{ICCV}, 2019.

\bibitem[Tian et~al.(2023)Tian, Jiang, Yun, Mao, Yang, Wang, Wang, and Zhao]{occ3d}
Xiaoyu Tian, Tao Jiang, Longfei Yun, Yucheng Mao, Huitong Yang, Yue Wang, Yilun Wang, and Hang Zhao.
\newblock Occ3d: A large-scale 3d occupancy prediction benchmark for autonomous driving.
\newblock In \emph{NeurIPS}, 2023.

\bibitem[Veicht et~al.(2024)Veicht, Sarlin, Lindenberger, and Pollefeys]{geocalib}
Alexander Veicht, Paul-Edouard Sarlin, Philipp Lindenberger, and Marc Pollefeys.
\newblock {GeoCalib: Single-image Calibration with Geometric Optimization}.
\newblock In \emph{ECCV}, 2024.

\bibitem[Vobecky et~al.(2023)Vobecky, Siméoni, Hurych, Gidaris, Bursuc, Pérez, and Sivic]{pop3d}
Antonin Vobecky, Oriane Siméoni, David Hurych, Spyros Gidaris, Andrei Bursuc, Patrick Pérez, and Josef Sivic.
\newblock Pop-3d: Open-vocabulary 3d occupancy prediction from images.
\newblock In \emph{NeurIPS}, 2023.

\bibitem[Wang and Agapito(2025)]{spann3r}
Hengyi Wang and Lourdes Agapito.
\newblock 3d reconstruction with spatial memory.
\newblock In \emph{3DV}, 2025.

\bibitem[Wang et~al.(2025{\natexlab{a}})Wang, Chen, Karaev, Vedaldi, Rupprecht, and Novotny]{vggt}
Jianyuan Wang, Minghao Chen, Nikita Karaev, Andrea Vedaldi, Christian Rupprecht, and David Novotny.
\newblock Vggt: Visual geometry grounded transformer.
\newblock In \emph{CVPR}, 2025{\natexlab{a}}.

\bibitem[Wang et~al.(2024{\natexlab{a}})Wang, Kim, Yang, Yu, Ivanovic, Waslander, Wang, Fidler, Pavone, and Karkus]{distillnerf}
Letian Wang, Seung~Wook Kim, Jiawei Yang, Cunjun Yu, Boris Ivanovic, Steven~L. Waslander, Yue Wang, Sanja Fidler, Marco Pavone, and Peter Karkus.
\newblock Distillnerf: Perceiving 3d scenes from single-glance images by distilling neural fields and foundation model features.
\newblock In \emph{NeurIPS}, 2024{\natexlab{a}}.

\bibitem[Wang et~al.(2025{\natexlab{b}})Wang, Zhang, Holynski, Efros, and Kanazawa]{cuter}
Qianqian Wang, Yifei Zhang, Aleksander Holynski, Alexei~A Efros, and Angjoo Kanazawa.
\newblock Continuous 3d perception model with persistent state.
\newblock In \emph{CVPR}, 2025{\natexlab{b}}.

\bibitem[Wang et~al.(2024{\natexlab{b}})Wang, Leroy, Cabon, Chidlovskii, and Revaud]{duster}
Shuzhe Wang, Vincent Leroy, Yohann Cabon, Boris Chidlovskii, and Jerome Revaud.
\newblock Dust3r: Geometric 3d vision made easy.
\newblock In \emph{CVPR}, 2024{\natexlab{b}}.

\bibitem[Wang and Tong(2024)]{Wang2024H2GFormerHV}
Yu Wang and Chao Tong.
\newblock H2gformer: Horizontal-to-global voxel transformer for 3d semantic scene completion.
\newblock In \emph{AAAI}, 2024.

\bibitem[Wang et~al.(2025{\natexlab{c}})Wang, Chen, Yang, Wang, Zhang, Zhao, and Zhao]{wang2025depthprior}
Zehan Wang, Siyu Chen, Lihe Yang, Jialei Wang, Ziang Zhang, Hengshuang Zhao, and Zhou Zhao.
\newblock Depth anything with any prior.
\newblock In \emph{arXiv}, 2025{\natexlab{c}}.

\bibitem[Wei et~al.(2023)Wei, Zhao, Zheng, Zhu, Zhou, and Lu]{surroundocc}
Yi Wei, Linqing Zhao, Wenzhao Zheng, Zheng Zhu, Jie Zhou, and Jiwen Lu.
\newblock Surroundocc: Multi-camera 3d occupancy prediction for autonomous driving.
\newblock In \emph{ICCV}, 2023.

\bibitem[Wimbauer et~al.(2023)Wimbauer, Yang, Rupprecht, and Cremers]{bts}
Felix Wimbauer, Nan Yang, Christian Rupprecht, and Daniel Cremers.
\newblock Behind the scenes: Density fields for single view reconstruction.
\newblock In \emph{CVPR}, 2023.

\bibitem[Wimbauer et~al.(2025)Wimbauer, Chen, Muhle, Rupprecht, and Cremers]{AnyCam}
Felix Wimbauer, Weirong Chen, Dominik Muhle, Christian Rupprecht, and Daniel Cremers.
\newblock Anycam: Learning to recover camera poses and intrinsics from casual videos.
\newblock In \emph{CVPR}, 2025.

\bibitem[Wu et~al.(2024)Wu, Jiang, Wang, Liu, Liu, Qiao, Ouyang, He, and Zhao]{wu2024point}
Xiaoyang Wu, Li Jiang, Peng-Shuai Wang, Zhijian Liu, Xihui Liu, Yu Qiao, Wanli Ouyang, Tong He, and Hengshuang Zhao.
\newblock Point transformer v3: Simpler faster stronger.
\newblock In \emph{CVPR}, 2024.

\bibitem[Xia et~al.(2023)Xia, Liu, Li, Zhu, Ma, Li, Hou, and Qiao]{scpnet}
Zhaoyang Xia, Youquan Liu, Xin Li, Xinge Zhu, Yuexin Ma, Yikang Li, Yuenan Hou, and Yu Qiao.
\newblock Scpnet: Semantic scene completion on point cloud.
\newblock In \emph{CVPR}, 2023.

\bibitem[Xiao et~al.(2021)Xiao, Shao, Hao, Zhang, Chai, Jiao, Li, Wu, Sun, Jiang, et~al.]{pandaset}
Pengchuan Xiao, Zhenlei Shao, Steven Hao, Zishuo Zhang, Xiaolin Chai, Judy Jiao, Zesong Li, Jian Wu, Kai Sun, Kun Jiang, et~al.
\newblock Pandaset: Advanced sensor suite dataset for autonomous driving.
\newblock In \emph{ITSC}, 2021.

\bibitem[Xie et~al.(2025)Xie, Zhang, Wei, and Wu]{higaussian}
Binjian Xie, Pengju Zhang, Hao Wei, and Yihong Wu.
\newblock Hi-gaussian: Hierarchical gaussians under normalized spherical projection for single-view 3d reconstruction.
\newblock In \emph{ICCV}, 2025.

\bibitem[Xue et~al.(2025)Xue, Pi, Zhang, Qin, Tang, Li, and Li]{sdformer}
Yujie Xue, Huilong Pi, Jiapeng Zhang, Yunchuan Qin, Zhuo Tang, Kenli Li, and Ruihui Li.
\newblock Sdformer: Vision-based 3d semantic scene completion via sam-assisted dual-channel voxel transformer.
\newblock In \emph{ICCV}, 2025.

\bibitem[Yang et~al.(2025)Yang, Sax, Liang, Henaff, Tang, Cao, Chai, Meier, and Feiszli]{fast3r}
Jianing Yang, Alexander Sax, Kevin~J. Liang, Mikael Henaff, Hao Tang, Ang Cao, Joyce Chai, Franziska Meier, and Matt Feiszli.
\newblock Fast3r: Towards 3d reconstruction of 1000+ images in one forward pass.
\newblock In \emph{CVPR}, 2025.

\bibitem[Yang et~al.(2024)Yang, Kang, Huang, Zhao, Xu, Feng, and Zhao]{da2}
Lihe Yang, Bingyi Kang, Zilong Huang, Zhen Zhao, Xiaogang Xu, Jiashi Feng, and Hengshuang Zhao.
\newblock Depth anything v2.
\newblock In \emph{NeurIPS}, 2024.

\bibitem[Yao et~al.(2023)Yao, Li, Sun, Cai, Li, Ouyang, and Li]{ndcscene}
Jiawei Yao, Chuming Li, Keqiang Sun, Yingjie Cai, Hao Li, Wanli Ouyang, and Hongsheng Li.
\newblock Ndc-scene: Boost monocular 3d semantic scene completion in normalized device coordinates space.
\newblock In \emph{ICCV}, 2023.

\bibitem[Ye et~al.(2025)Ye, Qin, Zhang, Gong, Zhu, Zhao, and Zhao]{gsocc3d}
Baijun Ye, Minghui Qin, Saining Zhang, Moonjun Gong, Shaoting Zhu, Hao Zhao, and Hang Zhao.
\newblock Gs-occ3d: Scaling vision-only occupancy reconstruction with gaussian splatting.
\newblock In \emph{ICCV}, 2025.

\bibitem[Ye et~al.(2024)Ye, Jiang, Xu, Li, and Zhao]{cvtocc}
Zhangchen Ye, Tao Jiang, Chenfeng Xu, Yiming Li, and Hang Zhao.
\newblock Cvt-occ: Cost volume temporal fusion for 3d occupancy prediction.
\newblock In \emph{ECCV}, 2024.

\bibitem[Yu et~al.(2024)Yu, Zhang, Ying, Yu, Hu, Luo, Cao, and Shen]{CGFormer}
Zhu Yu, Runmin Zhang, Jiacheng Ying, Junchen Yu, Xiaohai Hu, Lun Luo, Si-Yuan Cao, and Hui-liang Shen.
\newblock Context and geometry aware voxel transformer for semantic scene completion.
\newblock In \emph{NeurIPS}, 2024.

\bibitem[Zhang et~al.(2025{\natexlab{a}})Zhang, Yan, Wei, Li, Liu, Tang, Duan, and Lu]{occnerf}
Chubin Zhang, Juncheng Yan, Yi Wei, Jiaxin Li, Li Liu, Yansong Tang, Yueqi Duan, and Jiwen Lu.
\newblock Occnerf: Advancing 3d occupancy prediction in lidar-free environments.
\newblock \emph{IEEE TIP}, 2025{\natexlab{a}}.

\bibitem[Zhang et~al.(2025{\natexlab{b}})Zhang, Herrmann, Hur, Jampani, Darrell, Cole, Sun, and Yang]{monst3r}
Junyi Zhang, Charles Herrmann, Junhwa Hur, Varun Jampani, Trevor Darrell, Forrester Cole, Deqing Sun, and Ming-Hsuan Yang.
\newblock Monst3r: A simple approach for estimating geometry in the presence of motion.
\newblock In \emph{ICLR}, 2025{\natexlab{b}}.

\bibitem[Zhang et~al.(2023)Zhang, Zhu, and Du]{OccFormer}
Yunpeng Zhang, Zheng Zhu, and Dalong Du.
\newblock Occformer: Dual-path transformer for vision-based 3d semantic occupancy prediction.
\newblock In \emph{ICCV}, 2023.

\bibitem[Zheng et~al.(2024)Zheng, Tang, Wang, Wang, Ren, Feng, and Ma]{veon}
Jilai Zheng, Pin Tang, Zhongdao Wang, Guoqing Wang, Xiangxuan Ren, Bailan Feng, and Chao Ma.
\newblock Veon: Vocabulary-enhanced occupancy prediction.
\newblock In \emph{ECCV}, 2024.

\bibitem[Zhou et~al.(2025)Zhou, Wang, Wang, Wei, Dong, and Yang]{zhou2025autoocc}
Xiaoyu Zhou, Jingqi Wang, Yongtao Wang, Yufei Wei, Nan Dong, and Ming-Hsuan Yang.
\newblock Autoocc: Automatic open-ended semantic occupancy annotation via vision-language guided gaussian splatting.
\newblock In \emph{ICCV}, 2025.

\bibitem[Zuo et~al.(2025)Zuo, Zheng, Han, Yang, Pan, and Lu]{zuo2025quadricformer}
Sicheng Zuo, Wenzhao Zheng, Xiaoyong Han, Longchao Yang, Yong Pan, and Jiwen Lu.
\newblock Quadricformer: Scene as superquadrics for 3d semantic occupancy prediction.
\newblock \emph{NeurIPS}, 2025.

\bibitem[Zust et~al.(2025)Zust, Cabon, Marrie, Antsfeld, Chidlovskii, Revaud, and Csurka]{panst3r}
Lojze Zust, Yohann Cabon, Juliette Marrie, Leonid Antsfeld, Boris Chidlovskii, Jerome Revaud, and Gabriela Csurka.
\newblock Panst3r: Multi-view consistent panoptic segmentation.
\newblock In \emph{ICCV}, 2025.

\end{thebibliography}
}

\appendix
\section{Additional Details}
\label{app:tech_details}
\subsection{Datasets}
\occDnuscenes{} was built upon nuScenes~\cite{nuscenes}.
It contains $1,000$ 20-sec sequences captured by one LiDAR and six surrounding cameras.
The dataset provides 3D occupancy annotations of $18$ semantic classes, with 0.4\,m voxels covering $80\times80\times6.4$\,m areas at the resolution of $200\times200\times16$ voxels.
Evaluation is done on the official \emph{val} split~\cite{occ3d} of $150$ sequences.

\semantickitti{}, based on KITTI~\cite{semkitti}, consists of 22 sequences.
Each sequence is annotated at the resolution of $256\times256\times32$ with 0.2\,m voxels and $21$ semantic classes (19 semantics, 1 free, 1 unknown).
In our experiments, we only use images from the \texttt{cam2} camera.
Following~\cite{selfocc, scenerf}, we evaluate on the \emph{val} set, \ie sequence 8.

\subsection{Training}
The \emph{3D Reconstruction} stage (\cf Sec.~\textcolor{cvprblue}{3.1}) is trained in two consecutive steps:
\begin{itemize}
    \item \emph{Sequence-only training.} We only use mono-view sequences from all cameras across the five datasets. Training samples are drawn from frames within the same mono-view sequences.
    \item \emph{Mixed training.} This step continues \emph{Sequence-only training} while mixing surround-view data with sequential data (from the previous step) at a $1:1$ ratio.
    For surround-view data, we use frames from different cameras captured at the same timestep.
\end{itemize}

\begin{table}[t!]
\scriptsize
\setlength{\tabcolsep}{0.0055\linewidth}
\centering
\resizebox{\columnwidth}{!}{
\begin{tabular}{lll|c|>{\columncolor{important}}c>{\columncolor{important}}c|c|>{\columncolor{important}}c>{\columncolor{important}}c}
\toprule
\multirow{2}{*}{\textbf{Method}} &
\multirow{2}{*}{\textbf{Sem. feat.}} &
\multirow{2}{*}{\textbf{Params}} &
\multicolumn{3}{c|}{\textbf{Semantic KITTI sequence}} &
\multicolumn{3}{c}{\textbf{Occ3D-NuScenes surround-view}} \\
\cmidrule(lr){4-6} \cmidrule(lr){7-9}
&&&\textbf{Res.} & \textbf{mIoU} & \textbf{mIoU$^{\rm sc}$} & \textbf{Res.} & \textbf{mIoU} & \textbf{mIoU$^{\rm sc}$} \\
\midrule
\textbf{OccAny} & Distilled & 623M & 512x160 & \textbf{7.28} & \textbf{13.53} & 512x288 &  6.66 & 10.32 \\
\textbf{OccAny+} &  Distilled & 651M & 512x160 & 6.48 & 13.30 & 512x288 & \textbf{7.20} & \textbf{11.50} \\
\midrule
\textbf{OccAny} & Pretrained & 864M & 512x160 & 7.67 & 13.75 & 512x288 & 7.42 & 10.78 \\
\textbf{OccAny+}  &  Pretrained & 1.08B &  512x160 & \textbf{8.03} & \textbf{13.17} & 512x288 & \textbf{9.45} & \textbf{12.22} \\
\bottomrule
\end{tabular}}

\caption{\textbf{Using pretrained segmentation features} to boost semantic performance. OccAny+ is the variant using DA3 and SAM3 base models. Parameter counts reflect the forward path from the input to the predicted pointmaps and segmentation features. Note that using "pretrained" semantic features incurs a higher parameter cost due to the use of pretrained encoder.}
\label{tbl:boost_sem_perf}
\end{table}

The \emph{Novel-View Rendering} stage (\cf Sec.~\textcolor{cvprblue}{3.2}) is trained exclusively on sequential data. Empirically, we observed no gains when incorporating surround-view data in this stage.

Each stage is trained for $100$~epochs using the AdamW optimizer~\cite{adamw} with a learning rate of $7 \times 10^{-5}$. We utilize a cosine scheduler with a minimum learning rate of $1\times10^{-6}$ and a $3$-epoch warmup. The training set consists of $50,000$~samples (sequences or sets of surrounding images), with $10,000$~drawn from each dataset. Experiments are conducted on $16$ NVIDIA A100 40GB GPUs with an effective batch size of $64$. The \emph{3D Reconstruction} and \emph{Novel-View Rendering} stages required approximately $40$ and $30$~training hours, respectively.

\subsection{OccAny+ using DA3 and SAM3}
\label{supp:occanyplus}
For the \emph{3D Reconstruction} phase, we substitute the reconstruction encoder $\reconEncoder$ and decoder $\reconDecoder$ from \muster{} with DA3 backbone, fine-tuning the final eight transformer layers and the dual DPT head. For novel view rendering, we utilize the same projection and tokenization layers but replace both the rendering encoder $\renderEncoder$ and decoder $\renderDecoder$ with DA3 backbone.

To leverage the strong initialization of the pretrained DA3 model, we introduce a self-distillation branch that duplicates the last eight transformer layers. These duplicated layers serve as a "teacher", supervising the eight trainable transformer layers via a scale-invariant loss, matching the scale-invariant output of the pretrained DA3.

In the \emph{Novel View Rendering} phase, DA3 is initialized with the weights from the reconstruction phase. 
We train the first eight transformer layers while freezing the rest. 
Because DA3 lacks a memory mechanism, we tokenize the reconstruction outputs (pointmap, confidence, RGB, and segmentation features) from all reconstructed views using the same tokenizer; these are passed alongside the novel-view tokens $\{\novelInputfeatures_j\}_{j=1}^{\nren}$.
To facilitate cross-view information exchange, we modify the attention mechanism to alternate between global and local attention starting from the first layer, rather than the eighth layer as in the original DA3.

Regarding \emph{Segmentation Forcing}, we replaced the SAM2 encoder with a SAM3 encoder.
Our findings indicate that performance improves significantly when the linear head is replaced with a DPTHead, particularly when trained using a $10\times$ higher learning rate.
In all experiments, we use the DA3-LARGE variant.

\begin{table}[t!]
\centering
\setlength{\tabcolsep}{3pt}
\resizebox{\columnwidth}{!}{%
\begin{tabular}{ccl|ccccccc|>{\columncolor{important}}c >{\columncolor{important}}c}
\toprule
& & \textbf{Method}
& \rot{\textbf{Extr.}}
& \rot{\textbf{Intr.}}
& \rot{\textbf{\shortstack{Fixed\\Ratio}}}
& \rot{\textbf{\shortstack{Fixed\\Rig}}}
& \rot{\textbf{{\shortstack{GT\\LiDAR}}}}
& \rot{\textbf{\shortstack{Sem.\\Adapt.}}}
& \rot{\textbf{\shortstack{GT\\Occ.}}}
& \textbf{IoU}
& \textbf{mIoU}\\
\midrule
\multirow{15}{*}{\rot{{\textbf{\occDnuscenes} (ext.~\cref{tab:surround-occ3d-nuscenes})}}}&
\multirow{8}{*}{\rot{\textcolor{plt:red}{\textbf{in-domain}}}}
& SimpleOcc
& \textcolor{plt:red}{\tiny Req.}
& \textcolor{plt:red}{\tiny Req.}
& \textcolor{plt:red}{\tiny Req.}
& \textcolor{plt:red}{\tiny Req.}
& \textcolor{plt:red}{\tiny Req.}
& \textcolor{plt:red}{\tiny Req.}
& \textcolor{plt:green}{--}
& 33.92 & 7.05\\

&& DistillNeRF
& \textcolor{plt:red}{\tiny Req.}
& \textcolor{plt:red}{\tiny Req.}
& \textcolor{plt:red}{\tiny Req.}
& \textcolor{plt:red}{\tiny Req.}
& \textcolor{plt:red}{\tiny Req.}
& \textcolor{plt:green}{--}
& \textcolor{plt:green}{--}
& 29.11 & 8.93\\

&&  SelfOcc
& \textcolor{plt:red}{\tiny Req.}
& \textcolor{plt:red}{\tiny Req.}
& \textcolor{plt:red}{\tiny Req.}
& \textcolor{plt:red}{\tiny Req.}
& \textcolor{plt:green}{--}
& \textcolor{plt:red}{\tiny Req.}
& \textcolor{plt:green}{--}
& 45.01 & 9.30 \\

&& POP-3D
& \textcolor{plt:red}{\tiny Req.}
& \textcolor{plt:red}{\tiny Req.}
& \textcolor{plt:red}{\tiny Req.}
& \textcolor{plt:red}{\tiny Req.}
& \textcolor{plt:red}{\tiny Req.}
& \textcolor{plt:red}{\tiny Req.}
& \textcolor{plt:green}{--}
& 28.17 & 9.31\\

&& OccNeRF
& \textcolor{plt:red}{\tiny Req.}
& \textcolor{plt:red}{\tiny Req.}
& \textcolor{plt:red}{\tiny Req.}
& \textcolor{plt:red}{\tiny Req.}
& \textcolor{plt:green}{--}
& \textcolor{plt:red}{\tiny Req.}
& \textcolor{plt:green}{--}
& 39.20 & 9.53\\

&& GaussianOcc
& \textcolor{plt:green}{--}
& \textcolor{plt:red}{\tiny Req.}
& \textcolor{plt:red}{\tiny Req.}
& \textcolor{plt:red}{\tiny Req.}
& \textcolor{plt:green}{--}
& \textcolor{plt:red}{\tiny Req.}
& \textcolor{plt:green}{--}
& 51.22 & 9.94\\

&& VEON
& \textcolor{plt:red}{\tiny Req.}
& \textcolor{plt:red}{\tiny Req.}
& \textcolor{plt:red}{\tiny Req.}
& \textcolor{plt:red}{\tiny Req.}
& \textcolor{plt:red}{\tiny Req.}
& \textcolor{plt:red}{\tiny Req.}
& \textcolor{plt:red}{\tiny Req.}
& 57.92 & 12.38\\

&& GaussTR
& \textcolor{plt:red}{\tiny Req.}
& \textcolor{plt:red}{\tiny Req.}
& \textcolor{plt:red}{\tiny Req.}
& \textcolor{plt:red}{\tiny Req.}
& \textcolor{plt:green}{--}
& \textcolor{plt:red}{\tiny Req.}
& \textcolor{plt:green}{--}
& 45.19 & 12.27 \\

\cmidrule(lr){2-12}
&\multirow{7}{*}{\rot{\textcolor{plt:green}{\textbf{out-of-domain
}}}}
& MUSt3R
& \textcolor{plt:green}{--}
& \textcolor{plt:green}{--}
& \textcolor{plt:green}{--}
& \textcolor{plt:green}{--}
& \textcolor{plt:green}{--}
& \textcolor{plt:green}{--}
& \textcolor{plt:green}{--}
& 13.61 & 2.43\\

&& \cuter{}*
& \textcolor{plt:green}{--}
& \textcolor{plt:green}{--}
& \textcolor{plt:green}{--}
& \textcolor{plt:green}{--}
& \textcolor{plt:green}{--}
& \textcolor{plt:green}{--}
& \textcolor{plt:green}{--}
& 19.21 & 3.06\\

&& VGGT$^{\dagger}$
& \textcolor{orange}{\tiny Rescale}
& \textcolor{orange}{\tiny Rescale}
& \textcolor{plt:green}{--}
& \textcolor{plt:green}{--}
& \textcolor{plt:green}{--}
& \textcolor{plt:green}{--}
& \textcolor{plt:green}{--}
& 20.42 & 4.39\\

&& AnySplat*$^{\dagger}$
& \textcolor{orange}{\tiny Rescale}
& \textcolor{orange}{\tiny Rescale}
& \textcolor{plt:green}{--}
& \textcolor{plt:green}{--}
& \textcolor{plt:green}{--}
& \textcolor{plt:green}{--}
& \textcolor{plt:green}{--}
& 20.78 & 4.44\\

&& DA3
& \textcolor{plt:green}{--}
& \textcolor{plt:green}{--}
& \textcolor{plt:green}{--}
& \textcolor{plt:green}{--}
& \textcolor{plt:green}{--}
& \textcolor{plt:green}{--}
& \textcolor{plt:green}{--}
& 19.65 & 4.55\\

&& \textbf{\OURS{}}
& \textcolor{plt:green}{--}
& \textcolor{plt:green}{--}
& \textcolor{plt:green}{--}
& \textcolor{plt:green}{--}
& \textcolor{plt:green}{--}
& \textcolor{plt:green}{--}
& \textcolor{plt:green}{--}
& \best{34.10} & \second{6.62}\\

&& \textbf{\OURS{}+ (Pretrained)}
& \textcolor{plt:green}{--}
& \textcolor{plt:green}{--}
& \textcolor{plt:green}{--}
& \textcolor{plt:green}{--}
& \textcolor{plt:green}{--}
& \textcolor{plt:green}{--}
& \textcolor{plt:green}{--}
& \second{33.49} & \textbf{9.45}\\
\bottomrule
\multicolumn{12}{l}{\small{*: use \ttva{}\,\,\,\,\,$^\dagger$: scaled with Metric3Dv2~\cite{metric3d}.}}\\
\multicolumn{12}{l}{\small{\textcolor{plt:red}{Req.}: required in-domain data/priors. \textcolor{orange}{Rescale}: metric scaling needed}}\\
\end{tabular}}
\caption{\textbf{Detailed \surround{} results.} OccAny+ is the variant using DA3 and SAM3 base models.}
\label{tab:additional_method_comparison}
\end{table}

\section{Supplementary Studies}
\label{app:more_studies}
We present here the supplementary studies not presented in the main text due to the lack of space.

\begin{table*}
	\small
	\setlength{\tabcolsep}{0.0055\linewidth}
	\centering
    \begin{tabular}{l|c|cc >{\columncolor{important}}c|cc >{\columncolor{important}}c|c|cc >{\columncolor{important}}c|cc >{\columncolor{important}}c}
        \toprule
        \multirow{3}{*}{\textbf{Method}} &  \multicolumn{7}{c|}{\textbf{Semantic KITTI}} & \multicolumn{7}{c}{\textbf{Occ3D-NuScenes}} \\
        \cmidrule(lr){2-8} \cmidrule(lr){9-15} 
        & \multirow{2}{*}{\textbf{Res.}} & \multicolumn{3}{c|}{\textit{\sequence{}}} & \multicolumn{3}{c|}{\textit{\singleview{}}}&
        \multirow{2}{*}{\textbf{Res.}} & \multicolumn{3}{c|}{\textit{\sequence{}}}& \multicolumn{3}{c}{\textit{\surround{}}} \\
        \cmidrule(lr){3-5} \cmidrule(lr){6-8} \cmidrule(lr){10-12}\cmidrule(lr){13-15}
        & & \textbf{Prec.} & \textbf{Rec.} & \textbf{IoU}  & \textbf{Prec.} & \textbf{Rec.} & \textbf{IoU} &  & \textbf{Prec.} & \textbf{Rec.} & \textbf{IoU} & \textbf{Prec.} & \textbf{Rec.} & \textbf{IoU}  \\
        \midrule
        
        \OURS{}$_\text{depth completion}$ & $512\times160$ & 24.59 & 44.55 & 18.82  & 21.59 & \best{37.55} & 15.89  & $512\times288$ & 29.80, & 36.09 & 19.51 & 30.57 & 39.32 & 20.77 \\
        \textbf{\OURS{}} & $512\times160$ &  \best{36.79} & \best{46.79} & \best{25.91} & \best{45.64} & {33.66} & \best{24.03}   & $512\times288$ &  \best{36.09} & \best{40.39} & \best{23.55}  & \best{45.04} & \best{58.54} & \best{34.15}\\
        \bottomrule
    \end{tabular}
	\caption{\textbf{Novel-View Rendering \vs Depth Completion.} Occupancy prediction results on SemanticKITTI and Occ3D-NuScenes show the effectiveness of \emph{Novel-View Rendering}.}
	\label{tbl:nvr_depthcompletion}
\end{table*}  

\begin{table*}
	\small
	\centering
    \setlength{\tabcolsep}{0.0055\linewidth}
    \begin{tabular}{l|l|c|c|cc >{\columncolor{important}}c|c|cc >{\columncolor{important}}c}
        \toprule
        \multirow{2}{*}{\textbf{Label}} & \multirow{2}{*}{\textbf{Method}} & \multirow{2}{*}{\textbf{Venue}} & \multicolumn{4}{c|}{\textbf{\occDnuscenes{} \surround{}}} & \multicolumn{4}{c}{\textbf{\semantickitti{} \singleview{} }} \\
        \cmidrule(lr){4-7} \cmidrule(lr){8-11} 
        &  &  & \textbf{Res.} & \textbf{Prec.} & \textbf{Rec.} & \textbf{IoU} &  \textbf{Res.} & \textbf{Prec.} & \textbf{Rec.} & \textbf{IoU} \\
        \midrule
        \emph{Occ} & CVT-Occ~\cite{cvtocc} (Trained on \occDwaymo{}) & ECCV'24 &  $1600\times900$ &  \second{35.38}  & 25.86 & \second{17.56} & $1220\times370$ & 8.97 &  34.92 & 7.69  \\
        & CVT-Occ~\cite{cvtocc} (Trained on \occDwaymo{}) & ECCV'24 &  $960\times 540$ & 29.15   &  \second{28.33} & 16.78 &   $960\times 292$ & 8.92 &  36.84 & 7.73 \\
        & CVT-Occ~\cite{cvtocc} (Trained on \occDnuscenes{}) & ECCV'24 & \indomain{in-domain} & \indomain{--} & \indomain{--} & \indomain{--}  & $1220\times370$ & 11.73 & \best{59.97} & 9.43 \\
        & ALOcc~\cite{alocc} (Trained on \occDnuscenes{}) & ICCV'25 & \indomain{in-domain} & \indomain{--} & \indomain{--} & \indomain{--}  & $704\times256$ & \second{16.34} & \second{53.06} & \second{14.28} \\
        \emph{LiDAR} & \textbf{\OURS{}} & -- & $512\times288$ & \best{45.04} & \best{58.54} & \best{34.15} & $512\times160$ & \best{45.64} & 33.66 & \best{24.03} \\
        \bottomrule
    \end{tabular}
	\caption{\textbf{Generalization results of fully-supervised methods.} \emph{Occ} label is denser through temporal accumulation of LiDAR point-clouds and subsequent post-processing, whereas the \emph{LiDAR} label remains sparser at each timestep.
    \OURS{} works out of the box in any evaluation settings with different inference areas, voxel resolutions and sensor configurations.
    In contrast, other methods require manual code modifications to align testing and training conditions.
    Beyond being more versatile, \OURS{} clearly demonstrates superior generalization.
    }
	\label{tab:3D-ood-results}
\end{table*}

\subsection{Boosting semantic performance.}
\label{app:boost_sem_perf}
While the unified OccAny model conveniently uses distilled segmentation features, it can also be combined with the original features from segmentation foundation models at inference. Although this introduces additional overhead, it enables the use of higher-resolution segmentation features and improves semantic performance, as shown in~\cref{tbl:boost_sem_perf}.

\subsection{More \surround{} results}
\cref{tab:additional_method_comparison} details results and method constraints in the \surround{} setting, further including POP-3D~\cite{pop3d}, GaussianOcc~\cite{GaussianOcc}, and VEON~\cite{veon}.
Existing in-domain approaches, including self-supervised ones, rely heavily on domain-specific priors, and VEON further depends on binary occupancy ground truth for training.
In contrast, \OURS{} promotes a paradigm shift toward generalized and unconstrained occupancy prediction, enabling deployment of a \emph{unified model} across out-of-domain and heterogeneous sensor setups. Beyond being unconstrained, \OURS{} can benefit from continual advances in foundation models, and is therefore expected to progressively narrow the remaining performance gap.

As preliminary evidence, upgrading \muster{} to DA3 and replacing SAM2 with the more recent SAM3 yields an mIoU improvement of approximately 3 points, reaching performance comparable to recent self-supervised methods such as GaussianOcc~\cite{GaussianOcc}.

\subsection{Novel-View Rendering \vs Depth Completion}
In this experiment, we compare the effectiveness of our \emph{Novel-View Rendering} stage (\cf Sec.~\textcolor{cvprblue}{3.2}) with a baseline that performs depth completion on the projected pointmaps of the novel views.
To this end, we replace \emph{Novel-View Rendering} by using Prior Depth Anything~\cite{wang2025depthprior}, which takes as input the sparse projected pointmaps and the rendered RGB images produced by the state-of-the-art novel-view synthesis method AnySplat~\cite{anysplat}.
The Prior Depth Anything model outputs dense, completed depth maps for the novel views.
We name this baseline \OURS{}$_\text{depth completion}$ and present comparison results in~\cref{tbl:nvr_depthcompletion}.
Both models start from the first-stage-only \OURS{} and both adopt the TTVA strategy.
\OURS{} significantly outperforms the \OURS{}$_\text{depth completion}$ baseline, validating the effectiveness of our second stage.

\subsection{Generalization of State-of-the-art (SOTA) 3D Supervised Occupancy Models}

We assess the generalization capability of SOTA 3D fully-supervised models by evaluating models \emph{trained on a source dataset} directly on \emph{a different target dataset}.
We evaluate two settings: 
\begin{itemize}
    \item \occDwaymo{}$\rightarrow{}$\occDnuscenes{} (\surround{} $\rightarrow{}$\surround{}).
    \item \occDnuscenes{}/\occDwaymo{}$\rightarrow{}$\semantickitti{} (\surround{}$\rightarrow{}$\singleview{}).
\end{itemize}
As shown in Table~\ref{tab:3D-ood-results}, despite careful alignment of sensor configurations, inference areas, and voxel resolutions, these supervised methods exhibit limited generalization capabilities compared to \OURS{}.
Notably, \OURS{}'s inference is straightforward and does not require any prior knowledge of the sensor configurations (number of cameras, intrinsics/extrinsics and camera poses), adapting effortlessly to any inference areas and any voxel resolutions.

\condenseparagraph{\occDwaymo{}$\rightarrow$\occDnuscenes{}.}
In this setting, we evaluate CVT-Occ~\cite{cvtocc} using weights trained on \occDwaymo{} to perform inference on \occDnuscenes{}.
While the voxel resolutions and voxel sizes are consistent between these datasets, significant differences remain in sensor configurations.
To enable inference, we align the sensor setups by mapping the five \occDwaymo{} cameras to the six \occDnuscenes{} cameras.
Specifically, we map the \occDwaymo{} Front, Front-Right, and Front-Left to their \occDnuscenes{} counterparts, while the \occDwaymo{} Side-Left is mapped to both Back and Back-Left, and Side-Right to Back-Right.
Regarding image resolution, we follow the official implementation to scale \occDnuscenes{} input images to the \occDwaymo{} training resolution of $960\times640$. We also report inference performance at $1600\times900$, which yields slightly better results.
However, as detailed in Table~\ref{tab:3D-ood-results}, even with these manual adaptations, the model struggles to generalize to the new domain, achieving a peak IoU of only $17.56\%$, significantly lower than the $34.15\%$ achieved by our method.

\condenseparagraph{\occDnuscenes{}/\occDwaymo{}$\rightarrow$\semantickitti{}.}
Regarding the transfer from \surround{} to \singleview{}, we evaluate two SOTA 3D supervised methods: CVT-Occ~\cite{cvtocc} and ALOcc~\cite{alocc}.
We use checkpoints trained on \occDnuscenes{} (for both ALOcc and CVT-Occ) and \occDwaymo{} (for CVT-Occ) to perform inference on \semantickitti{}.
This scenario presents a significantly greater challenge than the previous setting: in addition to domain shifts and sensor discrepancies (using only the source front camera to align with the target setup), there are substantial divergences in voxel grid extents and resolutions.

For CVT-Occ~\cite{cvtocc}, we use two provided models, one trained on \occDnuscenes{}~($1600\times900$) and another trained on \occDwaymo{}~($960\times640$).
We evaluate the \occDnuscenes{}-trained model on \semantickitti{} at full image resolution~($1220\times370$), as it is closely aligned with the training resolution.
For the \occDwaymo{}-trained model, we conduct evaluations at both the full resolution and a resized resolution of $960\times540$, which preserves the \semantickitti{} aspect ratio while approximating the source training resolution.

For ALOcc~\cite{alocc}, only the model trained on \occDnuscenes{} is available.
Since ALOcc encodes the stereo cost volume's frustum grid within its parameters, the network is constrained to a fixed input resolution of $704\times256$.
Consequently, we evaluate ALOcc on \semantickitti{} at this exact resolution, adhering to the official implementation by using $16$ history frames and pairs of consecutive timesteps as stereo input.

The results in Table~\ref{tab:3D-ood-results} highlight a significant drop in performance when these models are inferred on the unseen \semantickitti{} dataset.
CVT-Occ and ALOcc achieve IoUs of only $9.43\%$ and $14.28\%$, respectively, whereas our proposed method demonstrates superior robustness with an IoU of $24.03\%$.

\subsection{Ego Vehicle Trajectory Prediction}
We assess the quality of ego-trajectory prediction using OccAny on the nuScenes validation set, following the evaluation protocol of~\cite{vista, gem}.
OccAny+ outperforms the base DA3-LARGE model in terms of Average Displacement Error (ADE), demonstrating clear advantages in urban scenes.
Furthermore, it approaches the accuracy of optimization-based RGB-D SLAM methods while remaining fully feed-forward and significantly simpler.

\begin{table}[h]
    \centering
    \resizebox{1.0\columnwidth}{!}{%
    \begin{tabular}{llc}
        \toprule
        Method  & \textbf{ADE (m)} \\ 
        \midrule
        GeoCalib~\cite{geocalib} + DroidSLAM~\cite{droidslam} + DA2~\cite{da2} & 1.63 \\
        DA3 large + DA3 metric large                               & 2.44 \\
        OccAny+    & 1.86 \\ 
        \bottomrule
    \end{tabular}
    }
    \caption{\textbf{Ego Vehicle Trajectory Prediction.}}
    \label{tab:traj_comparison}
\end{table}

\subsection{NVR complexity.}
We report in~\cref{tab:frames_time_memory} the memory consumption and running time of NVR inference using one A100 GPU in the \surround{} setting.
Similar to VGGT (\cf Tab.~\textcolor{cvprblue}{9}~in~\cite{vggt}), both memory \& time scale much slower \emph{w.r.t.} number of augmentation frames.

\begin{table}[t!]
\centering
\setlength{\tabcolsep}{3pt}
\resizebox{1.0\columnwidth}{!}{%
\begin{tabular}{lcccccccccc}
\hline
\textbf{\#Aug. Frames} & 1 & 2 & 4 & 8 & 10 & 20 & 50 & 100 & 200 \\
\hline
Time (s)   & 0.052 & 0.057 & 0.085  & 0.172 & 0.227 & 0.542 & 1.406 & 2.786 & 5.572 \\
Mem. (GB) & 1.175  & 1.920  & 3.422   & 6.471  & 8.005  & 9.936  & 14.314 & 17.013 & 22.418 \\
\hline
\end{tabular}%
}
\caption{\textbf{NVR inference complexity}, measured on one A100 GPU.}
\label{tab:frames_time_memory}
\end{table}

\begin{table}[t!]
\centering
\setlength{\tabcolsep}{4pt}
\resizebox{\columnwidth}{!}{
\begin{tabular}{lcccccc}
\toprule
\textbf{Method} &
\textbf{Train. GPUs} &
\textbf{Train. time} &
\textbf{Recon. time (ms)} &
\textbf{Render time (ms)} &
\textbf{Params (M)} \\
\midrule
CUT3R    & 8$\times$A100   & $\approx$30 days & 240.0 & 259.8 & 793.3 \\
VGGT     & 64$\times$A100  & $>$9 days        & 222.2 & --    & 1157.9 \\
AnySplat & 16$\times$A800  & $\approx$2 days  & 251.7 & 17.2  & 1190.7 \\
\OURS{}  & 16$\times$A100  & $\approx$1.5 days & 93.8 & 123.2 & 651.1 \\
\bottomrule
\end{tabular}
}
\caption{\textbf{Model size and speed.} Train times are from the original papers. Inference times are measured in the surround setting with 6 input views and 6 render views.}
\label{tab:measure_time}
\end{table}

\subsection{Model sizes and speeds}
We are report the model sizes and speeds of \OURS{} and baselines in~\cref{tab:measure_time}. \OURS{} has the fewest parameters~($\sim$651M) \vs CUT3R~($\sim$793M) and VGGT/AnySplat~($\sim$1.2B), and is the most runtime efficient in training/inference.
\OURS{}'s rendering is about $2\times$ faster than CUT3R, while AnySplat's is the fastest thanks to 3DGS.

\section{Qualitative Examples}
\label{app:supp}
\begin{figure*}[ht!]
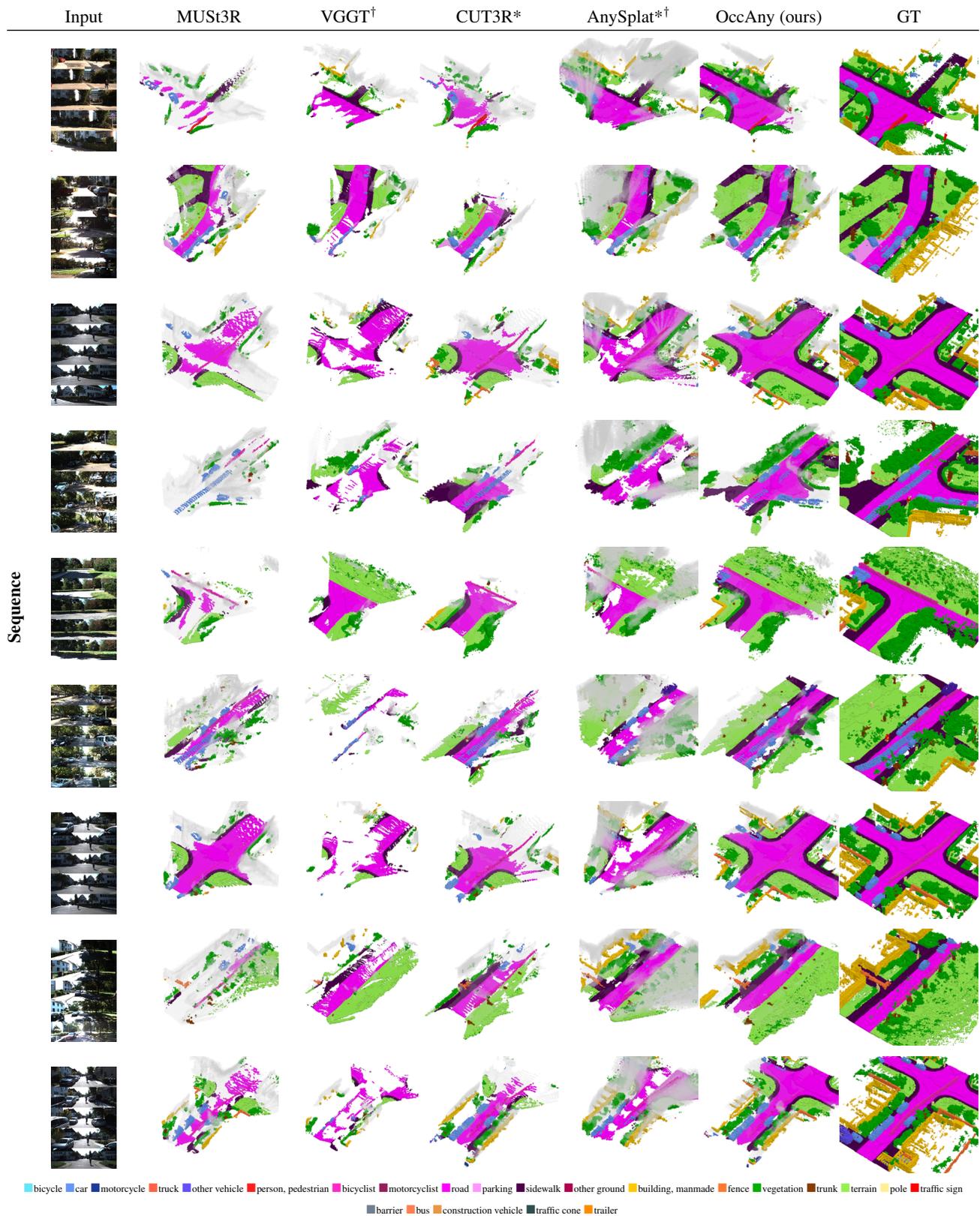

	\centering
	\small
	\setlength{\tabcolsep}{0.5pt}
	\renewcommand{\arraystretch}{0}
	\newcommand{\imgsizer}{.14\textwidth}
	\newcommand{\trimWidth}{0.45}
	\newcommand{\trimHeight}{0.42}
	\newcommand{\trimLeft}{0.1}
	\newcommand{\trimBot}{0.05}
	\newcommand{\kittiOccRow}[2]{%
        \makecell[t]{\adjincludegraphics[width=.6\linewidth]{figs/kitti_5frames_#2opa/#1_stacked_recon.png}} &
        \makecell{\adjincludegraphics[width=\linewidth,trim={\trimLeft\width} {\trimBot\height} {\trimWidth\width} {\trimHeight\height}, clip]{figs/kitti_5frames_#2opa/#1_must3r.png}} &
        \makecell{\adjincludegraphics[width=\linewidth,trim={\trimLeft\width} {\trimBot\height} {\trimWidth\width} {\trimHeight\height}, clip]{figs/kitti_5frames_#2opa/#1_VGGT.png}} &
        \makecell{\adjincludegraphics[width=\linewidth,trim={\trimLeft\width} {\trimBot\height} {\trimWidth\width} {\trimHeight\height}, clip]{figs/kitti_5frames_#2opa/#1_cut3r.png}} &
        \makecell{\adjincludegraphics[width=\linewidth,trim={\trimLeft\width} {\trimBot\height} {\trimWidth\width} {\trimHeight\height}, clip]{figs/kitti_5frames_#2opa/#1_AnySplat.png}} &
        \makecell{\adjincludegraphics[width=\linewidth,trim={\trimLeft\width} {\trimBot\height} {\trimWidth\width} {\trimHeight\height}, clip]{figs/kitti_5frames_#2opa/#1_occany512_original.png}} &
        \makecell{\adjincludegraphics[width=\linewidth,trim={\trimLeft\width} {\trimBot\height} {\trimWidth\width} {\trimHeight\height}, clip]{figs/kitti_5frames_#2opa/#1_ground_truth.png}}%
    }

	\newcommand{\trimNuScenesWidth}{0.35}
	\newcommand{\trimNuScenesHeight}{0.34}
	\newcommand{\trimNuScenesLeft}{0.22}
	\newcommand{\trimNuScenesBot}{0.26}

    \newcommand{\occRow}[2]{%
        \makecell{\adjincludegraphics[width=\linewidth]{figs/#2/#1_stacked_recon.png}} &
        \makecell{\adjincludegraphics[width=\linewidth,trim={\trimNuScenesLeft\width} {\trimNuScenesBot\height} {\trimNuScenesWidth\width} {\trimNuScenesHeight\height}, clip]{figs/#2/#1_must3r.png}} &
        \makecell{\adjincludegraphics[width=\linewidth,trim={\trimNuScenesLeft\width} {\trimNuScenesBot\height} {\trimNuScenesWidth\width} {\trimNuScenesHeight\height}, clip]{figs/#2/#1_VGGT.png}} &
        \makecell{\adjincludegraphics[width=\linewidth,trim={\trimNuScenesLeft\width} {\trimNuScenesBot\height} {\trimNuScenesWidth\width} {\trimNuScenesHeight\height}, clip]{figs/#2/#1_cut3r.png}} &
        \makecell{\adjincludegraphics[width=\linewidth,trim={\trimNuScenesLeft\width} {\trimNuScenesBot\height} {\trimNuScenesWidth\width} {\trimNuScenesHeight\height}, clip]{figs/#2/#1_AnySplat.png}} &
        \makecell{\adjincludegraphics[width=\linewidth,trim={\trimNuScenesLeft\width} {\trimNuScenesBot\height} {\trimNuScenesWidth\width} {\trimNuScenesHeight\height}, clip]{figs/#2/#1_occany512_original.png}} &
        \makecell{\adjincludegraphics[width=\linewidth,trim={\trimNuScenesLeft\width} {\trimNuScenesBot\height} {\trimNuScenesWidth\width} {\trimNuScenesHeight\height}, clip]{figs/#2/#1_ground_truth.png}}
    }
    
    \begin{tabular}{m{0.02\textwidth} m{0.11\textwidth}m{\imgsizer}m{\imgsizer}m{\imgsizer}m{\imgsizer}m{\imgsizer}m{\imgsizer}}
        & \multicolumn{1}{c}{Input\vphantom{$^\dagger$}} & \multicolumn{1}{c}{\muster{}\vphantom{$^\dagger$}} & \multicolumn{1}{c}{VGGT$^\dagger$} & \multicolumn{1}{c}{\cuter{}*\vphantom{$^\dagger$}} & \multicolumn{1}{c}{{AnySplat*}$^\dagger$} & \multicolumn{1}{c}{\OURS{} (ours)\vphantom{$^\dagger$}} & \multicolumn{1}{c}{GT\vphantom{$^\dagger$}} \\
        \midrule
        \multirow{12}{*}{\rotatebox[origin=c]{90}{\parbox{18cm}{\centering\textbf{\Sequence{}}}}} & \kittiOccRow{08_003720}{3} \\
        & \kittiOccRow{08_002835}{3} \\
        & \kittiOccRow{08_001275}{3} \\
        & \kittiOccRow{08_003045}{3} \\
        & \kittiOccRow{08_001565}{3} \\
        & \kittiOccRow{08_002960}{3} \\
        & \kittiOccRow{08_001265}{3} \\
        &\kittiOccRow{08_001670}{3} \\
        &\kittiOccRow{08_001235}{3} \\
    \end{tabular}
    {				
        \tiny
        \textcolor{bicycle}{$\blacksquare$}bicycle~%
        \textcolor{car}{$\blacksquare$}car~%
        \textcolor{motorcycle}{$\blacksquare$}motorcycle~%
        \textcolor{truck}{$\blacksquare$}truck~%
        \textcolor{other-vehicle}{$\blacksquare$}other vehicle~%
        \textcolor{person}{$\blacksquare$}person, pedestrian~%
        \textcolor{bicyclist}{$\blacksquare$}bicyclist~%
        \textcolor{motorcyclist}{$\blacksquare$}motorcyclist~%
        \textcolor{road}{$\blacksquare$}road~%
        \textcolor{parking}{$\blacksquare$}parking~%
        \textcolor{sidewalk}{$\blacksquare$}sidewalk~%
        \textcolor{other-ground}{$\blacksquare$}other ground~%
        \textcolor{building}{$\blacksquare$}building, manmade~%
        \textcolor{fence}{$\blacksquare$}fence~%
        \textcolor{vegetation}{$\blacksquare$}vegetation~%
        \textcolor{trunk}{$\blacksquare$}trunk~%
        \textcolor{terrain}{$\blacksquare$}terrain~%
        \textcolor{pole}{$\blacksquare$}pole~%
        \textcolor{traffic-sign}{$\blacksquare$}traffic sign~%
        
        \textcolor{barrier}{$\blacksquare$}barrier
        \textcolor{bus}{$\blacksquare$}bus
        \textcolor{construction-vehicle}{$\blacksquare$}construction vehicle
        \textcolor{traffic-cone}{$\blacksquare$}traffic cone
        \textcolor{trailer}{$\blacksquare$}trailer
		 }
	\caption{\textbf{Occupancy predictions} of \OURS{} and baselines on sequential data. We visualize here predicted voxels. For qualitative analysis, we overlay the semantic ground-truth colors on predicted voxels to better highlight class-wise gains. False positive voxels are painted in gray without any overlayed color. Compared to baselines, our occupancy predictions are denser and more accurate.}
	
	\label{fig:supp_qual_res_sequence}
\end{figure*}

\begin{figure*}[ht!]
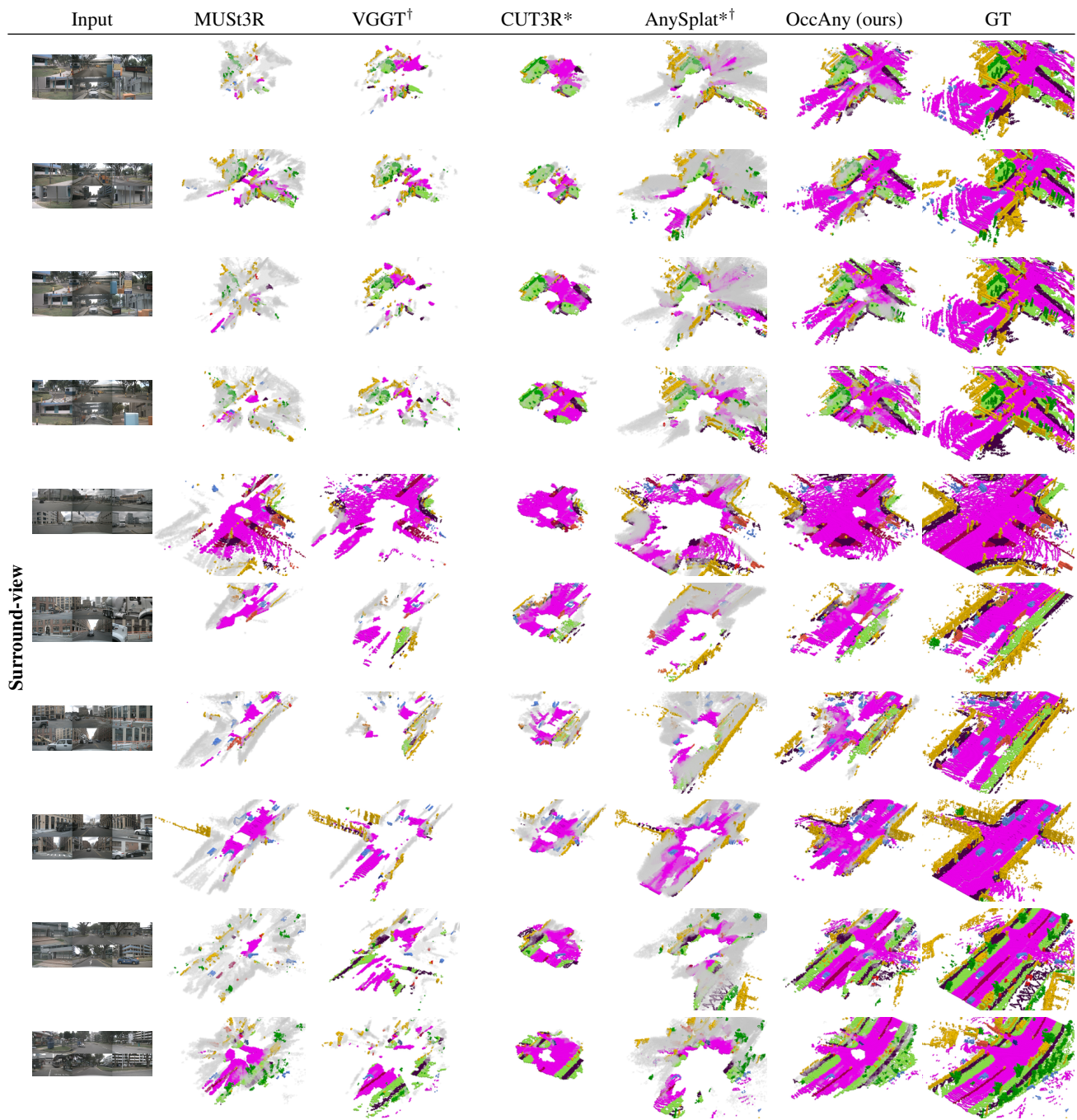

	\centering
	\small
	\setlength{\tabcolsep}{0.5pt}
	\renewcommand{\arraystretch}{0}
	\newcommand{\imgsizer}{.14\textwidth}
	\newcommand{\trimWidth}{0.45}
	\newcommand{\trimHeight}{0.42}
	\newcommand{\trimLeft}{0.1}
	\newcommand{\trimBot}{0.05}
	\newcommand{\kittiOccRow}[2]{%
        \makecell[t]{\adjincludegraphics[width=.6\linewidth]{figs/kitti_5frames_#2opa/#1_stacked_recon.png}} &
        \makecell{\adjincludegraphics[width=\linewidth,trim={\trimLeft\width} {\trimBot\height} {\trimWidth\width} {\trimHeight\height}, clip]{figs/kitti_5frames_#2opa/#1_must3r.png}} &
        \makecell{\adjincludegraphics[width=\linewidth,trim={\trimLeft\width} {\trimBot\height} {\trimWidth\width} {\trimHeight\height}, clip]{figs/kitti_5frames_#2opa/#1_VGGT.png}} &
        \makecell{\adjincludegraphics[width=\linewidth,trim={\trimLeft\width} {\trimBot\height} {\trimWidth\width} {\trimHeight\height}, clip]{figs/kitti_5frames_#2opa/#1_cut3r.png}} &
        \makecell{\adjincludegraphics[width=\linewidth,trim={\trimLeft\width} {\trimBot\height} {\trimWidth\width} {\trimHeight\height}, clip]{figs/kitti_5frames_#2opa/#1_AnySplat.png}} &
        \makecell{\adjincludegraphics[width=\linewidth,trim={\trimLeft\width} {\trimBot\height} {\trimWidth\width} {\trimHeight\height}, clip]{figs/kitti_5frames_#2opa/#1_occany512_original.png}} &
        \makecell{\adjincludegraphics[width=\linewidth,trim={\trimLeft\width} {\trimBot\height} {\trimWidth\width} {\trimHeight\height}, clip]{figs/kitti_5frames_#2opa/#1_ground_truth.png}}%
    }

	\newcommand{\trimNuScenesWidth}{0.35}
	\newcommand{\trimNuScenesHeight}{0.34}
	\newcommand{\trimNuScenesLeft}{0.22}
	\newcommand{\trimNuScenesBot}{0.26}

    \newcommand{\occRow}[2]{%
        \makecell{\adjincludegraphics[width=\linewidth]{figs/#2/#1_stacked_recon.png}} &
        \makecell{\adjincludegraphics[width=\linewidth,trim={\trimNuScenesLeft\width} {\trimNuScenesBot\height} {\trimNuScenesWidth\width} {\trimNuScenesHeight\height}, clip]{figs/#2/#1_must3r.png}} &
        \makecell{\adjincludegraphics[width=\linewidth,trim={\trimNuScenesLeft\width} {\trimNuScenesBot\height} {\trimNuScenesWidth\width} {\trimNuScenesHeight\height}, clip]{figs/#2/#1_VGGT.png}} &
        \makecell{\adjincludegraphics[width=\linewidth,trim={\trimNuScenesLeft\width} {\trimNuScenesBot\height} {\trimNuScenesWidth\width} {\trimNuScenesHeight\height}, clip]{figs/#2/#1_cut3r.png}} &
        \makecell{\adjincludegraphics[width=\linewidth,trim={\trimNuScenesLeft\width} {\trimNuScenesBot\height} {\trimNuScenesWidth\width} {\trimNuScenesHeight\height}, clip]{figs/#2/#1_AnySplat.png}} &
        \makecell{\adjincludegraphics[width=\linewidth,trim={\trimNuScenesLeft\width} {\trimNuScenesBot\height} {\trimNuScenesWidth\width} {\trimNuScenesHeight\height}, clip]{figs/#2/#1_occany512_original.png}} &
        \makecell{\adjincludegraphics[width=\linewidth,trim={\trimNuScenesLeft\width} {\trimNuScenesBot\height} {\trimNuScenesWidth\width} {\trimNuScenesHeight\height}, clip]{figs/#2/#1_ground_truth.png}}
    }

    \begin{tabular}{m{0.02\textwidth} m{0.11\textwidth}m{\imgsizer}m{\imgsizer}m{\imgsizer}m{\imgsizer}m{\imgsizer}m{\imgsizer}}
        & \multicolumn{1}{c}{Input\vphantom{$^\dagger$}} & \multicolumn{1}{c}{\muster{}\vphantom{$^\dagger$}} & \multicolumn{1}{c}{VGGT$^\dagger$} & \multicolumn{1}{c}{\cuter{}*\vphantom{$^\dagger$}} & \multicolumn{1}{c}{{AnySplat*}$^\dagger$} & \multicolumn{1}{c}{\OURS{} (ours)\vphantom{$^\dagger$}} & \multicolumn{1}{c}{GT\vphantom{$^\dagger$}} \\
        \midrule
        \multirow{2}{*}{\rotatebox[origin=c]{90}{\parbox{18cm}{\centering\textbf{\Surround{}}}}} & \occRow{scene-0003_1c8ec6c1dd614d3fb8b8849dc8756a9a}{nuscenes_surround_10opa}\\
        & \occRow{scene-0003_2f5de0aeca704127925cf8490ff5a21d}{nuscenes_surround_10opa}\\
        & \occRow{scene-0003_57a79294b87e4b55a7bcf58f7f8c6326}{nuscenes_surround_10opa}\\
        & \occRow{scene-0003_77f6617716654cfb8e313f0cd69e04a6}{nuscenes_surround_10opa}\\
        & \occRow{scene-0096_5edf36a058ce4849bc6371629bea8a69}{nuscenes_surround_10opa}\\
        & \occRow{scene-0101_09e58f5b8acb45a9a9c8c186895efa59}{nuscenes_surround_10opa}\\
        & \occRow{scene-0101_c0efe2319882495b9d13add1d352bee4}{nuscenes_surround_10opa}\\
        & \occRow{scene-0104_9709a4b7190f4e7a8f6edcd05316b811}{nuscenes_surround_10opa}\\
        & \occRow{scene-0277_3e991036257e4acda7cc1658dadd3e50}{nuscenes_surround_10opa}\\
        & \occRow{scene-0278_3eb6673bad8c44e2853283d958a68ca0}{nuscenes_surround_10opa}\\
    \end{tabular}
	\caption{\textbf{Occupancy predictions} of \OURS{} and baselines on surround-view data. Voxel colorization follows~\cref{fig:supp_qual_res_sequence}. Compared to baselines, our occupancy predictions are denser and more accurate.}
	
	\label{fig:supp_qual_res_surround}
\end{figure*}

\begin{figure*}
    \centering
    \large
    \setlength{\tabcolsep}{0pt}
    \renewcommand{\arraystretch}{0}
    \newcommand{\imgsizer}{.28\linewidth}
    \newcommand{\pcaRow}[2]{%
    \includegraphics[width=\imgsizer, trim=#2, clip]{figs/pca/#1_saved_colors.png} &
    \includegraphics[width=\imgsizer, trim=#2, clip]{figs/pca/#1_pca_image_embed.png} &
    \includegraphics[width=\imgsizer, trim=#2, clip]{figs/pca/#1_pca_high_res_1.png} &
    \includegraphics[width=\imgsizer, trim=#2, clip]{figs/pca/#1_pca_high_res_0.png}}
    
    \resizebox{.68\textwidth}{!}{%
    \begin{tabular}{cccc}    
        RGB & low & med & high \\ 
        \pcaRow{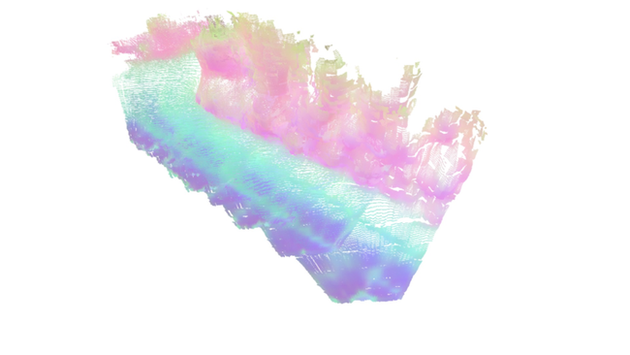}{1cm 1cm 1cm 1cm} \\
        \pcaRow{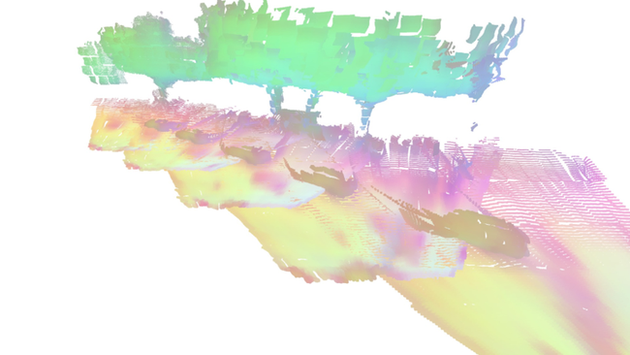}{1cm 1cm 1cm 1cm} \\
        \pcaRow{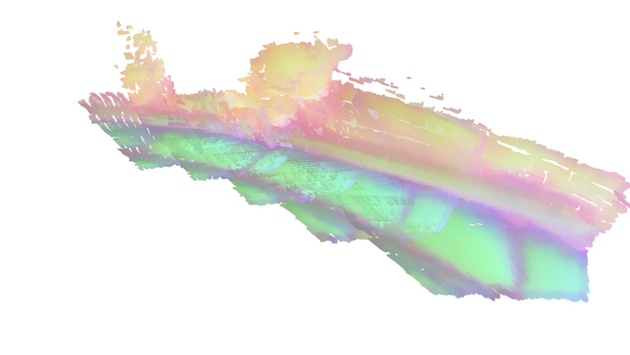}{1cm 1cm 1cm 1cm} \\
        \pcaRow{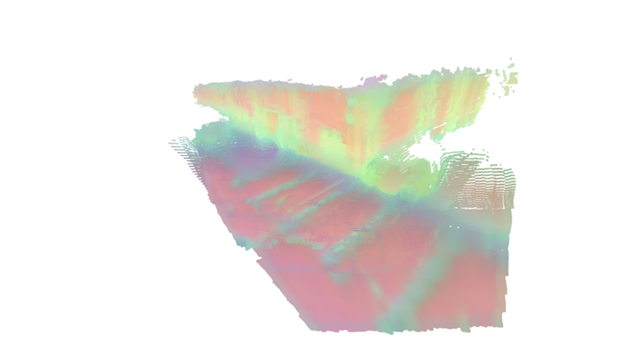}{1cm 1cm 1cm 1cm} \\
        \pcaRow{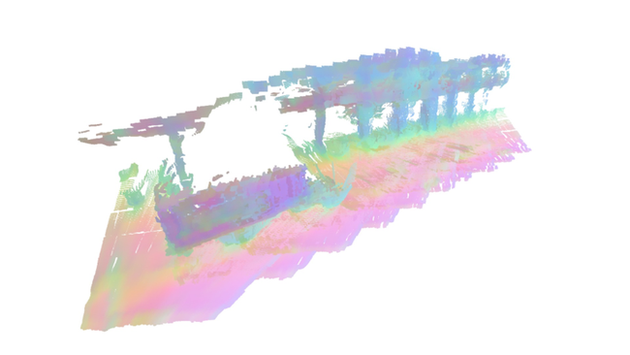}{1cm 1cm 1cm 1cm} \\
        \pcaRow{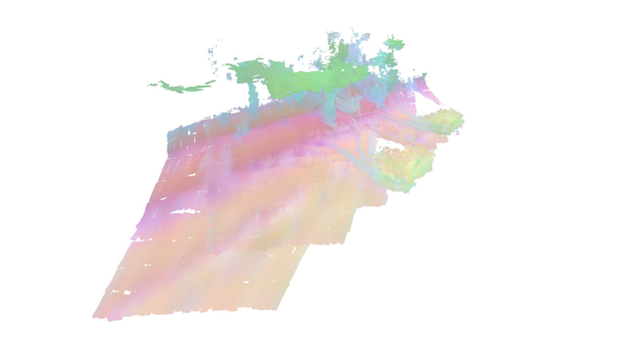}{1cm 1cm 1cm 1cm} \\
        \pcaRow{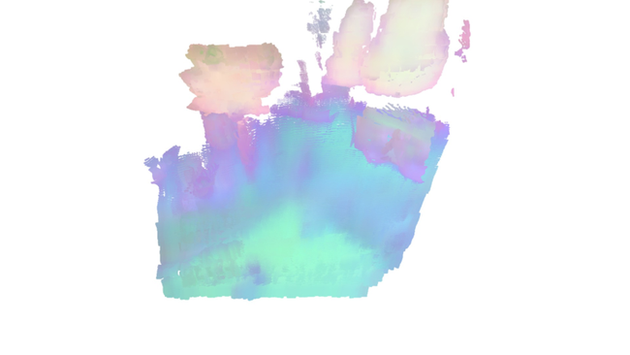}{1cm 1cm 1cm 1cm} \\
        \pcaRow{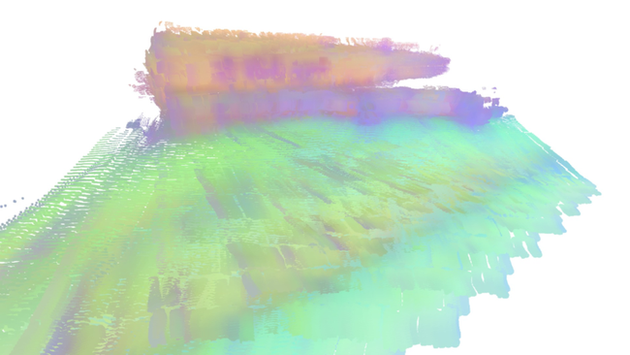}{1cm 1cm 1cm 1cm} \\
        \pcaRow{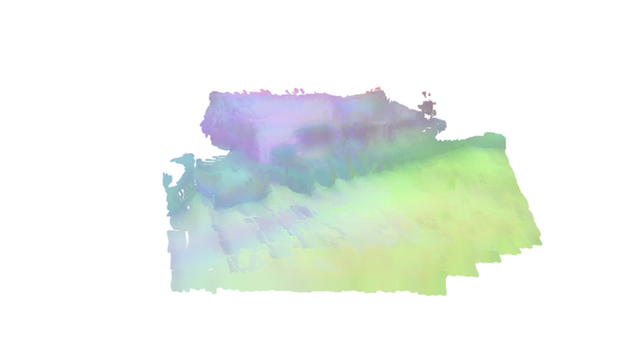}{1cm 1cm 1cm 1cm} \\
        \pcaRow{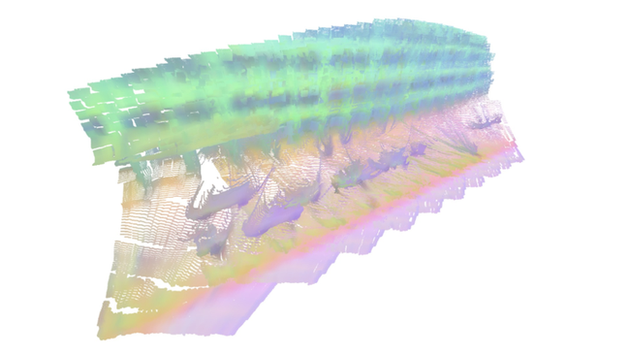}{1cm 1cm 1cm 1cm} \\
        \pcaRow{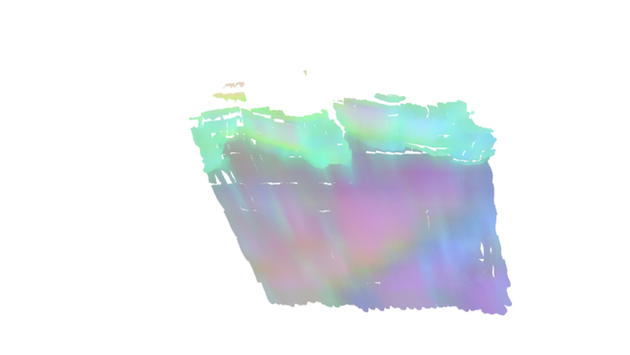}{1cm 1cm 1cm 1cm} \\
        \pcaRow{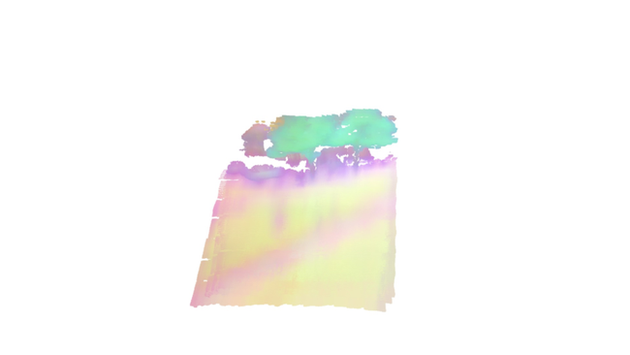}{1cm 1cm 1cm 1cm} \\

    \end{tabular}
    }
    \caption{\textbf{PCA visualization of predicted feature maps.} Low-resolution features capture high-level semantics (e.g., separating cars, buildings, and roads), while high-resolution features capture low-level details such as boundaries and textures. Features remain consistent across different views.}

    \label{fig:supp_pca}
\end{figure*}

We show additional qualitative results in~\cref{fig:supp_qual_res_sequence},~\cref{fig:supp_qual_res_surround},~\cref{fig:supp_qualitative_ablation_kitti},~\cref{fig:supp_qualitative_ablation_nuscenes}, and~\cref{fig:supp_pca}.

\begin{figure*}[ht!]
	\centering
	\small
	\setlength{\tabcolsep}{0pt}
	\renewcommand{\arraystretch}{0}
	\newcommand{\imgsizer}{.24\textwidth}
	\newcommand{\trimWidth}{0.45}
	\newcommand{\trimHeight}{0.4}
	\newcommand{\trimLeft}{0.1}
	\newcommand{\trimBot}{0.05}
	\newcommand{\kittiOccRow}[2]{%
        \makecell{\adjincludegraphics[width=0.6\linewidth]{figs/kitti_5frames_#2opa/#1_stacked_recon.png}} &
        \makecell{\adjincludegraphics[width=\linewidth,trim={\trimLeft\width} {\trimBot\height} {\trimWidth\width} {\trimHeight\height}, clip]{figs/kitti_5frames_#2opa/#1_occany512_noDistill.png}} &
        \makecell{\adjincludegraphics[width=\linewidth,trim={\trimLeft\width} {\trimBot\height} {\trimWidth\width} {\trimHeight\height}, clip]{figs/kitti_5frames_#2opa/#1_occany512_original.png}} &
        \makecell{\adjincludegraphics[width=\linewidth,trim={\trimLeft\width} {\trimBot\height} {\trimWidth\width} {\trimHeight\height}, clip]{figs/kitti_5frames_#2opa/#1_ground_truth.png}}
    }

	\newcommand{\kittiOccRowNoTTVA}[2]{%
        \makecell{\adjincludegraphics[width=0.6\linewidth]{figs/kitti_5frames_#2opa/#1_stacked_recon.png}} &
        \makecell{\adjincludegraphics[width=\linewidth,trim={\trimLeft\width} {\trimBot\height} {\trimWidth\width} {\trimHeight\height}, clip]{figs/kitti_5frames_#2opa/#1_occany512_noTTVA.png}} &
        \makecell{\adjincludegraphics[width=\linewidth,trim={\trimLeft\width} {\trimBot\height} {\trimWidth\width} {\trimHeight\height}, clip]{figs/kitti_5frames_#2opa/#1_occany512_original.png}} &
        \makecell{\adjincludegraphics[width=\linewidth,trim={\trimLeft\width} {\trimBot\height} {\trimWidth\width} {\trimHeight\height}, clip]{figs/kitti_5frames_#2opa/#1_ground_truth.png}}
    }

	\newcommand{\trimNuScenesWidth}{0.35}
	\newcommand{\trimNuScenesHeight}{0.33}
	\newcommand{\trimNuScenesLeft}{0.22}
	\newcommand{\trimNuScenesBot}{0.25}
	\newcommand{\occRow}[2]{%
		\adjincludegraphics[width=\linewidth]{figs/#2/#1_stacked_recon.png} &
		\adjincludegraphics[width=\linewidth,trim={\trimNuScenesLeft\width} {\trimNuScenesBot\height} {\trimNuScenesWidth\width} {\trimNuScenesHeight\height}, clip]{figs/#2/#1_occany512_noDistill.png} &
		\adjincludegraphics[width=\linewidth,trim={\trimNuScenesLeft\width} {\trimNuScenesBot\height} {\trimNuScenesWidth\width} {\trimNuScenesHeight\height}, clip]{figs/#2/#1_occany512_original.png} &
		\adjincludegraphics[width=\linewidth,trim={\trimNuScenesLeft\width} {\trimNuScenesBot\height} {\trimNuScenesWidth\width} {\trimNuScenesHeight\height}, clip]{figs/#2/#1_ground_truth.png} 
	}

	\newcommand{\occRowNoTTVA}[2]{%
		\adjincludegraphics[width=\linewidth]{figs/#2/#1_stacked_recon.png} &
		\adjincludegraphics[width=\linewidth,trim={\trimNuScenesLeft\width} {\trimNuScenesBot\height} {\trimNuScenesWidth\width} {\trimNuScenesHeight\height}, clip]{figs/#2/#1_occany512_noTTVA.png} &
		\adjincludegraphics[width=\linewidth,trim={\trimNuScenesLeft\width} {\trimNuScenesBot\height} {\trimNuScenesWidth\width} {\trimNuScenesHeight\height}, clip]{figs/#2/#1_occany512_original.png} &
		\adjincludegraphics[width=\linewidth,trim={\trimNuScenesLeft\width} {\trimNuScenesBot\height} {\trimNuScenesWidth\width} {\trimNuScenesHeight\height}, clip]{figs/#2/#1_ground_truth.png} 
	}
	
	\begin{subfigure}{0.49\textwidth}
		\centering
        \setlength{\tabcolsep}{2pt}
		\begin{tabular}{m{0.03\textwidth} m{0.16\textwidth}m{\imgsizer}m{\imgsizer}m{\imgsizer}}
			& \multicolumn{1}{c}{Input\vphantom{$^\dagger$}} & \multicolumn{1}{c}{w/o forcing} & \multicolumn{1}{c}{w/ forcing\vphantom{$^\dagger$}} & \multicolumn{1}{c}{GT\vphantom{$^\dagger$}} \\
			\midrule
			\multirow{2}{*}{\rotatebox[origin=c]{90}{\parbox{18cm}{\centering\textbf{\Sequence{}}}}} 
            & \kittiOccRow{08_000020}{3} \\
            & \kittiOccRow{08_000745}{3} \\
            & \kittiOccRow{08_001515}{3} \\
            & \kittiOccRow{08_003340}{3} \\
            & \kittiOccRow{08_000020}{5} \\
            & \kittiOccRow{08_000645}{5} \\
            & \kittiOccRow{08_000745}{5} \\
            & \kittiOccRow{08_001475}{5} \\
            & \kittiOccRow{08_001515}{5} \\
            & \kittiOccRow{08_003340}{5} \\
		\end{tabular}
		\caption{Segmentation Forcing}
		\label{fig:qualitative-occupancy-a}
	\end{subfigure}
	\hfill
	\begin{subfigure}{0.49\textwidth}
		\centering
        \setlength{\tabcolsep}{2pt}
		\begin{tabular}{m{0.03\textwidth} m{0.16\textwidth}m{\imgsizer}m{\imgsizer}m{\imgsizer}}
			 & \multicolumn{1}{c}{Input\vphantom{$^\dagger$}} & \multicolumn{1}{c}{w/o NVR} & \multicolumn{1}{c}{w/ NVR\vphantom{$^\dagger$}} & \multicolumn{1}{c}{GT\vphantom{$^\dagger$}} \\
			\midrule
            \multirow{2}{*}{\rotatebox[origin=c]{90}{\parbox{18cm}{\centering\textbf{\Sequence{}}}}}
            & \kittiOccRowNoTTVA{08_001165}{3} \\
            & \kittiOccRowNoTTVA{08_001175}{3} \\
            & \kittiOccRowNoTTVA{08_001240}{3} \\
            & \kittiOccRowNoTTVA{08_001675}{3} \\
            & \kittiOccRowNoTTVA{08_002920}{3} \\
            & \kittiOccRowNoTTVA{08_003225}{3} \\
            & \kittiOccRowNoTTVA{08_003250}{3} \\ 
            & \kittiOccRowNoTTVA{08_003345}{3}\\
            & \kittiOccRowNoTTVA{08_002045}{3}\\
            & \kittiOccRowNoTTVA{08_001575}{3}\\
		\end{tabular}
		\caption{Novel-View Rendering}
		\label{fig:qualitative-occupancy-b}
	\end{subfigure}
	
	\caption{\textbf{Qualitative ablation on Semantic KITTI} shows the gains from \emph{Segmentation Forcing} and \emph{Novel-View Rendering}. Voxel colorization follows~\cref{fig:supp_qual_res_sequence}. The two proposed strategies significantly improve the density and the accuracy of occupancy predictions.}
	\label{fig:supp_qualitative_ablation_kitti}
\end{figure*}

\begin{figure*}[ht!]
	\centering
	\small
	\setlength{\tabcolsep}{0pt}
	\renewcommand{\arraystretch}{0}
	\newcommand{\imgsizer}{.24\textwidth}
	\newcommand{\trimWidth}{0.45}
	\newcommand{\trimHeight}{0.4}
	\newcommand{\trimLeft}{0.1}
	\newcommand{\trimBot}{0.05}
	\newcommand{\kittiOccRow}[2]{%
        \makecell{\adjincludegraphics[width=0.6\linewidth]{figs/kitti_5frames_#2opa/#1_stacked_recon.png}} &
        \makecell{\adjincludegraphics[width=\linewidth,trim={\trimLeft\width} {\trimBot\height} {\trimWidth\width} {\trimHeight\height}, clip]{figs/kitti_5frames_#2opa/#1_occany512_noDistill.png}} &
        \makecell{\adjincludegraphics[width=\linewidth,trim={\trimLeft\width} {\trimBot\height} {\trimWidth\width} {\trimHeight\height}, clip]{figs/kitti_5frames_#2opa/#1_occany512_original.png}} &
        \makecell{\adjincludegraphics[width=\linewidth,trim={\trimLeft\width} {\trimBot\height} {\trimWidth\width} {\trimHeight\height}, clip]{figs/kitti_5frames_#2opa/#1_ground_truth.png}}
    }

	\newcommand{\kittiOccRowNoTTVA}[2]{%
        \makecell{\adjincludegraphics[width=0.6\linewidth]{figs/kitti_5frames_#2opa/#1_stacked_recon.png}} &
        \makecell{\adjincludegraphics[width=\linewidth,trim={\trimLeft\width} {\trimBot\height} {\trimWidth\width} {\trimHeight\height}, clip]{figs/kitti_5frames_#2opa/#1_occany512_noTTVA.png}} &
        \makecell{\adjincludegraphics[width=\linewidth,trim={\trimLeft\width} {\trimBot\height} {\trimWidth\width} {\trimHeight\height}, clip]{figs/kitti_5frames_#2opa/#1_occany512_original.png}} &
        \makecell{\adjincludegraphics[width=\linewidth,trim={\trimLeft\width} {\trimBot\height} {\trimWidth\width} {\trimHeight\height}, clip]{figs/kitti_5frames_#2opa/#1_ground_truth.png}}
    }

	\newcommand{\trimNuScenesWidth}{0.35}
	\newcommand{\trimNuScenesHeight}{0.33}
	\newcommand{\trimNuScenesLeft}{0.22}
	\newcommand{\trimNuScenesBot}{0.25}
	\newcommand{\occRow}[2]{%
		\adjincludegraphics[width=\linewidth]{figs/#2/#1_stacked_recon.png} &
		\adjincludegraphics[width=\linewidth,trim={\trimNuScenesLeft\width} {\trimNuScenesBot\height} {\trimNuScenesWidth\width} {\trimNuScenesHeight\height}, clip]{figs/#2/#1_occany512_noDistill.png} &
		\adjincludegraphics[width=\linewidth,trim={\trimNuScenesLeft\width} {\trimNuScenesBot\height} {\trimNuScenesWidth\width} {\trimNuScenesHeight\height}, clip]{figs/#2/#1_occany512_original.png} &
		\adjincludegraphics[width=\linewidth,trim={\trimNuScenesLeft\width} {\trimNuScenesBot\height} {\trimNuScenesWidth\width} {\trimNuScenesHeight\height}, clip]{figs/#2/#1_ground_truth.png} 
	}

	\newcommand{\occRowNoTTVA}[2]{%
		\adjincludegraphics[width=\linewidth]{figs/#2/#1_stacked_recon.png} &
		\adjincludegraphics[width=\linewidth,trim={\trimNuScenesLeft\width} {\trimNuScenesBot\height} {\trimNuScenesWidth\width} {\trimNuScenesHeight\height}, clip]{figs/#2/#1_occany512_noTTVA.png} &
		\adjincludegraphics[width=\linewidth,trim={\trimNuScenesLeft\width} {\trimNuScenesBot\height} {\trimNuScenesWidth\width} {\trimNuScenesHeight\height}, clip]{figs/#2/#1_occany512_original.png} &
		\adjincludegraphics[width=\linewidth,trim={\trimNuScenesLeft\width} {\trimNuScenesBot\height} {\trimNuScenesWidth\width} {\trimNuScenesHeight\height}, clip]{figs/#2/#1_ground_truth.png} 
	}
	
	\begin{subfigure}{0.49\textwidth}
		\centering
        \setlength{\tabcolsep}{2pt}
		\begin{tabular}{m{0.03\textwidth} m{0.16\textwidth}m{\imgsizer}m{\imgsizer}m{\imgsizer}}
			& \multicolumn{1}{c}{Input\vphantom{$^\dagger$}} & \multicolumn{1}{c}{w/o forcing} & \multicolumn{1}{c}{w/ forcing\vphantom{$^\dagger$}} & \multicolumn{1}{c}{GT\vphantom{$^\dagger$}} \\
            \midrule
			\multirow{10}{*}{\rotatebox[origin=c]{90}{\parbox{12cm}{\centering\textbf{\Surround{}}}}} 
            & \occRow{scene-0003_08254b97198045d98009af51d144d147}{nuscenes_surround_10opa}\\
            & \occRow{scene-0003_b214d0d42c404d4f925aa56ee9cac607}{nuscenes_surround_10opa}\\
            & \occRow{scene-0273_49ceba17e48d41f08da1c994618f84da}{nuscenes_surround_10opa}\\
            & \occRow{scene-0277_10341c91fa1347e1a0b607ae5eff471c}{nuscenes_surround_10opa}\\
            & \occRow{scene-0277_74f6b583f9eb441db5b95b67cbec71f5}{nuscenes_surround_10opa}\\
            & \occRow{scene-0346_a4a9d61254d148fba35d53277b5246f8}{nuscenes_surround_10opa}\\
            & \occRow{scene-0928_0ed87ec201a04c3581bf8328600f0f7b}{nuscenes_surround_10opa}\\
            & \occRow{scene-0969_93d2488a9e3e47088d7334164e1625d4}{nuscenes_surround_10opa}\\
            & \occRow{scene-0968_0046092508b14f40a86760d11f9896bb}{nuscenes_surround_10opa}\\
            & \occRow{scene-0968_22cc5dafc805425b9bcec94512093825}{nuscenes_surround_10opa}\\
		\end{tabular}
		\caption{Segmentation Forcing}
		\label{fig:qualitative-occupancy-a}
	\end{subfigure}
	\hfill
	\begin{subfigure}{0.49\textwidth}
		\centering
        \setlength{\tabcolsep}{2pt}
		\begin{tabular}{m{0.03\textwidth} m{0.16\textwidth}m{\imgsizer}m{\imgsizer}m{\imgsizer}}
			 & \multicolumn{1}{c}{Input\vphantom{$^\dagger$}} & \multicolumn{1}{c}{w/o NVR} & \multicolumn{1}{c}{w/ NVR\vphantom{$^\dagger$}} & \multicolumn{1}{c}{GT\vphantom{$^\dagger$}} \\
            \midrule
			\multirow{10}{*}{\rotatebox[origin=c]{90}{\parbox{12cm}{\centering\textbf{\Surround{}}}}} 
            & \occRowNoTTVA{scene-0039_e46803a34fa744ad9372b5227798e00a}{nuscenes_surround_10opa} \\  
            & \occRowNoTTVA{scene-0273_c0ac3dc491e7414184f44c2b32d24ad8}{nuscenes_surround_10opa} \\  
            & \occRowNoTTVA{scene-0278_c1255d993df4492a9dfa2f67a50f3073}{nuscenes_surround_10opa} \\  
            & \occRowNoTTVA{scene-0522_2ba2a8fb2f0f475ba9ca6d6889ec4d27}{nuscenes_surround_10opa} \\  
            & \occRowNoTTVA{scene-0905_75fc8c40884f4132b4017d3b29500a71}{nuscenes_surround_10opa} \\  
            & \occRowNoTTVA{scene-0915_20fac9beda19441d87bd522015a11b67}{nuscenes_surround_10opa} \\  
            & \occRowNoTTVA{scene-0930_5a624b8b0bbc4435a57f30b192a0fd46}{nuscenes_surround_10opa} \\    
            & \occRowNoTTVA{scene-0968_311d6ba64bd145d8b552328d8e3e3339}{nuscenes_surround_10opa} \\
            & \occRowNoTTVA{scene-0968_22cc5dafc805425b9bcec94512093825}{nuscenes_surround_10opa} \\
            & \occRowNoTTVA{scene-0035_de550f4c35284f90844c07a64253639b}{nuscenes_surround_10opa} \\
		\end{tabular}
		\caption{Novel-View Rendering}
		\label{fig:qualitative-occupancy-b}
	\end{subfigure}
	
	\caption{\textbf{Qualitative ablation on Occ3D-NuScenes} shows the gains from \emph{Segmentation Forcing} and \emph{Novel-View Rendering}. Voxel colorization follows~\cref{fig:supp_qual_res_sequence}. The two proposed strategies significantly improve the density and the accuracy of occupancy predictions.}
	\label{fig:supp_qualitative_ablation_nuscenes}
\end{figure*}

    \end{document}